\documentclass{article}
\usepackage{geometry}
\usepackage[american]{babel}

\usepackage{hyperref}
\usepackage{natbib} 
\usepackage{mathtools} 
\usepackage{booktabs} 
\usepackage{tikz} 
\usepackage{arxiv}

\title{Achieving Limited Adaptivity for Multinomial Logistic Bandits}

\author{%
  \begin{tabular}[t]{@{}c@{\hspace{1cm}}c@{\hspace{1cm}}c@{}}
    \begin{tabular}[t]{@{}c@{}}
      \textbf{Sukruta Prakash Midigeshi}\\
      \textnormal{Microsoft Research India}\\
      \texttt{\href{mailto:t-smidigeshi@microsoft.com}{t-smidigeshi@microsoft.com}}
    \end{tabular}
    &
    \begin{tabular}[t]{@{}c@{}}
      \textbf{Tanmay Goyal}\\
      \textnormal{Microsoft Research India}\\
      \texttt{\href{mailto:t-tangoyal@microsoft.com}{t-tangoyal@microsoft.com}}
    \end{tabular}
    &
    \begin{tabular}[t]{@{}c@{}}
      \textbf{Gaurav Sinha}\\
      \textnormal{Microsoft Research India}\\
      \texttt{\href{mailto:gauravsinha@microsoft.com}{gauravsinha@microsoft.com}}
    \end{tabular}
  \end{tabular}
}

\date{}

\renewcommand{\headeright}{}
\renewcommand{\undertitle}{}

\usepackage{amssymb, amsmath}
\usepackage{amsthm}
\usepackage{bm}
\usepackage{wrapfig}
\usepackage{multirow}
\usepackage{graphicx}
\usepackage{subcaption}

\usepackage{algorithmic, algorithm}
\usepackage{bbm}

\theoremstyle{definition}
\newtheorem{definition}{Definition}[section]

\theoremstyle{plain}
\newtheorem{lemma}{Lemma}[section]
\newtheorem{theorem}{Theorem}[section]

\newtheorem{remark}{Remark}[section]

\DeclareMathOperator*{\argmax}{arg\,max}
\DeclareMathOperator*{\argmin}{arg\,min}
\DeclareMathOperator*{\E}{\mathbb{E}}

\def\DEBUG{\FALSE} 

\ifdefined\DEBUG
  \usepackage{xcolor} 
  \newcommand{\tanmay}[1]{\textcolor{red}{[tanmay:#1]}}
  \newcommand{\sukruta}[1]{\textcolor{magenta}{[sukruta:#1]}}
  \newcommand{\gaurav}[1]{\textcolor{blue}{[gaurav:#1]}}
\else
  \newcommand{\tanmay}[1]{}
  \newcommand{\sukruta}[1]{}
  \newcommand{\gaurav}[1]{}
\fi

\newcommand{\R}{\mathbb{R}}

\renewcommand{\P}{\mathbb{P}}

\newcommand{\X}{\mathcal{X}}

\renewcommand{\O}{\mathcal{O}}

\newcommand{\inv}{^{-1}}
\newcommand*\diff{\mathop{}\!\mathrm{d}}

\newcommand{\eigmax}[1]{\lambda_{max}\pbrak{#1}}
\newcommand{\eigmin}[1]{\lambda_{min}\pbrak{#1}}

\newcommand{\mleq}{\preccurlyeq}
\newcommand{\mgeq}{\succcurlyeq}

\newcommand{\twonorm}[1]{\left|\left|#1\right|\right|_2}
\newcommand{\norm}[1]{\left|\left|#1\right|\right|}
\newcommand{\matnorm}[2]{\left|\left|#1\right|\right|_{#2}}

\newcommand{\sbrak}[1]{\left[#1\right]}
\newcommand{\pbrak}[1]{\left(#1\right)}
\newcommand{\cbrak}[1]{\left\{#1\right\}}
\newcommand{\modulus}[1]{\left|#1\right|}

\newcommand{\thetastar}{\bm{\theta}^\star}

\begin{document}
\maketitle

\begin{abstract}
Multinomial Logistic Bandits have recently attracted much attention due to their ability to model problems with multiple outcomes. 
In this setting, each decision is associated with many possible outcomes, modeled using a multinomial logit function. Several recent works on multinomial logistic bandits have simultaneously achieved optimal regret and computational efficiency. However, motivated by real-world challenges and practicality, there is a need to develop algorithms with limited adaptivity, wherein we are allowed only $M$ policy updates. 
To address these challenges, we present two algorithms, \texttt{B-MNL-CB} and \texttt{RS-MNL}, that operate in the batched and rarely-switching paradigms, respectively. The batched setting involves choosing the $M$ policy update rounds at the start of the algorithm, while the rarely-switching setting can choose these $M$ policy update rounds in an adaptive fashion. Our first algorithm, \texttt{B-MNL-CB} extends the notion of distributional optimal designs to the multinomial setting and achieves $\tilde{O}(\sqrt{T})$ regret assuming the contexts are generated stochastically when presented with $\Omega(\log \log T)$ update rounds. Our second algorithm, \texttt{RS-MNL} works with adversarially generated contexts and can achieve $\tilde{O}(\sqrt{T})$ regret with $\tilde{O}(\log T)$ policy updates.
Further, we conducted experiments that demonstrate that our algorithms (with a fixed number of policy updates) are extremely competitive (and often better) than several state-of-the-art baselines (which update their policy every round), showcasing the applicability of our algorithms in various practical scenarios. 

\end{abstract}

\section{Introduction}\label{sec:intro}

Contextual Bandits help incorporate additional information that a learner may have with the standard Multi-Armed Bandit (MAB) setting. In this setting, at each round, the learner is presented with a set of arms and is expected to choose an arm. She is also presented with a context vector that helps guide the decisions she makes. For each decision, the learner receives a reward, which is generated using a hidden optimal parameter. The goal of the learner is to minimize her cumulative regret (or equivalently, maximize her cumulative reward), over a specified number of rounds $T$. Contextual Bandits have long been studied under various notions of reward models and settings. For instance, one of the simplest models is to assume that the expected reward is a linear function of the arms and the hidden parameter \citep{AbbasiYadkori2011, auer2003, chu2011}. This was later extended to non-linear settings such as the logistic setting
\citep{Faury2020, Abeille2021, Faury2022}, generalized linear setting 
\citep{Filippi2010, Li2017}, and the multinomial setting \citep{Amani2021, Zhang2023}. In this work, we specifically focus on the multinomial setting that can model problems with multiple outcomes, which makes this setting incredibly useful in the fields of machine and reinforcement learning, as well as, in real life.

Though significant progress has been made in designing algorithms for the contextual setting, the algorithms do not demonstrate a lot of applicability. There has been growing interest in constraining the budget available for algorithmic updates. This \emph{limited adaptivity} setting is crucial in real-world applications, where frequent updates can hinder parallelism and large-scale deployment. Additionally, practical and computational constraints may make it infeasible to make policy updates at every time step. For example, in clinical trials \citep{international1997international}, the treatments made available to the patients cannot be changed with every patient. Thus, the updates are made after administering the treatment to a group of patients, observing the effects and outcomes, and then updating the treatment. We observe a similar tendency in online advertising and recommendations, where it is difficult to update the policy at each round due to resource constraints. A recent line of work \citep{Ruan2021, sawarni24} has introduced algorithms for contextual bandits in the linear and generalized linear settings, respectively. They introduce algorithms for two different settings: the \emph{batched} setting, wherein the policy update rounds are fixed at the start of the algorithm, and the \emph{rarely-switching} algorithm, wherein the policy update rounds are decided in an adaptive fashion. Since multinomial logistic bandits are not generalized linear models, it is not clear if the algorithms developed in past works would apply in this setting \citep{Amani2021}. Hence, the major focus of this work is to develop algorithms with limited adaptivity for the multinomial setting. We now list our contributions:

\subsection{Contributions}
\begin{itemize}
    \item We propose a new algorithm \texttt{B-MNL-CB}, which operates in the batched setting where the contexts are generated stochastically. The algorithm achieves $\tilde{O}(\sqrt T)$ regret with high probability, with $\Omega(\log \log T)$ policy updates. In order to accommodate time-varying contexts, we adapt the recently introduced concept of distributional optimal designs \citep{Ruan2021} to the multinomial logistic setting. This is done by introducing a new scaling technique to counter the non-linearity associated with the reward function. Note that the leading term of the regret bound is free of the instance-dependent non-linearity parameter $\kappa$, which can scale exponentially with the instance parameters (refer to Section \ref{section:notations} for more details).
    \item Our second algorithm, \texttt{RS-MNL} operates in the rarely-switching setting, where the contexts are generated adversarially. The algorithm achieves $\tilde{O}(\sqrt T)$ regret while performing $\tilde{O}(\log T)$ policy updates, each determined by a simple switching criterion. Further, our algorithm does not require a warmup switching criterion, unlike the rarely-switching algorithm in \cite{sawarni24}, which helps in reducing the number of switches from $\tilde{O}(\log^2 T)$ to $\tilde{O}(\log T)$.
    \item We empirically demonstrate the performance of our rarely-switching algorithm \texttt{RS-MNL}. Across a range of randomly selected instances, our algorithm achieves regret comparable to, and often better than, several logistic and multinomial logistic state-of-the-art baseline algorithms. Our algorithm manages to do so with a limited number of policy updates as compared to the baselines, which perform an update at each time round. We also empirically show that the number of switches made by our algorithm is $\tilde{O}(\log T)$, which is in agreement with our theoretical results.
    
\end{itemize}
\subsection{Related works}
The multinomial logistic setting was first studied by \cite{Amani2021}. They proposed an algorithm that achieved a regret bound of $\tilde{{O}}(\sqrt{\kappa T}) $, where $\kappa$ is the instance-dependent non-linearity parameter (defined in Section \ref{section:notations}). This was further improved by \cite{Zhang2023}, who proposed a computationally efficient algorithm with a regret bound of $\tilde{{O}}(\sqrt{T})$, thus achieving a $\kappa-$free bound (the leading term is free of $\kappa$). However, both of these algorithms face challenges in real-world deployment due to infrastructural and practical constraints associated with updating the policy at every round.

Thus, the limited adaptivity framework was introduced to combat this challenge, wherein the algorithm could only undergo a limited number of policy switches. This framework consists of two paradigms: the first being the \emph{Batched} Setting, where the batch lengths are predetermined and was first studied by \cite{Gao2019}, who showed that $\Omega(\log\log T)$ batches are necessary to obtain optimal minimax regret. The second setting is the \emph{Rarely Switching} Setting, first introduced by \cite{AbbasiYadkori2011}, where batch lengths are determined adaptively, based on a switching criterion, such as the determinant doubling trick, wherein the policy is updated every time the determinant of the information matrix doubles.

In the contextual setting, \cite{Ruan2021} used optimal designs to study the case where the arm sets themselves were generated stochastically, providing a bound of $\tilde{{O}}(\sqrt{dT\log d })$ for the batched setting. This idea was then extended to the generalized linear setting by \cite{sawarni24}, who proposed algorithms that could achieve $\kappa-$free regret in both the batched and rarely-switching settings (independent of $\kappa$ in the leading term). However, to the best of our knowledge, the limited adaptivity framework has not yet been explored in the multinomial setting. The primary focus of this work is to propose optimal limited-adaptivity algorithms for the multinomial setting. We achieve this by extending the results of \cite{sawarni24} and \cite{Ruan2021} to the multinomial setting in the batched setting while preserving the regret bound of \cite{Zhang2023} in the first-order term. In the rarely-switching setting, we further build upon the work of \cite{AbbasiYadkori2011} and \cite{sawarni24} to adapt it for the multinomial case. This maintains the regret bound of \cite{Zhang2023} while also reducing the number of switches as compared to \cite{sawarni24}.

\section{Preliminaries}
\label{section:notations}

\textbf{Notations:} We denote all vectors with bold lower case letters, matrices with bold upper case letters, and sets with upper case calligraphic symbols. We write $\bm{M}\mgeq 0$, if matrix $\bm{M}$ is positive semi-definite (p.s.d). For a p.s.d matrix $\bm{M}$, we define the norm of a vector $\bm{x}$ with respect to $\bm{M}$ as $\matnorm{\bm{x}}{\bm{M}} = \sqrt{\bm{x}^\top{\bm{M}}\bm{x}}$
and the spectral norm of $\bm{M}$ as $\twonorm{\bm{M}} = \sqrt{\eigmax{\bm{M}^\top\bm{M}}}$ where $\eigmax{\bm{M}}$ denotes the maximum eigenvalue of $\bm{M}$. We denote the set $\cbrak{1,\ldots, N}$ as $[N]$. The Kronecker product of matrices \(\bm A \in \mathbb{R}^{p \times q}\) and \(\bm B \in \mathbb{R}^{r \times s}\) is defined as \((\bm A \otimes \bm B)_{pr+v,\;qs+w} = \bm A_{rs} \cdot \bm B_{vw}\), resulting in a \(pr \times qs\) matrix, where $\bm M_{ij}$ denotes the element of the matrix $\bm M$ present at the $i^{th}$ row and the $j^{th}$ column. 
Finally, we use $\Delta(\mathcal{X})$
to denote the set of all probability distributions over $\mathcal{X}$. We use $\bm{I}_n$ to denote an identity matrix of dimension $n$, and we simply use $\bm{I}$ when the dimensions are clear from context.

\textbf{Multinomial Logistic Bandits}: In the Multinomial Logistic Bandit Setting, at each round $t \in [T]$, the learner is presented with a set of arms $\mathcal{X}_t \subseteq \mathbb{R}^d $
, and is expected to choose an arm $\bm{x}_t \in \X_t$. Based on the learner's choice, the environment provides an outcome $y_t \in [K]\cup \{0\}$\footnote{The outcome $0$ indicates \emph{no outcome}.}. While choosing the arm at round $t$, the learner can utilize all prior information, which can be encoded in the filtration $\mathcal{F}_t = \sigma\pbrak{\mathcal{F}_0 , \bm{x}_1 , y_1 , \ldots , \bm{x}_{t-1} , y_{t-1}}$, where $\mathcal{F}_0$ represents any prior information the learner had before starting the algorithm. The probability distribution over these $K+1$ outcomes is modeled using a multinomial logistic function\footnote{The multinomial logistic function is also referred to as the link function and would be used interchangeably throughout.} as follows:
\[\P\cbrak{y_t = i \mid \bm{x}_t , \mathcal{F}_t} = 
    \begin{cases}
    \frac{\exp\pbrak{\bm{x}_t^\top\bm{\theta}^*_i}}{1 + \sum\limits_{j=1}^{K}\exp\pbrak{\bm{x}_t^\top\bm{\theta}^*_j}}, & 1 \leq i \leq K,\\
    \frac{1}{1 + \sum\limits_{j=1}^{K}\exp\pbrak{\bm{x}_t^\top\bm{\theta}^*_j}}, & i = 0,
    \end{cases}
\]
where $\thetastar = \pbrak{{\bm{\theta}^\star_1}^\top , \ldots , {\bm{\theta}^\star_K}^\top}^\top \in \R^{dK}$ comprises the hidden optimal parameter vectors associated with each of the $K$ outcomes. Based on the outcome $y_t$, the learner receives a reward $\rho_{y_t} \geq 0$. It is standard to set $\rho_0 = 0$. We assume that the reward vector $\rho = \pbrak{\rho_1 , \ldots , \rho_K}^\top$ is fixed and known. We assume that $\twonorm{\thetastar} \leq S$, $\twonorm{\bm{\rho}} \leq R$, and $\twonorm{\bm{x}} \leq 1$, for all $\bm{x}\in \X_t$, where $R$ and $S$ are fixed and known beforehand.  Note that when $K=1$, the problem reduces to the binary logistic setting. For simplicity, we denote the probability of the $i^{\text{th}}$ outcome $\P\cbrak{y_t = i \mid \bm{x}_t , \mathcal{F}_t}$ as  $z_i(\bm{x}_t , \thetastar)$ and denote the probability vector over the $K$ outcomes as $\bm{z}(\bm{x}_t , \thetastar) = \pbrak{z_1(\bm{x}_t , \thetastar) , \ldots , z_K(\bm{x}_t , \thetastar)}^\top$. Then, it is easy to see that the expected reward of the learner at round $t$ is given by $\bm{\rho}^\top\bm{z}(\bm{x}_t , \thetastar)$. 
The goal of the learner is to choose an arm $\bm{x}_t, t\in[T]$ so as to minimize her regret, which can have different formulations based on the problem setting:
\begin{enumerate}
    \item Stochastic Contextual setting \label{Batched setting}: In this setting, at each time step, the feasible action sets are sampled from the same (unknown) distribution $\mathcal{D}$. Thus, the learner wishes to minimize her expected cumulative regret which is given by
    \[R(T) = \E\sbrak{\sum\limits_{t=1}^{T}\sbrak{\max\limits_{\bm{x}\in\mathcal{X}_t}\bm{\rho}^\top\bm{z}(\bm{x} , \thetastar) - \bm{\rho}^\top\bm{z}(\bm{x}_t , \thetastar)}}.\]
    Here, the expectation is over the distribution of the arm set $\mathcal{D}$ and the randomness inherently present in the algorithm. In this setting, we assume that only $M$ (fixed beforehand) policy updates can be made and the rounds at which these updates can happen need to be decided prior to starting the algorithm. 

    \item Adversarial Contextual setting \label{Adversarial setting}: In this setting, there are no assumptions made on how the feature vectors of the arms are generated. Thus, allowing $M$ policy updates, the algorithm can choose the rounds at which it updates its policy during the course of the algorithm. These dynamic updates are based on a simple switching criterion similar to the one presented in \cite{AbbasiYadkori2011}. In this setting, the learner wishes to minimize her cumulative regret given by
     \[R(T) = \sum\limits_{t=1}^{T}\sbrak{\max\limits_{\bm{x}\in\mathcal{X}_t}\bm{\rho}^\top\bm{z}(\bm{x} , \thetastar) - \bm{\rho}^\top\bm{z}(\bm{x}_t , \thetastar)}.\]
\end{enumerate}

\textbf{Discussion on the Instance-Dependent Non-Linearity Parameter $\kappa$: } Several works on the binary logistic model and generalized linear model
\citep{Filippi2010, Faury2020} as well as the multinomial logistic model \citep{Amani2021, Zhang2023} have mentioned the importance of an instance dependent, non-linearity parameter $\kappa$, and have stressed on the need to obtain regret guarantees independent of $\kappa$ (at least in the leading term). $\kappa$ was first defined for the binary logistic reward model setting \citep{Filippi2010}. A natural extension to the multinomial logit setting was recently proposed in \cite{Amani2021}. We use the same definition as \cite{Amani2021}, i.e.,

\[\kappa = \sup\bigg\{\frac{1}{\lambda_{min}(\bm{A}(\bm{x} , \bm{\theta)})}: \bm{x}\in\mathcal{X}_1\cup\ldots\cup\mathcal{X}_T, \bm{\theta}\in\Theta\bigg\},
\]

where 
$\bm{A}(\bm{x} , \bm{\theta}) = \nabla \bm{z}(\bm{x} , \bm{\theta)} = diag(\bm{z}(\bm{x} , \bm{\theta})) - \bm{z}(\bm{x} , \bm{\theta})\bm{z}(\bm{x} , \bm{\theta})^\top$, is the gradient of the link function $\bm{z}$ with respect to the vector $(\bm{I}_K \otimes \bm x^\top) \theta$. In Section $2$,
\cite{Faury2020}, it was highlighted that that $\kappa$ can grow exponentially in the instance parameters such as $S$ and therefore regret proportional to $\kappa$ could be detrimental when these parameters are large. 
In Section $3$ of \cite{Amani2021}, the authors show that $\kappa$ in the multinomial setting also scales exponentially with the diameter of the parameter and action sets. We direct the reader to Section 3 of \cite{Amani2021} for a more elaborate discussion on the importance of $\kappa$ in the multinomial setting.

\textbf{Optimal Design policies: } 
Optimal Experimental Designs are concerned with efficiently selecting the best data points so as to minimize the variance (or equivalently, maximize the information)  of estimated parameters. For a set of points $\mathcal{X} \subseteq \mathbb{R}^d$ 
and some distribution $\pi$ defined on $\mathcal{X}$, The information matrix is defined as  $ (\E_{\bm{x} \sim \pi}\bm{x}\bm{x}^\top)^{-1}$. Several criteria are used to maximize the information, some of which are A-Criterion (minimize trace of the information matrix), E-Criterion (maximize the minimum eigenvalue of the information matrix), and D-Criterion (maximize the determinant of the information matrix). One of the popular criteria used in bandit literature is the G-Optimal Design which is defined as follows: 
\begin{definition}
\textbf{G-Optimal Design:}  
For a set $\mathcal{X} \subseteq \mathbb{R}^d$, the G-Optimal design $\pi_G(\mathcal{X})$ is the solution to the following optimization problem:
\[
\min_{\pi \in \Delta(\mathcal{X})} \max_{\bm{x} \in \mathcal{X}} \|\bm{x}\|^2_{V(\pi)^{-1}}, \quad \text{where} \quad \bm{V}(\pi) = \mathbb{E}_{\bm{x} \sim \pi}[\bm{x}\bm{x}^\top].
\]
\end{definition}

The General Equivalence Theorem \citep{Kiefer_Wolfowitz_1960, Lattimore_Szepesvári_2020} establishes an equivalence between the G-Optimal and D-Optimal criteria. Specifically, it shows that for any set $\mathcal{X} \subseteq \mathbb{R}^d$, there exists a G-Optimal design $\pi_G(\mathcal{X}) \in \Delta(\mathcal{X})$ such that:
\[
\|\bm{x}\|^2_{V(\pi)^{-1}} \leq d \quad \forall \bm{x} \in \mathcal{X}.
\]
Furthermore, if $\mathcal{X}$ is a discrete set with finite cardinality, one can find a G-Optimal design in poly-time with respect to $|\mathcal{X}|$ such that the right-hand side can be relaxed to $2d$ (Lemma 3, \cite{Ruan2021}).

\textbf{Distributional Optimal design}: The extension of the G-Optimal design to the stochastic contextual setting worsens the bound on $\|\bm{x}\|^2_{V(\pi)^{-1}}$, i.e, in the worst case, the expected variance $\|\bm{x}\|^2_{V(\pi)^{-1}}$ is now upper bounded by $d^2$, where the expectation is over the arm set $\mathcal{X}$. To address this, \cite{Ruan2021} introduces the Distributional Optimal Design, formalized in the following result:

\begin{lemma}(Theorem 5, \cite{Ruan2021})
Let $\pi$ be the \textsc{Distributional Optimal Design} policy that has been learned from $s$ independent samples $\mathcal{X}_1, \ldots, \mathcal{X}_s \sim \mathcal{D}$. Let $\bm{V}$ denote the expected design matrix,
\[
\bm{V} = \E\limits_{\mathcal{X} \sim \mathcal{D}} \, \E\limits_{\bm x \sim \pi(\mathcal{X})} \left[ \bm x \bm x^\top \mid \mathcal{X} \right].
\]
Then,
\[
\mathbb{P} \left( \E\limits_{\mathcal{X} \sim \mathcal{D}} \left[ \max_{\bm x \in \mathcal{X}} \|\bm x\|_{\bm V ^{-1}} \right] \leq \mathcal{O}(\sqrt{d \log d}) \right) \geq 1 - \exp\left( \mathcal{O}(d^4 \log^2 d - s d^{-1.2} \cdot 2^{-16}) \right).
\]
\end{lemma}

We utilize the \textbf{CoreLearning for Distributional G-Optimal Design} algorithm (Algorithm 3, \cite{Ruan2021}) to learn the distributional optimal design over a given set of context vectors. In this paper, we extend both the G-Optimal and Distributional Optimal Design frameworks to the multinomial logistic (MNL) setting by introducing \emph{directionally scaled sets}. These sets are then used to construct the design policies employed in our batched algorithm.

\section{Batched Multinomial Contextual Bandit Algorithm: \texttt{B-MNL-CB}}\label{sec:B-MNL-CB}
\begin{algorithm}[ht!]
\caption{Batched Multinomial Contextual Bandit Algorithm: \texttt{B-MNL-CB}}
\label{alg : B-MNL-CB}
\begin{algorithmic}[1]
\STATE \textbf{Input:} $M$, $\bm{\rho}$, $S$, $T$
\STATE Initialize $\cbrak{\mathcal{T}_m}_{m=1}^M$ as per \ref{eqn: batch_length}, $\lambda = \sqrt{Kd\log T}$, and policy $\pi_0$ as G-OPTIMAL DESIGN
\FOR{batches $\beta \in [M]$}
    \FOR{each round $t \in \mathcal{T}_\beta$}
        \STATE Observe arm set $\mathcal{X}_t$
        \FOR{$j = 1$ to $\beta- 1$}
            \STATE Update arm set $\mathcal{X}_t\gets \textrm{UL}_{j}(\mathcal{X}_t)$ (defined in \ref{eqn: UL}) 
        \ENDFOR
        \STATE Sample $\bm{x}_t \sim \pi_{\beta-1}(\mathcal{X}_t)$ and obtain outcome $y_t$ along with the corresponding reward $\rho_{y_t}$.
    \ENDFOR
    \STATE Divide $\mathcal{T}_\beta$ into two sets $C$ and $D$ such that $\lvert C\rvert = \lvert D \rvert$, $C \cup D = \mathcal{T}_\beta$, and $C \cap D = \emptyset$.
    \STATE Compute $\hat{\bm{\theta}}_{\beta} \gets \argmin\sum\limits_{s \in C} \ell(\bm{\theta}, \bm{x}_s, y_s)$, $\bm{H}_{\beta} = \lambda \bm{I} + \sum\limits_{s \in C}  \frac{\bm{A}(\bm{x}_t,\hat{\bm{\theta}}_\beta) \otimes \bm{x}_t\bm{x}_t^\top}{B_\beta(\bm{x}_t)}$, and $\pi_\beta$ using Algorithm \ref{alg: D-Opt_MNL} with the inputs $\pbrak{\beta, \{\mathcal{X}_t\}_{t \in D}}$
\ENDFOR
\end{algorithmic}
\end{algorithm}

In this section, we present our first algorithm, \texttt{B-MNL-CB}. This section is structured in the following manner: we introduce the algorithm and explain each step in detail. This is followed by a few salient remarks and the regret guarantee for the algorithm. We provide a proof sketch for this guarantee and guide the reader to the full proof in the appendix.

\texttt{B-MNL-CB} operates in the stochastic contextual setting (described in Section \ref{section:notations}), building upon \texttt{BATCHLINUCB-DG} (Algorithm 5, \cite{Ruan2021}) and \texttt{B-GLinCB} (Algorithm 1, \cite{sawarni24}), both of which are batched algorithms. In the batched paradigm, the rounds at which the policy updates occur are fixed beforehand. We will refer to all the rounds between two consecutive policy updates as a \emph{batch}. The horizon is divided into $M = O(\log\log T)$ disjoint batches denoted by $\{\mathcal{T}_\beta\}_{\beta=1}^{M}$, and the lengths of the batches are denoted by $\tau_\beta = \modulus{\mathcal{T}_\beta}$. 

The input to \texttt{B-MNL-CB} includes the number of batches $M$, the fixed (known) reward vector $\bm{\rho}$, the known upper bound on $\twonorm{\bm{\theta}^\star}$ denoted by $S$, and the total number of rounds $T$. We denote the policy learned in each batch $\beta$ by $\pi_\beta$, initializing $\pi_0$ with the G-Optimal design. We also initialize $\lambda$ to $\sqrt{Kd \log T}$ and define the batch lengths $\cbrak{\tau_\beta}_{\beta=1}^{M}$ as follows: 
\begin{equation}
\tau_\beta = \lfloor T^{1-2^{-\beta}} \rfloor \; \forall \beta \in [1,M].
    \label{eqn: batch_length}
\end{equation}
We now provide a detailed explanation of the steps involved in the algorithm. In \emph{Steps 3-13}, we iterate over all batches $\beta\in [M]$ and rounds $t\in \mathcal{T}_{\beta}$.
During batch $\beta$ and round $t\in \mathcal{T}_{\beta}$, first, in \emph{Step 5}, we obtain the set of feasible arms $\X_t$ at round $t$. Then in \emph{Steps 6-8}, we iterate over all the previous batches $j\in [\beta-1]$ to prune $\X_t$ and retain only a subset of it via a \emph{Successive Elimination} procedure described next.

\subsection{Successive Eliminations}
\label{subsection:successive_elimination}For each prior batch $j\in [\beta-1]$, we compute an upper confidence bound $\textrm{UCB}(j, \bm{x}, \lambda)$ and a lower confidence bound $\textrm{LCB}(j, \bm{x}, \lambda)$ as follows, 

\begin{equation}
    \textrm{UCB}(j , \bm{x} , \lambda) = \bm{\rho}^T\hat{\bm{\theta}}_j + \epsilon_1(j , \bm{x} ,\lambda) + \epsilon_2(j , \bm{x} , \lambda),
    \label{eqn: UCB}
\end{equation}
\begin{equation}
    \textrm{LCB}(j , \bm{x} , \lambda) = \bm{\rho}^T\hat{\bm{\theta}}_j - \epsilon_1(j , \bm{x} ,\lambda) - \epsilon_2(j , \bm{x} , \lambda),
    \label{eqn: LCB}
\end{equation}
where the bonus terms $\epsilon_1(j , \bm{x} , \lambda)$ and $\epsilon_2(j , \bm{x} , \lambda)$ are defined as,
\begin{equation}
    \epsilon_1(j , \bm{x} , \lambda) = \gamma(\lambda)\|\bm{H}_j^{\;-\frac{1}{2}}(\bm{I} \otimes \bm{x})\bm{A}(\bm{x} , \hat{\bm{\theta}}_j)\bm{\rho}\|_2,
    \epsilon_2(j , \bm{x} , \lambda) = 3\gamma(\lambda)^2\|\bm{\rho}\|_2\|(\bm{I} \otimes \bm{x}^\top)\bm{H}_j^{\;-\frac{1}{2}}\|_2^2.
    \label{eqn: epsilon}
\end{equation}
Here, $\hat{\bm\theta}_j$ and $\bm{H}_j$ are the estimators (computed during \emph{Steps 11,12} at the end of batch $j$) of the true parameter vector $\thetastar$ and an optimal batch-level Hessian matrix $\bm{H}_j^\star$ and $\gamma(\lambda)$ is defined to be $O(\sqrt{Kd \log T})$. We provide more details on these in Section \ref{subsection:batch-hessian}. In \emph{Step 7}, for batch $j$, we eliminate a subset of $\X_t$ using the upper and lower confidence bounds just defined. In particular, we eliminate all $\bm{x} \in \X_t$ for which $\textrm{UCB}(j, \bm{x}, \lambda) \leq \max_{\bm{x}^\prime}\textrm{LCB}(j, \bm{x}^\prime, \lambda)$. Thus, in \emph{Step 7}, $\X_t$ is updated to $\textrm{UL}_j(\mathcal{X}_t)$, defined as,
\begin{equation}
    \textrm{UL}_j(\mathcal{X}) = \mathcal{X} \setminus \cbrak{\bm{x}\in\mathcal{X}: \textrm{UCB}(j , \bm{x} , \lambda) \leq \max\limits_{\bm{y}\in\mathcal{X}}\textrm{LCB}(j , \bm{y} , \lambda)},
    \label{eqn: UL}
\end{equation}
 The idea behind this step is to eliminate the arms that were not suitable candidates for the confidence regions learned in each of the previous batches. Following the successive eliminations over all prior batches $j \in [\beta - 1]$, in \emph{Step 9}, we select an arm $\bm{x}_t$ from the pruned arm set according to the policy computed at the end of batch $\beta - 1$ using Algorithm~\ref{alg: D-Opt_MNL}. The environment then returns the outcome $y_t$ and the corresponding reward $\rho_{y_t}$. Details of the policy computation (Algorithm~\ref{alg: D-Opt_MNL}) are provided in Section~\ref{subsec: policy_calc}. After completing all rounds in batch $\beta$ (i.e., $\mathcal{T}_{\beta}$), we proceed to \emph{Step 11}, where we partition these rounds equally into two sets, $C$ and $D$. The set $C$ is used to define a batch-level Hessian matrix $\bm{H}_{\beta}^\star$, compute an estimator $\hat{\theta}_{\beta}$ of $\theta^\star$, and construct a matrix $\bm{H}_{\beta}$ that estimates $\bm{H}_{\beta}^\star$ as described in the next section.
 
\subsection{Batch Level Hessian and Parameter Estimation}
\label{subsection:batch-hessian}
In batch $\beta$, we define a batch level Hessian matrix  $\bm{H}_\beta^\star = \lambda\bm{I} + \sum_{t \in C}\bm{A}(\bm{x}_t , \thetastar)\otimes\bm{x}_t\bm{x}_t^\top$, which is constructed using the set $C$.  Since $\thetastar$ is unknown, we maintain an online proxy to estimate $\bm{H}_{\beta}^{\star}$ by calculating a scaled Hessian matrix $\bm{H}_\beta = \lambda \bm{I} + \sum_{t \in C}\frac{\bm{A}(\bm{x}_t , \hat{\bm{\theta}}_\beta)}{B_\beta(\bm{x})} \otimes\bm{x}_t\bm{x}_t^\top$. 
Here, $B_\beta(\bm{x})$ is a normalizing factor which is obtained using the self-concordance properties of the link function and is given by:
\begin{equation}
    B_\beta(\bm{x}) = \exp\pbrak{\sqrt 6\min\cbrak{\gamma(\lambda)\sqrt{\kappa}\matnorm{\bm{x}}{\bm{V}_\beta\inv} , 2S}} 
    \label{eqn: self-concordance normalization},
\end{equation}
where $\gamma(\lambda) = \O(\sqrt{Kd\log T})$ is the confidence radius for the permissible set of $\bm{\theta}$ and $\bm{V}_\beta$ is the design matrix given by $\bm{V}_\beta = \lambda\bm{I} + \sum_{t\in C}\bm{x}_t\bm{x}_t^\top$. Using the self-concordance properties of the link function, we can show that $\bm{H}_\beta \mleq \bm{H}^\star_\beta$. The set  $C$ is also used to update the estimator $\hat{\bm{\theta}}_\beta$, which is done by minimizing the negative log likelihood $\sum\limits_{t\in C} \ell(\bm\theta, \bm{x}_t, y_t)$, where $\ell(\bm\theta, \bm{x}, y)$ is defined as,
\begin{equation}
    \ell(\bm{\theta} ,\bm{x} , y) = -\sum\limits_{i=1}^{K}\mathbbm{1}\cbrak{y = i}\log\frac{1}{z_i(\bm{x} , \bm{\theta)}} + \frac{\lambda}{2}\|\bm{\theta}\|_2^2,
    \label{eqn: loss}
\end{equation}
Next, we explain how the policy is updated to $\pi_{\beta}$ at the end of batch $\beta$ using the rounds in set $D$.

\subsection{Policy calculation}
\label{subsec: policy_calc}
\begin{algorithm}[!ht]
\caption{Distributional Optimal Design for MNL bandits}
\label{alg: D-Opt_MNL}
\begin{algorithmic}[1]
\STATE \textbf{Input} Batch $\beta$ and collection of arm sets $\{\mathcal{X}_j\}_j$ 
\STATE Create the sets $\{F_i(\{\mathcal{X}_j\}_j , \beta)\}_{i =1}^K$ as defined in Equation \ref{equation: scaled sets}.
\STATE Compute the distributional optimal design policy $\pi_i$ for each of the sets $F_i(\{\mathcal{X}_j \}_j, \beta)$.
\STATE Compute the distributional optimal design policy $\pi_0$ for the set $\{\mathcal{X}_j\}_j$.
\STATE  \textbf{Return} $\pi = \frac{1}{K+1}\sum\limits_{i=0}^{K}\pi_i$
\end{algorithmic}
\end{algorithm}

To compute our final policy at the end of each batch, we utilize the idea of distributional optimal design, first introduced in \cite{Ruan2021} (See Section \ref{section:notations}). Recently, \cite{sawarni24} used distributional optimal designs to develop limited adaptivity algorithms for stochastic contextual bandits for generalized linear bandits. A key step in their algorithm (Step $13$ and Equation $4$, Algorithm $1$ in \cite{sawarni24}) involves scaling the arm set (after pruning using \emph{successive eliminations}) with the derivative of the link function and a suitable normalization factor. 
Generalizing this idea to the MNL setting results in a matrix $\tilde{\bm{X}} = \frac{\bm{A}(\bm{x} , \hat{\bm{\theta}}_t)^\frac{1}{2}}{B_\beta(\bm{x})}\otimes \bm{x}$. Since optimal design concepts apply only to vectors, the technique used in \cite{sawarni24} can not be trivially extended to the MNL case. Hence, to combat this problem and to use distributional optimal designs in the MNL bandit case, we simulate the idea of designs on matrices by introducing the concept of directionally scaled sets ( Algorithm \ref{alg: D-Opt_MNL}). Through this, we create a set of $K$ scaled sets, learn the distributional optimal designs of each of these sets individually, and then combine them to create the final distribution. We now proceed with the description of the algorithm.


In \emph{Step 12} of Algorithm \ref{alg : B-MNL-CB}, we invoke this algorithm (Algorithm \ref{alg: D-Opt_MNL}) with inputs as the batch number $\beta$ and the collection of all the pruned arm sets $\{\X_t\}_{t \in D}$ (\emph{Step 7}, Algorithm \ref{alg : B-MNL-CB}). We then create $K$ different sets $F_i(\{\mathcal{X}_t\}_{t \in D} , \beta)$ ($i\in [K]$), which comprises of the arms in each arm set scaled by the $i^{\text{th}}$ column of the gradient matrix. In particular, 
\begin{equation}
    F_i(\{\mathcal{X}_t\}_{t \in D}, \beta) = \left\{ \left\{\frac{\bm{A}(\bm{x} ,\hat{\bm{\theta}}_\beta)^{\frac{1}{2}}}{\sqrt{B_\beta(\bm{x})}} \bm{e}_i\otimes \bm{x} : \bm{x} \in \mathcal{X}_t\right\}: t \in D \right\},
    \label{equation: scaled sets}
\end{equation}
where $\bm{e}_i \in \mathbb{R}^{K}$ is the $i^{\text{th}}$ standard basis vector.  We calculate the distributional optimal design for each of the sets $F_i(\{\mathcal{X}_t\}_{t \in D} , \beta)$ using Algorithm 2 in \cite{Ruan2021}. In such a case, it is easy to see that calculating the distributional optimal design over $\tilde{\bm{X}}$ can be done by calculating the distributional optimal designs for each of the sets $F_i(\{\mathcal{X}_t\}_{t \in D} , \beta)$. We provide the proof for the same in Section \ref{appendix: optimal design}. We also calculate the distributional optimal design over $\{\mathcal{X}_t\}_{t \in D}$. Finally, the policy returned is a convex combination (in this case, a uniform combination) over all the $K+1$ designs that were calculated. 

This completes our explanation of Algorithm \ref{alg : B-MNL-CB}. We provide a regret guarantee in Theorem \ref{thm: regret_b_mnl}.

\begin{remark}
    A direct application of the scaling techniques introduced in \cite{sawarni24} for learning distributional optimal designs in the multinomial setting results in the creation of a scaled matrix. Since the notion of distributional optimal design introduced in \cite{Ruan2021} applies only to vectors, Algorithm \ref{alg: D-Opt_MNL} scales the original context vectors into $K$ different sets and then learns the optimal designs for each of them.
     \label{Remark 1}
\end{remark}

\begin{remark}
    \cite{sawarni24} introduces a warm-up round with length $O(\kappa^{1/3})$. Since $\kappa$ can scale exponentially with several instance-dependent parameters, the warm-up round can result in a long exploration phase. Using the regret decomposition in \cite{Zhang2023}, we can eliminate the dependence on $\kappa$, resulting in $\kappa-$free batch lengths, including the length of the warm-up round.
     \label{Remark 2}
\end{remark}

\begin{remark}
    While \cite{Zhang2023} introduced a novel method of regret decomposition into the error terms (refer \ref{eqn: epsilon}), a straightforward application to the limited adaptivity setting is not easy. Hence,  with some additional insights, we incorporate their method into the batched setting while being able to match the leading term of their regret bound. 
    \label{Remark 3}
\end{remark}

\begin{theorem}
    (Regret of \texttt{B-MNL-CB}) With high probability, at the end of $T$ rounds, the regret incurred by Algorithm \ref{alg : B-MNL-CB} is bounded as $R_T \leq R_1 + R_2$ where 
    \[
        R_1 = \tilde{O}\left( R S^{5/4} K^{5/2} d\sqrt{T} \right)
        \text{ and }
        R_2 = \tilde{O}\left(R S^{5/2} K^2 d^2 \kappa^{1/2} T^{1/4} \max\{e^{3S} K^{3/2} S^{-1}, \kappa^{1/2} d\} \right).
    \]
    \label{thm: regret_b_mnl}
\end{theorem}

\textbf{Proof Sketch:}

We know that the expected regret during batch $\beta+1$ is given by:
\[R_{\beta+1} = \E\sbrak{\sum\limits_{t\in\beta} \bm{\rho}^\top \bm{z}(\bm{x}_t^\star , \bm{\theta}^\star) - \bm{\rho}^\top \bm{z}(\bm{x}_t , \bm{\theta}^\star)},\]

where $\bm{x}_t^\star = \argmax\limits_{\bm{x}\in\mathcal{X}_t}\bm{\rho}^\top\bm{z}(\bm{x}, \bm{\theta}^\star)$ is the best arm at round $t$ and the expectation is taken over the distribution of the arm set $\mathcal{D}$. Using ideas similar to \cite{Zhang2023}, we can decompose the regret into 
\[R(T) \leq 4\sum\limits_{t\in\beta}\cbrak{\E\sbrak{\max\limits_{\bm{x}\in\mathcal{X}_t}\epsilon_1(\beta , \bm{x} , \lambda)} + \E\sbrak{\max\limits_{\bm{x}\in\mathcal{X}_t}\epsilon_2(\beta , \bm{x} , \lambda)} },\]
where $\epsilon_1(\beta , \bm{x} , \lambda)$ and $\epsilon_2(\beta , \bm{x} , \lambda)$ are as defined in \ref{eqn: epsilon}. We proceed to bound each of these terms using the extension of distributional optimal design we introduced in Algorithm \ref{alg: D-Opt_MNL}.

Directly extending the ideas of \cite{Ruan2021} and \cite{sawarni24} to construct the distributional optimal designs results in an attempt to learn the design for matrices $\tilde{\bm{X}}_\beta = \frac{\bm{A}(\bm{x} , \hat{\bm{\theta}_\beta})^\frac{1}{2}}{B_\beta(\bm{x})}\otimes \bm{x}$. Hence, we create $K$ different sets $F_i(\mathcal{X}) \text{ for all } i \in [K]$ (defined in \ref{equation: scaled sets}), such that $$\tilde{\bm{X}}_\beta\tilde{\bm{X}}_\beta^\top = \sum\limits_{i=1}^{K} \cbrak{\frac{\bm{A}(\bm{x} , \hat{\bm{\theta}}_\beta)^{\frac{1}{2}}}{\sqrt{B_\beta(\bm{x})}} \bm{e}_i\otimes \bm{x}}\cbrak{\frac{\bm{A}(\bm{x} , \hat{\bm{\theta}}_\beta)^{\frac{1}{2}}}{\sqrt{B_\beta(\bm{x})}} \bm{e}_i\otimes \bm{x}}^T .$$ 
Thus, learning the optimal design over $\tilde{\bm{X}}$ is equivalent to creating a convex combination of the designs learned over $F_i(\mathcal{X}) \text{ for all } i \in [K]$. 
This gives us a way of bounding the scaled Hessian matrix $\bm{H}_\beta$ by the scaled Hessian matrices $\bm{H}^i_\beta$ constructed over $F_i(\mathcal{X})$ for all $i \in [K]$. We then use methods similar to \cite{sawarni24} and \cite{Ruan2021} to obtain the bound on the regret for the batch $\beta+1$ as:
\begin{align*}R_{\beta+1} &\leq + 32 R K \kappa^{1/2} d\gamma^2(\lambda) \cbrak{e^{3S} K^{3/2} S^{-1} \sqrt{\log (Kd) \log d} + 12 \kappa^{1/2} d} \pbrak{\frac{\tau_{\beta+1}}{\tau_\beta}} 
\\
&+ 16 R K^2 \gamma(\lambda) \sqrt{d \log (Kd)}\pbrak{\frac{\tau_{\beta+1}}{\sqrt{\tau_\beta}}} 
\end{align*}
Finally, using the batch lengths defined in \ref{eqn: batch_length} and summing over all the $M$ batches completes the proof. For the sake of brevity, we provide the complete proof in Section \ref{section: regret_b_mnl}.

\section{Rarely Switching Multinomial Contextual Bandit Algorithm: \texttt{RS-MNL}}\label{sec:RS-MNL-UCB}

\begin{algorithm}[!ht]
\caption{RS-MNL}
\label{alg : RS-MNL}
\begin{algorithmic}[1]
\STATE \textbf{Inputs: } $\bm{\rho} , S , T$ 
\STATE \textbf{Initialize:} $\bm{H}_1 = \lambda \bm{I}$, $\tau = 1$, $\lambda := K d S^{-1/2} \log(T/\delta)$, $\gamma := C S^{5/4} \sqrt{Kd \log (T/\delta)}$
\FOR{$t = 1, \ldots, T$}
    \STATE Observe arm set $\mathcal{X}_t$
    \IF{$\det(\bm{H}_t) > 2 \det(\bm{H}_\tau)$}
        \STATE Set $\tau = t$ 
        \STATE Update $\hat{\bm{\theta}}_\tau \gets \argmin\limits_{\bm{\theta}}\sum\limits_{s \in [t-1]}\ell(\bm\theta, \bm{x}_s, y_s)$ and $\bm{H}_t = \sum\limits_{s \in [t-1]}\frac{\bm{A}(\bm{x}_s,\hat{\bm{\theta}}\tau)}{B_\tau(\bm{x}_s)}\otimes \bm{x}_s\bm{x}_s^\top + \lambda \bm{I}_{Kd}$
    \ENDIF
    \STATE Select $\bm{x}_t = \argmax\limits_{\bm{x} \in \mathcal{X}_t} \textrm{UCB}(t , \tau , \bm{x})$, observe $y_t$, and update $\bm{H}_{t+1} \gets \bm{H}_t + \frac{\bm{A}(\bm{x}_t,\hat{\bm{\theta}}_\tau)}{B_\tau(\bm{x}_t)}\otimes\bm{x}_t\bm{x}_t^\top$
\ENDFOR
\end{algorithmic}
\end{algorithm}

In this section, we present our second algorithm \texttt{RS-MNL}. We introduce the algorithm and explain the workings in a step-by-step fashion. We then mention a few salient remarks about our algorithm. We conclude with the regret guarantee of our algorithm, a proof sketch for the same, and guide the reader to the complete proof in the Appendix.

Our second algorithm, \texttt{RS-MNL} (Algorithm \ref{alg : RS-MNL}) operates in the Adversarial Contextual setting. In this setting, there are no assumptions on the generation of the feature vectors. \texttt{RS-MNL} also limits the number of policy updates in a rarely-switching fashion, i.e, the rounds where these updates are made are decided dynamically, based on a simple switching criterion, similar to the one used in \cite{AbbasiYadkori2011}. While the algorithm is based on \texttt{RS-GLinCB} in \cite{sawarni24}, a unique regret decomposition method allows for the removal of the warmup criterion, in turn, helping in the reduction in the number of switches made by the algorithm from $O(\log^2 T)$ to $O(\log T)$. Further, we successfully remove the idea of \emph{successive eliminations} based on the previous confidence regions and replace the idea with the maximization of the Upper Confidence Bound (UCB) of each arm. 

The inputs to the algorithm are $\bm{\rho}$, the fixed and known reward vector, $S$, the fixed and known upper bound on $\|\bm{\theta}\|_2$, and $T$, the number of rounds for which the algorithm is played. In \emph{Step 2}, we initialize the scaled Hessian matrix $\bm{H}_1$ to $\bm{I}$, $\lambda$ to $K d S^{-1/2} \log(T/\delta)$, and $\gamma$ to $C S^{5/4} \sqrt{Kd \log (T/\delta)}$. Next, at every round $t\in [T]$, we receive the arm set $\mathcal{X}_t$ in \emph{Step 4}. During \emph{Steps 5-8}, we check if the switching condition is met and update the policy accordingly. 

\subsection{Switching Criterion and Policy Update:} We use $\tau \leq t$ to denote the last time round at which a switch occurred for some round $t$. In \emph{Step 5}, we check for the switching condition: if the determinant of the scaled Hessian matrix $\bm{H}_t = \lambda\bm{I} + \sum_{s\in[t-1]}\frac{\bm{A}(\bm{x}_s , \bm{\hat\theta_\tau})}{B(\bm{x}_s)}\otimes\bm{x}_s\bm{x}_s^\top$ has increased by a constant factor (in this case, 2) as compared to $\bm{H}_\tau$. In case the switching condition is triggered, we set $\tau = t$ in \emph{Step 6} (since a switch was made in round $t$). We then compute $\hat{\bm{\theta}}_\tau$ by minimizing the negative log likelihood $\sum_{s\in [t-1]}\ell(\bm\theta, \bm{x}_s, y_s)$ (see \ref{eqn: loss} for definition of $\ell(\bm{\theta}, \bm{x}_s, y_s)$)  over all previous rounds $s\in [t-1]$, and recompute the matrix $\bm{H}_t$ with respect to the newly calculated $\hat{\bm{\theta}}_\tau$ (\emph{Step 7}). The switching criterion is similar to the one used in \cite{AbbasiYadkori2011} and helps to reduce the number of policy updates to $O(\log{T})$.

\subsection{Arm Selection:}
Next, in \emph{Step 9}, we determine the arm $\bm{x}_t$ to be played based on the Upper Confidence Bound (UCB). The upper confidence bound $\textrm{UCB}(t , \tau , \bm{x})$ for an arm $\bm{x}\in \X_t$ with respect to the previous switching round $\tau (\leq t)$ is defined as:
\begin{equation}
    \textrm{UCB}(t , \tau , \bm{x}) = \bm{\rho}^T\hat{\bm{\theta}}_\tau + \epsilon_1(t , \tau , \bm{x}) + \epsilon_2(t , \tau , \bm{x}),
\end{equation}
where the error terms $\epsilon_1(t , \tau , \bm{x})$ and $\epsilon_2(t , \tau , \bm{x})$ are defined as:
    \begin{equation}
    \epsilon_1(t , \tau , \bm{x}) = \sqrt{2}\gamma(\delta)\|\bm{H}^{-\frac{1}{2}}_{t}(\bm{I} \otimes \bm{x})\bm{A}(\bm{x} , \hat{\bm{\theta}}_\tau)\bm{\rho}\|_2, \hspace{0.5em}
        \epsilon_2(t  ,\tau , \bm{x}) = 6R\gamma(\delta)^2\|(\bm{I} \otimes \bm{x}^\top)\bm{H}^{-\frac{1}{2}}_{t}\|_2^2.
        \label{eqn: RS_epsilon}
    \end{equation}
We then obtain the outcome $y_t$, which is sampled from $\bm{z}(\bm{x}_t , \bm{\theta}^\star)$, and receives the corresponding reward $\rho_{y_t}$. The algorithm then updates the scaled Hessian matrix $\bm{H}_{t+1}$. In Theorem \ref{thm: regret_rs_mnl}, we provide the regret guarantee for \texttt{RS-MNL}.

\begin{remark}
The goal of a rarely-switching algorithm is to reduce the number of switches (policy updates) that are made. Our algorithm successfully reduces the number of switches from $\O(\log^2 T)$ to $O(\log T)$ due to the removal of the warm-up switching criterion. Additionally, the number of switches is independent of $\kappa$.
\end{remark}

\begin{remark}
Similar to the batched setting, using the regret decomposition method introduced in \cite{Zhang2023} in the rarely-switching paradigm is non-trivial. We manage to extend their results to match the leading term of their regret bound while performing a switch $O(\log T)$ times.
\end{remark}

\begin{theorem}
    With high probability, after $T$ rounds, Algorithm \ref{alg : RS-MNL} achieves the following regret:

    \begin{align*}
    R_T &\leq \tilde{O}\left( R K^{3/2} S^{5/4} d \sqrt{T} \right).
    \end{align*}
    \label{thm: regret_rs_mnl}
\end{theorem}

\textbf{Proof Sketch: }
 
The expression for total regret is given by 
\[R(T) = \sum\limits_{t=1}^T \bm{\rho}^\top \bm{z}(\bm{x}_t^\star , \bm{\theta}^\star) - \bm{\rho}^\top\bm{z}(\bm{x}_t , \bm{\theta}^\star),\]
where $\bm{x}_t^\star = \argmax\limits_{\bm{x}\in\mathcal{X}_t}\bm{\rho}^\top\bm{z}(\bm{x} , \bm{\theta}^\star)$ is the best arm at any given round $t$. Using a method similar to the one used in \cite{Zhang2023}, we can upper bound the regret as 
\[R(T) \leq 2\sum\limits_{t=1}^T\cbrak{\epsilon_1(t,\tau,\bm{x}_t) + \epsilon_2(t , \tau , \bm{x}_t)}, \]
where $\epsilon_1(t , \tau , \bm{x}_t)$ and $\epsilon_2(t , \tau , \bm{x}_t)$ are as defined in \ref{eqn: RS_epsilon}. We now wish to upper bound both the terms separately. 

Bounding $\epsilon_1(t , \tau , \bm{x}_t)$ using the switching criterion in \cite{AbbasiYadkori2011} along with the selection rule in our algorithm can result in an exponential dependency in $S$. \cite{sawarni24} was able to circumvent this exponential dependency by using an additional switching criterion, referred to as a \emph{warmup criterion}. However, this results in the increase in the number of switches from $O(\log T)$ to $O(\log^2 T)$. It also slows down the algorithm due to the successive eliminations done at each round (similar to the ones in Algorithm \ref{alg : B-MNL-CB}). Our algorithm gets rid of the exponential dependency from the first order term and the warm-up criterion by decomposing $\epsilon_1(t , \tau , \bm{x}_t)$ in an alternate manner, resulting in an improved runtime as well as $O(\log T)$ switches.

We bound both  $\epsilon_1(t , \tau , \bm{x}_t)$ and $\epsilon_2(t , \tau , \bm{x}_t)$ using an analysis similar to the one used for Theorem \ref{thm: regret_b_mnl}, where we attempt to upper bound the scaled Hessian matrix $\bm{H}_t$ using the scaled Hessian matrices calculated over the $K$ different scaled sets introduced in \ref{alg: D-Opt_MNL}. Note that nowhere do these $K$ different sets appear in the algorithm. They only serve to ease the analysis.
Combining the bounds on each of the error terms finishes the proof. For the sake of brevity, we provide the complete proof in Section \ref{section: regret_rs_mnl} of the Appendix.

\section{Experiments}
\label{sec: Experiments}
In this section, we compare our algorithm \texttt{RS-MNL} to several contextual logistic and MNL bandit algorithms\footnote{The code for the experiments can be found \href{https://github.com/tanmaygoyal258/MNL_Limited_Adaptivity.git}{here}.}. We describe the experiments in detail below:
\begin{figure*}
	\centering
	\begin{subfigure}[b]{0.32\columnwidth}  
		\centering 
		\includegraphics[width=45mm]{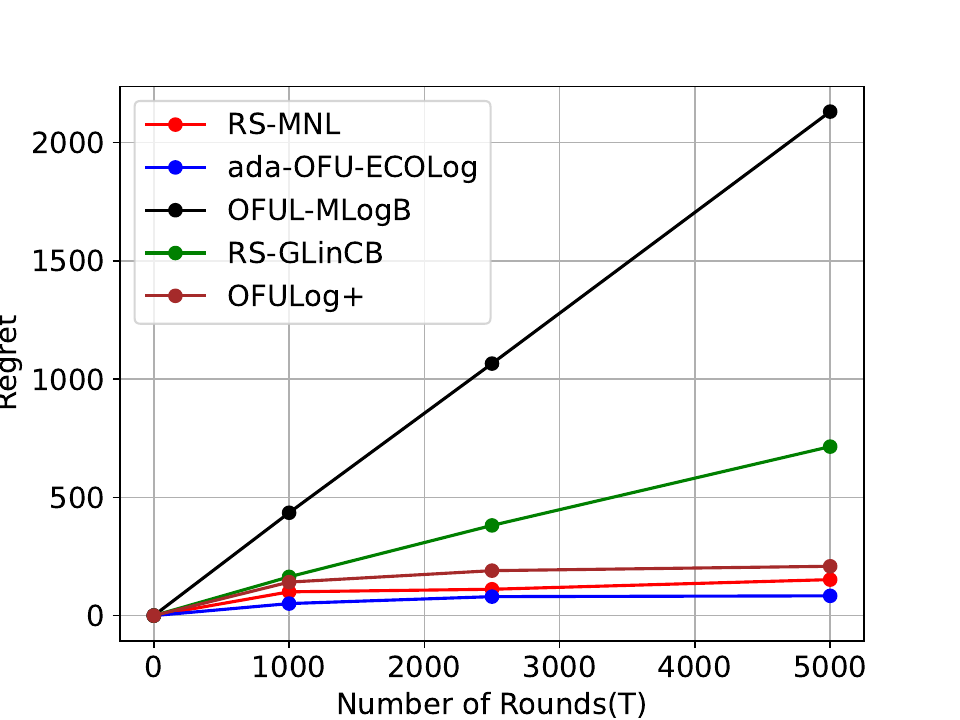}
		\caption[]{{\small Regret vs.\ $T$: Logistic Setting}}   
		\label{fig:logistic}
	\end{subfigure}
	\hfill
	\begin{subfigure}[b]{0.32\columnwidth}   
		\centering 
	\includegraphics[width=45mm]{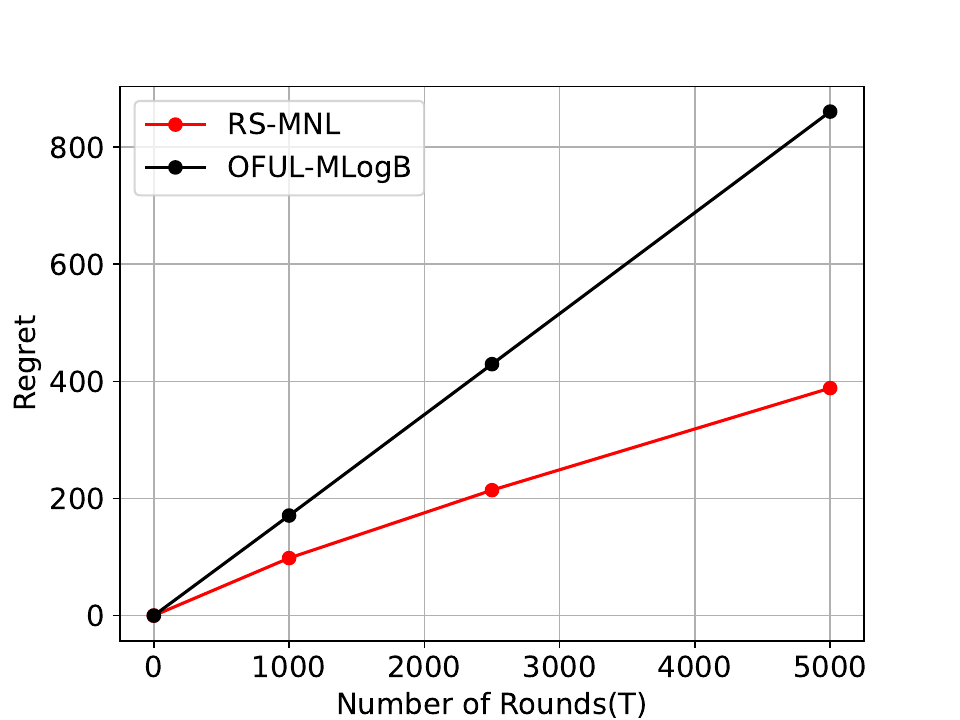}
		\caption[]{{\small Regret vs.\ $T$: $K = 3$}}   
		\label{fig:3 outcomes}
	\end{subfigure}
    \hfill
    \begin{subfigure}[b]{0.32\columnwidth}
    \centering
    \includegraphics[width=45mm]{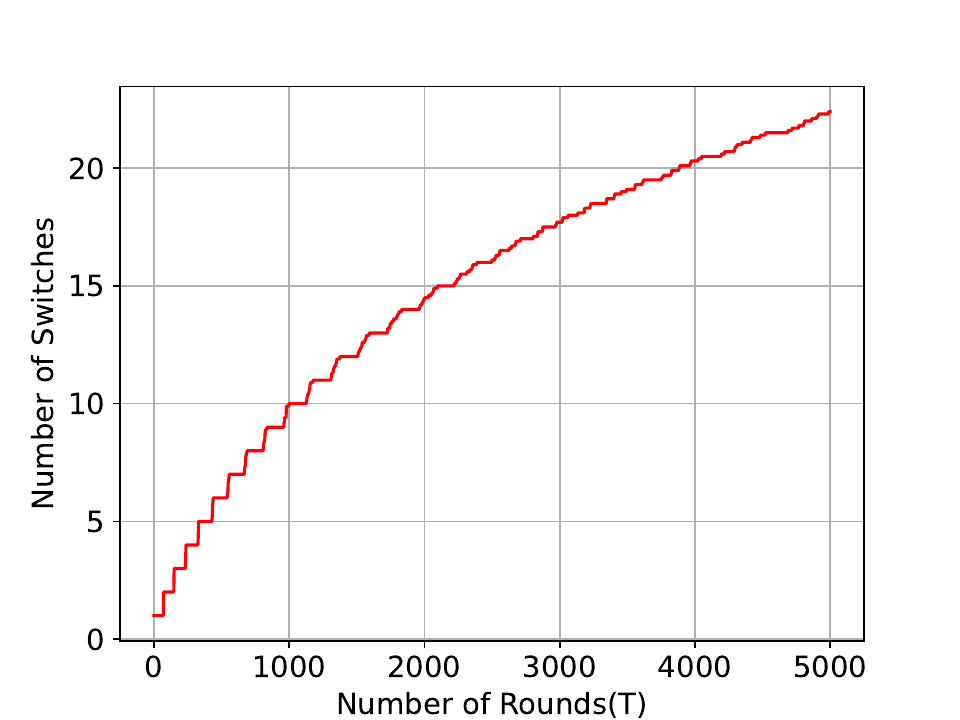}
    \caption[]{{\small Switches vs.\ T}}  \label{fig:switches_vs_T}
            
    \end{subfigure} 
\end{figure*}

\textbf{Experiment 1 ($R(T)$ vs.\ $T$ for the Logistic $(K=1)$ Setting):}
In this experiment, we compare our algorithm \texttt{RS-MNL} to several state-of-the-art contextual logistic bandit algorithms such as \texttt{ada-OFU-ECOLog} (Algorithm 2, \cite{Faury2022}), \texttt{RS-GLinCB} (Algorithm 2, \cite{sawarni24}, \texttt{OFUL-MLogB} (Algorithm 2, \cite{Zhang2023}), and \texttt{OFULog+} (Algorithm 1, \cite{lee2024}). The dimension of the arms $d$ is set to $3$ and the number of outcomes $K$ is set to $1$, which reduces the problem to the logistic setting. The arm set $\mathcal{X}$ is constructed by sampling 10 different arms from $[-1,1]^3$ and normalizing them to unit vectors. The optimal parameter $\thetastar$ is chosen randomly from $[-1,1]^3$ and normalized so that $\lVert \bm\theta^\star \rVert = S = 2$. We run all the algorithms for $T \in \{1000,2500,4500\}$ rounds and average the results over 10 different seeds (for sampling rewards). The results are plotted in Figure \ref{fig:logistic}. We see that \texttt{RS-MNL} is incredibly competitive with \texttt{ada-OFU-ECOLog} and \texttt{OFULog+}, while incurring much lower regret than \texttt{RS-GLinCB} and \texttt{OFUL-MLogB}. We showcase the results with two standard deviations in Section \ref{appendix: Additional Experiments}.

\textbf{Experiment 2 ($R(T)$ vs.\ $T$ for $K =3$):}  
In this experiment, we compare our algorithm \texttt{RS-MNL} to \texttt{OFUL-MLogB}, the only algorithm that achieves an optimal ($\kappa-$free) regret while being computationally efficient for MNL Bandits (to the best of our knowledge). We set the number of outcomes $K$ as $3$ and the dimension of the arms $d$ is set to $3$. The arm set $\mathcal{X}$ is constructed by sampling 10 different arms from $[-1,1]^3$ and normalizing them to unit vectors. The optimal parameter $\thetastar$ is sampled from $[-1,1]^{9}$ (since $\thetastar \in \R^{Kd}$ and normalized so that $\lVert \bm\theta^\star \rVert = S = 2$. The reward vector $\bm\rho$ is sampled from $[0,1]^3$ and normalized so that $\lVert \rho \rVert = R = 2$. We run both the algorithms for $T \in \{1000,2500,4500\}$ rounds and average these results over 10 different seeds (for sampling rewards and $\bm\rho$). The results are plotted in \ref{fig:3 outcomes}. We see that \texttt{RS-MNL} incurs much lower regret than \texttt{OFUL-MLogB}. We also showcase the results with two standard deviations in Section \ref{appendix: Additional Experiments}.

\textbf{Experiment 3 (Number of Switches vs.\  T): } In this experiment, we plot the number of switches \texttt{RS-MNL} makes as a function of the number of rounds $T$. We assume that the instance is simulated in the same manner as \textbf{Experiment 2}. We run the algorithm for $T = 5000$  rounds and average over 10 different seeds. The results are shown in Figure \ref{fig:switches_vs_T}. We see that the number of switches made by \texttt{RS-MNL} exhibits a strong logarithmic dependence with $t \in [T]$. This is in agreement with Lemma \ref{lemma: number_switches}, where we show that \texttt{RS-MNL} switches $O(\log t)$ times, as compared to other algorithms, which switch (update) $O(t)$ times.

\section{Conclusions and Future Work}
In this paper, we present two algorithms \texttt{B-MNL-CB} and \texttt{RS-MNL}, for the multinomial logistic setting in the batched and rarely-switching paradigms, respectively. The batched setting involves fixing the policy update rounds at the start of the algorithm, while the rarely switching setting chooses the policy update rounds adaptively. Our first algorithm, \texttt{B-MNL-CB} manages to extend the notion of distributional optimal designs to the multinomial logit setting while being able to achieve an optimal regret of $\tilde{O}(\sqrt T)$ in $\Omega(\log\log T)$ batches. Our second algorithm, \texttt{RS-MNL}, builds upon the rarely-switching algorithm presented in \cite{sawarni24} and obtains an optimal regret of $\tilde{O}(\sqrt T)$ while being able to reduce the number of switches to $O(\log T)$ using alternate ways of regret decomposition. The regret of our algorithms scales with the number of outcomes $K$ as $K^{7/2}$ and $K^{5/2}$ respectively, which can be detrimental for problems with a large number of outcomes. We believe that this dependence on $K$ can be further improved, which is an interesting line for future work.

\bibliography{main}

\begin{thebibliography}{18}
\providecommand{\natexlab}[1]{#1}
\providecommand{\url}[1]{\texttt{#1}}
\expandafter\ifx\csname urlstyle\endcsname\relax
  \providecommand{\doi}[1]{doi: #1}\else
  \providecommand{\doi}{doi: \begingroup \urlstyle{rm}\Url}\fi

\bibitem[Abbasi-Yadkori et~al.(2011)Abbasi-Yadkori, Pál, and Szepesvári]{AbbasiYadkori2011}
Yasin Abbasi-Yadkori, Dávid Pál, and Csaba Szepesvári.
\newblock Improved algorithms for linear stochastic bandits.
\newblock In \emph{Advances in Neural Information Processing Systems 24 (NeurIPS)}, pages 2312--2320, 2011.

\bibitem[Abeille et~al.(2021)Abeille, Faury, and Calauzenes]{Abeille2021}
Marc Abeille, Louis Faury, and Clement Calauzenes.
\newblock Instance-wise minimax-optimal algorithms for logistic bandits.
\newblock In Arindam Banerjee and Kenji Fukumizu, editors, \emph{Proceedings of The 24th International Conference on Artificial Intelligence and Statistics}, volume 130 of \emph{Proceedings of Machine Learning Research}, pages 3691--3699. PMLR, 13--15 Apr 2021.
\newblock URL \url{https://proceedings.mlr.press/v130/abeille21a.html}.

\bibitem[Amani and Thrampoulidis(2021)]{Amani2021}
Sanae Amani and Christos Thrampoulidis.
\newblock {UCB}-based algorithms for multinomial logistic regression bandits.
\newblock In A.~Beygelzimer, Y.~Dauphin, P.~Liang, and J.~Wortman Vaughan, editors, \emph{Advances in Neural Information Processing Systems}, 2021.
\newblock URL \url{https://openreview.net/forum?id=Jhp38rtUTV}.

\bibitem[Auer(2003)]{auer2003}
Peter Auer.
\newblock Using confidence bounds for exploitation-exploration trade-offs.
\newblock \emph{J. Mach. Learn. Res.}, 3\penalty0 (null):\penalty0 397–422, March 2003.
\newblock ISSN 1532-4435.

\bibitem[Chu et~al.(2011)Chu, Li, Reyzin, and Schapire]{chu2011}
Wei Chu, Lihong Li, Lev Reyzin, and Robert Schapire.
\newblock Contextual bandits with linear payoff functions.
\newblock In Geoffrey Gordon, David Dunson, and Miroslav Dudík, editors, \emph{Proceedings of the Fourteenth International Conference on Artificial Intelligence and Statistics}, volume~15 of \emph{Proceedings of Machine Learning Research}, pages 208--214, Fort Lauderdale, FL, USA, 11--13 Apr 2011. PMLR.
\newblock URL \url{https://proceedings.mlr.press/v15/chu11a.html}.

\bibitem[Faury et~al.(2020)Faury, Abeille, Calauzenes, and Fercoq]{Faury2020}
Louis Faury, Marc Abeille, Clement Calauzenes, and Olivier Fercoq.
\newblock Improved optimistic algorithms for logistic bandits.
\newblock In Hal~Daumé III and Aarti Singh, editors, \emph{Proceedings of the 37th International Conference on Machine Learning}, volume 119 of \emph{Proceedings of Machine Learning Research}, pages 3052--3060. PMLR, 13--18 Jul 2020.
\newblock URL \url{https://proceedings.mlr.press/v119/faury20a.html}.

\bibitem[Faury et~al.(2022)Faury, Abeille, Jun, and Calauzenes]{Faury2022}
Louis Faury, Marc Abeille, Kwang-Sung Jun, and Clement Calauzenes.
\newblock Jointly efficient and optimal algorithms for logistic bandits.
\newblock In Gustau Camps-Valls, Francisco J.~R. Ruiz, and Isabel Valera, editors, \emph{Proceedings of The 25th International Conference on Artificial Intelligence and Statistics}, volume 151 of \emph{Proceedings of Machine Learning Research}, pages 546--580. PMLR, 28--30 Mar 2022.
\newblock URL \url{https://proceedings.mlr.press/v151/faury22a.html}.

\bibitem[Filippi et~al.(2010)Filippi, Cappe, Garivier, and Szepesv\'{a}ri]{Filippi2010}
Sarah Filippi, Olivier Cappe, Aur\'{e}lien Garivier, and Csaba Szepesv\'{a}ri.
\newblock Parametric bandits: The generalized linear case.
\newblock In J.~Lafferty, C.~Williams, J.~Shawe-Taylor, R.~Zemel, and A.~Culotta, editors, \emph{Advances in Neural Information Processing Systems}, volume~23. Curran Associates, Inc., 2010.
\newblock URL \url{https://proceedings.neurips.cc/paper_files/paper/2010/file/c2626d850c80ea07e7511bbae4c76f4b-Paper.pdf}.

\bibitem[Gao et~al.(2019)Gao, Han, Ren, and Zhou]{Gao2019}
Zijun Gao, Yanjun Han, Zhimei Ren, and Zhengqing Zhou.
\newblock Batched multi-armed bandits problem.
\newblock In H.~Wallach, H.~Larochelle, A.~Beygelzimer, F.~d'Alché Buc, E.~Fox, and R.~Garnett, editors, \emph{Advances in Neural Information Processing Systems}, volume~32, pages 503--513. Curran Associates, Inc., 2019.

\bibitem[Group et~al.(1997)]{international1997international}
International Stroke Trial~Collaborative Group et~al.
\newblock The international stroke trial (ist): a randomised trial of aspirin, subcutaneous heparin, both, or neither among 19 435 patients with acute ischaemic stroke.
\newblock \emph{The Lancet}, 349\penalty0 (9065):\penalty0 1569--1581, 1997.

\bibitem[Hanna et~al.(2023)Hanna, Yang, and Fragouli]{Hanna2023}
Osama~A. Hanna, Lin~F. Yang, and Christina Fragouli.
\newblock Efficient batched algorithm for contextual linear bandits with large action space via soft elimination.
\newblock In \emph{Proceedings of the 37th International Conference on Neural Information Processing Systems}, NIPS '23, Red Hook, NY, USA, 2023. Curran Associates Inc.

\bibitem[Kiefer and Wolfowitz(1960)]{Kiefer_Wolfowitz_1960}
J.~Kiefer and J.~Wolfowitz.
\newblock The equivalence of two extremum problems.
\newblock \emph{Canadian Journal of Mathematics}, 12:\penalty0 363–366, 1960.
\newblock \doi{10.4153/CJM-1960-030-4}.

\bibitem[Lattimore and Szepesvári(2020)]{Lattimore_Szepesvári_2020}
Tor Lattimore and Csaba Szepesvári.
\newblock \emph{Bandit Algorithms}.
\newblock Cambridge University Press, 2020.

\bibitem[Lee et~al.(2024)Lee, Yun, and Jun]{lee2024}
Junghyun Lee, Se-Young Yun, and Kwang-Sung Jun.
\newblock Improved regret bounds of (multinomial) logistic bandits via regret-to-confidence-set conversion.
\newblock In Sanjoy Dasgupta, Stephan Mandt, and Yingzhen Li, editors, \emph{Proceedings of The 27th International Conference on Artificial Intelligence and Statistics}, volume 238 of \emph{Proceedings of Machine Learning Research}, pages 4474--4482. PMLR, 02--04 May 2024.
\newblock URL \url{https://proceedings.mlr.press/v238/lee24d.html}.

\bibitem[Li et~al.(2017)Li, Lu, and Zhou]{Li2017}
L.~Li, Y.~Lu, and D.~Zhou.
\newblock Provably optimal algorithms for generalized linear contextual bandits.
\newblock In \emph{Proceedings of the 34th International Conference on Machine Learning-Volume 70}, pages 2071--2080. JMLR.org, 2017.

\bibitem[Ruan et~al.(2021)Ruan, Yang, and Zhou]{Ruan2021}
Yufei Ruan, Jiaqi Yang, and Yuan Zhou.
\newblock Linear bandits with limited adaptivity and learning distributional optimal design.
\newblock In \emph{Proceedings of the 53rd Annual ACM SIGACT Symposium on Theory of Computing}, STOC 2021, page 74–87, New York, NY, USA, 2021. Association for Computing Machinery.
\newblock ISBN 9781450380539.
\newblock \doi{10.1145/3406325.3451004}.
\newblock URL \url{https://doi.org/10.1145/3406325.3451004}.

\bibitem[Sawarni et~al.(2024)Sawarni, Das, Barman, and Sinha]{sawarni24}
Ayush Sawarni, Nirjhar Das, Siddharth Barman, and Gaurav Sinha.
\newblock Generalized linear bandits with limited adaptivity.
\newblock In A.~Globerson, L.~Mackey, D.~Belgrave, A.~Fan, U.~Paquet, J.~Tomczak, and C.~Zhang, editors, \emph{Advances in Neural Information Processing Systems}, volume~37, pages 8329--8369. Curran Associates, Inc., 2024.
\newblock URL \url{https://proceedings.neurips.cc/paper_files/paper/2024/file/0faa0019b0a8fcab8e6476bc43078e2e-Paper-Conference.pdf}.

\bibitem[Zhang and Sugiyama(2023)]{Zhang2023}
Yu-Jie Zhang and Masashi Sugiyama.
\newblock Online (multinomial) logistic bandit: Improved regret and constant computation cost.
\newblock In A.~Oh, T.~Naumann, A.~Globerson, K.~Saenko, M.~Hardt, and S.~Levine, editors, \emph{Advances in Neural Information Processing Systems}, volume~36, pages 29741--29782. Curran Associates, Inc., 2023.
\newblock URL \url{https://proceedings.neurips.cc/paper_files/paper/2023/file/5ef04392708bb2340cb9b7da41225660-Paper-Conference.pdf}.

\end{thebibliography}

\newpage
\appendix
\appendixtitle

Throughout the appendix, for a matrix $\bm{A}$, we shall define $\eigmax{\bm{A}}$ and $\eigmin{\bm{A}}$ as the maximum and minimum eigenvalue of $\bm{A}$ respectively. Further, the norm of a matrix $\bm{A}$ is defined as $\lVert \bm{A} \rVert^2_2 = \eigmax{\bm{A}^\top\bm{A}}$.

Without loss of generality, we also assume that $\kappa, K, d, R, S$, and $T$ are greater than $1$ throughout the appendix.

\section{Batched Multinomial Contextual Bandit Algorithm: \texttt{B-MNL-CB}}
\label{section: regret_b_mnl}

\subsection{Notations}
\label{appendix: b_mnl_notation}
We first list a few matrices, vectors, and scalars that are commonly used throughout this section:
\begin{enumerate}
    \item $\bm{V}_\beta = \lambda\bm{I}_{d \times d} + \sum\limits_{t \in \mathcal{T}_\beta}{} \bm{x}_t \bm{x}_t^\top$
    \item $\tilde{\bm{V}}_\beta = \bm{I}_{K \times K} \otimes \bm{V}_\beta$
    \item $\bm{A}(\bm{x} , \bm{\theta}) = \textrm{diag}(\bm{z}(\bm{x} , \bm{\theta})) - \bm{z}(\bm{x} , \bm{\theta}) \bm{z}(\bm{x} , \bm{\theta})^\top$
    \item $\bm{M}(\bm{x} , \bm{\theta}_1 , \bm{\theta}_2) = \int\limits_{0}^1 \bm{A}(\bm{x} , v\bm{\theta}_1 + (1-v)\bm{\theta}_2) \; \diff v$
    \item $\bm{H}^\star_{\beta} := \lambda \bm{I}_{Kd \times Kd} + \sum\limits_{t \in \mathcal{T}_\beta} \bm{A}(\bm{x}_t, \bm{\theta}^\star)\otimes \bm{x}_t \bm{x}_t^\top$
    \item $\gamma(\lambda) = 12S \sqrt{\log T + Kd} + 8S \lambda^{-1/2}(\log T  + Kd) + 2S^{3/2} \lambda^{1/2}$ \label{definition:gamma}
    \item $B_\beta(\bm{x}) = \exp\pbrak{\sqrt{6}\min\cbrak{ \kappa^{1/2} \gamma(\delta) \matnorm{\bm{x}}{\bm{V}_\beta\inv} , 2S}}$
    \item $\bm{H}_\beta = \lambda\bm{I}_{Kd \times Kd} + \sum\limits_{t \in \mathcal{T}_\beta} \frac{\bm{A}(\bm{x}_t , \hat{\bm{\theta}}_\beta)}{\bm{B}_\beta(\bm{x}_t)} \otimes \bm{x}_t\bm{x}_t^\top$
    \item $\tilde{\bm{X}}_\beta = \frac{\bm{A}(\bm{x} , \hat{\bm{\theta}}_\beta)^{1/2}}{\sqrt{B_\beta(\bm{x})}} \otimes \bm{x}$
    \item $\tilde{\bm{x}}^{(i)}_\beta = \frac{\bm{A}(\bm{x} , \hat{\bm{\theta}}_\beta)^{1/2}}{\sqrt{B_\beta(\bm{x})}} \bm{e}_i\otimes \bm{x}$
    \item $\bm{m}_s = \left( \mathbbm{1}\{y_s = 1\} , \ldots , \mathbbm{1}\{y_s = K\} \right)^\top$
\end{enumerate}

We now present the regret upper bound for \texttt{B-MNL-CB} by restating Theorem \ref{thm: regret_b_mnl}:

\begin{theorem}
    (Regret of \texttt{B-MNL-CB)} With high probability, at the end of $T$ rounds, the regret incurred by Algorithm \ref{alg : B-MNL-CB} is bounded above by $R_T$ where
    \[
        R_T \leq \tilde{\mathcal{O}}\left( R S^{5/4} K^{5/2} d \sqrt{T} +  R S^{5/2} K^2 d^2 \kappa^{1/2} T^{1/4} \max\{e^{3S} K^{3/2} S^{-1} , \kappa^{1/2} d \} \right)
    \]
    \label{theorem: bs_mnl_cb regret}
\end{theorem}
\begin{proof}
    
From Lemma \ref{Lemma: regret_per_round}, we have an upper bound for the regret incurred for any round $t \in \mathcal{T}_{\beta+1}$. Thus, the regret incurred in batch $\beta + 1$ is given by:

\[ R_{\beta+1} \leq 16 R K^2 \gamma(\lambda) \sqrt{d \log (Kd)}\pbrak{\frac{\tau_{\beta+1}}{\sqrt{\tau_\beta}}} + 32 R K \kappa^{1/2} d\gamma^2(\lambda) \cbrak{e^{3S} K^{3/2} S^{-1} \sqrt{\log (Kd) \log d} + 12 \kappa^{1/2} d} \pbrak{\frac{\tau_{\beta+1}}{\tau_\beta}} \]

Choosing the batch lengths as $\tau_\beta = T^{1-2^{-\beta}}$ results in the following observation \citep{Hanna2023,Gao2019}:
\[\frac{\tau_{\beta+1}}{\sqrt{\tau_\beta}} \leq 2\sqrt{T} \qquad  \qquad \frac{\tau_{\beta+1}}{\tau_\beta} \leq T^{\frac{1}{4}}\]

Thus, the regret incurred in batch $\beta + 1$ is bounded by:
 \[R_{\beta+1} \leq 32 R K^2 \gamma(\lambda) \sqrt{d \log (Kd)} \sqrt{T} + 32 R K \kappa^{1/2} d\gamma^2(\lambda) \cbrak{e^{3S} K^{3/2} S^{-1} \sqrt{\log (Kd) \log d} + 12 \kappa^{1/2} d} T^{1/4}\]

We now trivially upper bound the regret for $\mathcal{T}_1$ as $R\tau_1 = R\sqrt{T}$. Thus, adding the regret incurred in each batch over all batches $\beta \in [1,\log\log T+1]$ results in:
\begin{align*}
    R_T \leq &\left( 32 R K^2 \gamma(\lambda) \sqrt{d \log (Kd)}+ R \right) \sqrt{T} \log \log T 
    \\
    &+ 32 R K \kappa^{1/2} d\gamma^2(\lambda) \cbrak{e^{3S} K^{3/2} S^{-1} \sqrt{\log (Kd) \log d} + 12 \kappa^{1/2} d} T^{1/4}  \log \log T
\end{align*}

From Lemma \ref{lemma: confidence radius}, setting $\lambda = S^{-1/2} Kd \log T$ along with the fact that $Kd + \log T \leq Kd\log T$ results in $\gamma(\lambda) \leq 22 S^{5/4} \sqrt{Kd \log T}$. Substituting the value of $\gamma(\lambda)$ gives us:

\begin{align*}
    R_T \leq &\left( 704 S^{5/4} R K^{5/2} d \sqrt{ \log T \log (Kd)}+ R \right) \sqrt{T} \log \log T 
    \\
    &+ 14784 R S^{5/2} K^2 d^2 \kappa^{1/2}  \cbrak{e^{3S} K^{3/2} S^{-1} \sqrt{\log (Kd) \log d} + 12 \kappa^{1/2} d} T^{1/4} \log^2 T  \log \log T
\end{align*}

This concludes the proof.
\end{proof}

\subsection{Supporting Lemmas for Theorem \ref{theorem: bs_mnl_cb regret}}

\begin{lemma}
    For batch $\beta$, denoted by $\mathcal{T}_\beta$, let $\{\bm{x}_1 , \ldots , \bm{x}_{\tau_\beta}\}$ be a set of i.i.d arms and $\{r_1 , \ldots , r_{\tau_\beta}\}$ be the corresponding rewards associated with these arms, where $\tau_\beta =|\mathcal{T}_\beta|$. Define $\hat{\bm\theta}_\beta$ to be the MLE estimate for this batch, i.e
    \[\hat{\bm\theta}_\beta = \argmin_{\bm\theta} \sum_{s \in \mathcal{T}_\beta} \sum_{i=1}^K \mathbbm{1} \{y_s = i\} \log z_i(\bm{x}_s , \bm\theta) + \frac{\lambda}{2} \lVert \bm\theta \rVert^2_2\]
    Let the optimal Hessian matrix for batch $\beta$, $\bm{H}_\beta^\star$, be defined as in Section \ref{appendix: b_mnl_notation}. Then, with probability greater than $1 - \frac{1}{T^2}$, we have:
    \[\matnorm{\thetastar - \hat{\bm{\theta}}_\beta}{\bm{H}_\beta^\star} \leq 12S\sqrt{\log T + Kd} + 8S \lambda^{-1/2}(\log T  + Kd) + 2S^{3/2}\lambda^{1/2}\]
    \label{lemma: confidence radius}
\end{lemma}
\begin{proof}
    For a batch $\beta$, we define the following quantity:

$$\bm{G}_{\beta}(\bm{\theta}_1 , \bm{\theta}_2) = \sum\limits_{t \in \mathcal{T}_\beta} \bm{M}(\bm{x} , \bm{\theta}_1 , \bm{\theta}_2) \otimes \bm{x}_t\bm{x}_t^\top + \lambda \bm{I}_{Kd \times Kd}$$
Then,

\begin{align*}
    \matnorm{\thetastar - \hat{\bm{\theta}}_\beta}{\bm{H}_\beta^\star} &\overset{(i)}{\leq} \sqrt{1 + 2S}\matnorm{\thetastar - \hat{\bm{\theta}}_\beta}{\bm{G}_\beta(\thetastar, \hat{\bm{\theta}}_\beta)} 
    \\
    &\leq \sqrt{1 + 2S}\matnorm{\bm{G}_\beta(\bm{\theta}^\star , \hat{\bm\theta}_\beta) \pbrak{\thetastar - \hat{\bm{\theta}}_\beta}}{\bm{G}\inv_\beta(\thetastar, \hat{\bm{\theta}}_\beta)}
    \\
    &\leq \sqrt{1 + 2S}\matnorm{\sum\limits_{t \in \mathcal{T}_\beta} \sbrak{\bm{M}(\bm{x} , \thetastar , \hat{\bm{\theta}}_\beta) \otimes \bm{x}_t\bm{x}_t^\top + \bm{I}_{Kd \times Kd}} \pbrak{\thetastar - \hat{\bm{\theta}}_\beta}}{\bm{G}\inv_\beta(\thetastar, \hat{\bm{\theta}}_\beta)}
    \\
    &\overset{(ii)}{\leq} \sqrt{1 + 2S}\matnorm{\sum\limits_{t \in \mathcal{T}_\beta} \sbrak{\bm{M}(\bm{x} , \thetastar , \hat{\bm{\theta}}_\beta) \otimes \bm{x}_t^\top}\pbrak{\thetastar - \hat{\bm{\theta}}_\beta} \otimes \bm{x}_t + \lambda \pbrak{\thetastar - \hat{\bm\theta}_\beta }}{\bm{G}\inv_\beta(\thetastar, \hat{\bm{\theta}}_\beta)}
    \\
    &\overset{(iii)}{\leq} \sqrt{1 + 2S}\matnorm{\sum\limits_{t \in \mathcal{T}_\beta} \sbrak{\bm{z}(\bm{x_t} , \thetastar) - \bm{z}(\bm{x}_t  , \hat{\bm{\theta}}_\beta)} \otimes \bm{x}_t  - \lambda \hat{\bm{\theta}}_\beta}{\bm{G}\inv_\beta(\thetastar, \hat{\bm{\theta}}_\beta)} + \lambda \sqrt{1+2S}\matnorm{\thetastar}{\bm{G}\inv_\beta(\thetastar, \hat{\bm{\theta}}_\beta)}
    \\ 
    &\overset{(iv)}{\leq} (1 + 2S)\matnorm{\sum\limits_{t \in \mathcal{T}_\beta} \sbrak{\bm{z}(\bm{x_t} , \thetastar) - \bm{z}(\bm{x}_t  , \hat{\bm{\theta}}_\beta)} \otimes \bm{x}_t - \lambda \hat{\bm{\theta}}_\beta}{\bm{H}_\beta^{\star \; -1}} + \sqrt{\lambda (1 + 2S)} \twonorm{\thetastar}
    \\
    &\overset{(v)}{\leq} 3S \matnorm{\sum\limits_{t \in \mathcal{T}_\beta} \sbrak{\bm{z}(\bm{x_t} , \thetastar) - \bm{m}_s} \otimes \bm{x}_t}{\bm{H}_\beta^{\star \; -1}} + \sqrt{3} \lambda^{1/2} S^{3/2}
\end{align*}
where $(i)$ follows from Lemma \ref{lemma: H-G relation} , $(ii)$ follows from Mixed Product Property, $(iii)$ follows from the Mean value Theorem and the triangle inequality, $(iv)$ follows from the fact that $\bm{G}_\beta \mgeq \lambda \bm{I}$ and Lemma \ref{lemma: H-G relation}, and $(v)$ follows from Lemma \ref{lemma: MLE} and the fact that $\twonorm{\bm{\theta}^\star} \leq S$.

Now, consider the following term:
\begin{align*}
    \matnorm{\sum\limits_{t \in \mathcal{T}_\beta} \sbrak{\bm{z}(\bm{x_t} , \thetastar) - \bm{m}_s} \otimes \bm{x}_t}{\bm{H}_\beta^{\star \; -1}} &= \twonorm{\sum\limits_{t \in \mathcal{T}_\beta} \bm{H}_\beta^{\star \; -1/2}\pbrak{\sbrak{\bm{z}(\bm{x_t} , \thetastar) - \bm{m}_s} \otimes \bm{x}_t}}
    \\
    &= \max\limits_{\bm{y} \in \mathcal{B}_2(Kd)} \left\langle{\bm{y} , \sum\limits_{t \in \mathcal{T}_\beta} \bm{H}_\beta^{\star \; -1/2}\pbrak{\sbrak{\bm{z}(\bm{x_t} , \thetastar) - \bm{m}_s} \otimes \bm{x}_t}}\right\rangle 
\end{align*}
where $\mathcal{B}_2(Kd)$ represents the $Kd-$dimensional unit ball with respect to the $\ell_2$ norm. We construct an $\epsilon-$net for this unit ball, denoted as $C_\epsilon$. For any $\bm{y} \in \mathcal{B}_2(Kd)$, we define  $\bm{y}_\epsilon = \argmin\limits_{\bm{x} \in C_\epsilon} \twonorm{\bm{y} - \bm{x}}$, then,

\begin{align*}
     \matnorm{\sum\limits_{t \in \mathcal{T}_\beta} \sbrak{\bm{z}(\bm{x_t} , \thetastar) - \bm{m}_s} \otimes \bm{x}_t}{\bm{H}_\beta^{\star \; -1}} 
     &= \max\limits_{\bm{y} \in \mathcal{B}_2(Kd)} \left\langle{\bm{y} , \sum\limits_{t \in \mathcal{T}_\beta}\bm{H}_\beta^{\star \; -1/2}\pbrak{\sbrak{\bm{z}(\bm{x_t} , \thetastar) - \bm{m}_s} \otimes \bm{x}_t}}\right\rangle  
     \\
     &= \max\limits_{\bm{y} \in \mathcal{B}_2(Kd)} \left\langle{(\bm{y} - \bm{y}_\epsilon) + \bm{y}_\epsilon , \sum\limits_{t \in \mathcal{T}_\beta} \bm{H}_\beta^{\star \; -1/2}\pbrak{\sbrak{\bm{z}(\bm{x_t} , \thetastar) - \bm{m}_s} \otimes \bm{x}_t}}\right\rangle 
\end{align*}

Thus, an application of the Cauchy-Schwarz inequality along with the fact that $\twonorm{\bm{y} - \bm{y}_\epsilon} \leq \epsilon$ gives us
\[ \matnorm{\sum\limits_{t \in \mathcal{T}_\beta} \sbrak{\bm{z}(\bm{x_t} , \thetastar) - \bm{m}_s} \otimes \bm{x}_t}{\bm{H}_\beta^{\star \; -1}}  \leq  \frac{1}{1-\epsilon} \left\langle{\bm{y}_\epsilon , \sum\limits_{t \in \mathcal{T}_\beta}\bm{H}_\beta^{\star \; -\frac{1}{2}}\pbrak{\sbrak{\bm{z}(\bm{x_t} , \thetastar) - \bm{m}_s} \otimes \bm{x}_t}}\right\rangle \]

The above term can be bounded using the Bernstein Inequality (Lemma \ref{lemma: bernstein}), which has been done in Lemma \ref{lemma: application:Bernstein}. We note that $|C_\epsilon| \leq \pbrak{\frac{2}{\epsilon}}^{Kd}$. We now set $\epsilon = 0.5$ and $\delta = (T^2 |C_\epsilon|)^{-1}$ and then perform a union bound over $C_\epsilon$. We get that with probability greater than $1 - \frac{1}{T^2}$, we have:

\begin{align*}
    \matnorm{\sum\limits_{t \in \mathcal{T}_\beta} \sbrak{\bm{z}(\bm{x_t} , \thetastar) - \bm{m}_s} \otimes \bm{x}_t}{\bm{H}_\beta^{\star \; -1}} &\leq 2\pbrak{\sqrt{2\log\pbrak{T^2 4^{Kd}}} +\frac{4}{3}\lambda^{-1/2}\log (T^2 4^{Kd})}
    \\
    &\leq 4\sqrt{\log T + Kd} + \frac{8}{3}\lambda^{-1/2}(\log T  + Kd)
\end{align*}
Substituting this into the original bound finishes the proof.
\end{proof} 

\begin{lemma}
    Let $\bm{y}$ be a fixed vector with $\twonorm{\bm{y}} \leq 1$, then, with probability at least $1 - \delta$
    $$\sum\limits_{t \in \mathcal{T}_\beta} \left[\bm{y}^\top\bm{H}_\beta^{\star \; -\frac{1}{2}}\sbrak{\bm{z}(\bm{x_t} , \thetastar) - \bm{m}_s} \otimes \bm{x}_t \right] \leq \sqrt{2\log\frac{1}{\delta}} + \frac{4}{3\sqrt{\lambda}}\log\frac{1}{\delta}$$
    \label{lemma: application:Bernstein} 
\end{lemma}
\begin{proof}
    
Denote $\varphi_t = \bm{y}^\top\bm{H}_\beta^{\star \; -\frac{1}{2}}\pbrak{\sbrak{\bm{z}(\bm{x_t} , \thetastar) - \bm{m}_s} \otimes \bm{x}_t}$. From Lemma \ref{lemma: expectation of m}, we have that $\E\sbrak{\varphi_t} = 0$.

Also,
\begin{align*}
    \mathbb{V}\sbrak{\varphi_t} &= \E\sbrak{\varphi_t^2} - \E\sbrak{\varphi_t}^2
    \overset{(i)}{=} \E\sbrak{\varphi_t\varphi_t^\top}
    \\
    &= \E\sbrak{\bm{y}^\top\bm{H}_\beta^{\star \; -\frac{1}{2}} \pbrak{\sbrak{\bm{z}(\bm{x_t} , \thetastar) - \bm{m}_s} \otimes \bm{x}_t}\pbrak{\sbrak{\bm{z}(\bm{x_t} , \thetastar) - \bm{m}_s} \otimes \bm{x}_t}^\top \bm{H}_\beta^{\star \; -\frac{1}{2}} \bm{y}}
    \\
    &\overset{(ii)}{=} \bm{y}^\top\bm{H}_\beta^{\star \; -\frac{1}{2}} \E\sbrak{\sbrak{\bm{z}(\bm{x_t} , \thetastar) - \bm{m}_s}\sbrak{\bm{z}(\bm{x_t} , \thetastar) - \bm{m}_s}^\top \otimes \bm{x}_t\bm{x}_t^\top}\bm{H}_\beta^{\star \; -\frac{1}{2}} \bm{y}
    \\
    &= \bm{y}^\top\bm{H}_\beta^{\star \; -\frac{1}{2}} \pbrak{\E\sbrak{\sbrak{\bm{z}(\bm{x_t} , \thetastar) - \bm{m}_s}\sbrak{\bm{z}(\bm{x_t} , \thetastar) - \bm{m}_s}^\top} \otimes \bm{x}_t\bm{x}_t^\top} \bm{H}_\beta^{\star \; -\frac{1}{2}} \bm{y}
    \\
    &\overset{(iii)}{=} \bm{y}^\top\bm{H}_\beta^{\star \; -\frac{1}{2}} \pbrak{\bm{A}(\bm{x}_t , \thetastar) \otimes \bm{x}_t\bm{x}_t^\top} \bm{H}_\beta^{\star \; -\frac{1}{2}} \bm{y}
    \overset{(iv)}{=} \bm{y}^\top\bm{H}_\beta^{\star \; -\frac{1}{2}} \pbrak{\bm{H}_\beta^{\star} - \lambda \bm{I}} \bm{H}_\beta^{\star \; -\frac{1}{2}} \bm{y} 
    \\
    &\leq \bm{y}^\top\bm{y}
    \leq 1
\end{align*}

where $(i)$ follows from the fact that $\varphi_t$ is a scalar and $\E[\varphi_t] = 0$, $(ii)$ follows from the fact that $\pbrak{\bm{A} \otimes \bm{B}}^\top = \bm{A}^\top \otimes \bm{B}^\top$ and the mixed-product property of the Kronecker Product, $(iii)$ follows from Lemma \ref{lemma: expectation of m}, and $(iv)$ follows from the definition of $\bm{H}_\beta^\star$.

Finally, we note that
\begin{align*}
    \modulus{\varphi_t - \E\sbrak{\varphi_t}} &= \modulus{\varphi_t}
    = \modulus{\bm{y}^\top\bm{H}_\beta^{\star \; -\frac{1}{2}}\pbrak{\sbrak{\bm{z}(\bm{x_t} , \thetastar) - \bm{m}_s} \otimes \bm{x}_t}}
    \overset{(i)}{\leq} \twonorm{\bm{y}}\twonorm{\bm{H}_\beta^{\star \; -\frac{1}{2}}\pbrak{\sbrak{\bm{z}(\bm{x_t} , \thetastar) - \bm{m}_s} \otimes \bm{x}_t}}
    \\
    &\overset{(ii)}{\leq}\norm{\bm{H}_\beta^{\star \; -\frac{1}{2}}}\twonorm{\pbrak{\bm{z}(\bm{x_t} , \thetastar) - \bm{m}_s} \otimes \bm{x}_t}
    \overset{(iii)}{\leq}\frac{1}{\sqrt{\lambda}}\twonorm{\bm{z}(\bm{x_t} , \thetastar) - \bm{m}_s} \twonorm{\bm{x}_t}
    \\
    &\overset{(iv)}{\leq}\frac{1}{\sqrt{\lambda}} \pbrak{\twonorm{\bm{z}(\bm{x_t} , \thetastar)} + \twonorm{\bm{m}_s}}
    \overset{(v)}{\leq} \frac{2}{\sqrt{\lambda}}
    \end{align*}

where $(i)$ follows from Cauchy-Schwarz, $(ii)$ follows from the fact that $\twonorm{\bm{y}} \leq 1$ and $\twonorm{\bm{A}\bm{x}} \leq \norm{\bm{A}}\twonorm{\bm{x}}$, $(iii)$ follows from $\bm{H}_\beta^\star \mgeq \lambda\bm{I}$ and the fact that $\twonorm{\bm{a}\otimes\bm{b}} = \twonorm{\bm{a}}\twonorm{\bm{b}}$, $(iv)$ follows from $\twonorm{\bm{x}} \leq 1$ and uses the triangle inequality, and $(v)$ follows from the fact $\lVert \bm{z}(\bm{x} , \bm\theta) \rVert_2 \leq \lVert \bm{z}(\bm{x} , \bm\theta) \rVert_1 \leq 1 $. 

Substituting $v = 1$ and $b = \frac{2}{\sqrt{\lambda}}$ in Lemma \ref{lemma: bernstein} finishes the proof.
\end{proof}

\begin{lemma}
Let $\tilde{\bm{V}}_\beta$ and $\bm{H}_\beta^\star$ be the design and optimal Hessian matrices defined as in Section \ref{appendix: b_mnl_notation}. Then, we have that
\[\tilde{\bm{V}}_\beta \mleq \kappa \bm{H}^\star_\beta\]    \label{lemma : relation between V and H}
\end{lemma}
\begin{proof}
    From the definition of $\kappa$, we know that $\bm{A}(\bm{x} , \bm{\theta}) \mgeq \frac{1}{\kappa}\bm{I}$.

Hence, using the fact that $\kappa > 1$, we can say that
\begin{align*}
    \tilde{\bm{V}}_\beta &= \bm{I}_{K \times K} \otimes \bm{V}_\beta
    =\bm{I}_{K \times K} \otimes \pbrak{\lambda\bm{I}_{d\times d} + \sum\limits_{t \in \mathcal{T}_\beta}\bm{x}_t \bm{x}_t^\top}
    = \lambda\bm{I}_{Kd \times Kd} +  \bm{I}_{K\times K}\otimes \sum\limits_{t \in \mathcal{T}_\beta} \bm{x}_t \bm{x}_t^\top
    \\
    &\mleq \lambda \bm{I}_{Kd \times Kd} + \kappa \sum\limits_{t \in \mathcal{T}_\beta} \bm{A}(\bm{x}_t , \thetastar) \otimes \bm{x}_t \bm{x}_t ^\top
    \mleq \kappa \bm{H}^\star_\beta
\end{align*}

\end{proof}

\begin{lemma}  
    Let $\bm{H}_\beta^\star$ and $\bm{H}_\beta$ be the optimal and proxy Hessian matrices in batch $\beta$ as defined in Section \ref{appendix: b_mnl_notation}. Then, we have that
    \[\bm{H}_\beta \mleq \bm{H}^\star_\beta\]
    \label{lemma: relation between H and H}
\end{lemma}
\begin{proof}
    From Lemma \ref{lemma: self-concordance}, we have that
$$\bm{A}(\bm{x} , \hat{\bm{\theta}}_\beta) \mleq \bm{A}(\bm{x} , \thetastar) \exp\pbrak{\sqrt{6}\twonorm{(\bm{I} \otimes \bm{x}^\top)(\thetastar - \hat{\bm{\theta}}_\beta)}}$$

We can bound $\twonorm{(\bm{I} \otimes \bm{x}^\top)(\thetastar - \hat{\bm{\theta}}_\beta)}$ as follows:

\begin{align*}
    \twonorm{(\bm{I} \otimes \bm{x}^\top)(\thetastar - \hat{\bm{\theta}}_\beta)} &\overset{(i)}{\leq} 2S\twonorm{\bm{I}\otimes\bm{x}^\top}
    \overset{(ii)}{=} 2S \sqrt{\eigmax{(\bm{I}\otimes\bm{x})(\bm{I}\otimes\bm{x}^\top)}}
    \\
     &\overset{(iii)}{=} 2S \sqrt{\eigmax{\bm{I} \otimes \bm{x}\bm{x}^\top}}
     \overset{(iv)}{\leq} 2S
\end{align*}
where $(i)$ uses the sub-multiplicativity of the norm, a triangle inequality, and the fact that $\twonorm{\thetastar} \leq S$, $(ii)$ uses the definition of the norm, i.e., $\twonorm{\bm{A}} = \sqrt{\eigmax{\bm{A}^\top\bm{A}}}$, $(iii)$ follows from the Mixed-Product property of Kronecker Products, and $(iv)$ follows from the fact that $\lambda\pbrak{\bm{A}\otimes\bm{B}} = \lambda(\bm{A})\lambda(\bm{B})$ and since $\bm{x}\bm{x}^\top$ is a rank-one matrix, the only eigenvalues are $\lVert \bm{x} \rVert^2_2$ and $0$, and $0 \leq \twonorm{\bm{x}} \leq 1$. 

We can also bound  $\twonorm{(\bm{I} \otimes \bm{x}^\top)(\thetastar - \hat{\bm{\theta}_\beta})}$ as follows:
\
\begin{align*}
   \twonorm{(\bm{I} \otimes \bm{x}^\top)(\thetastar - \hat{\bm{\theta}}_\beta)} &=  \twonorm{(\bm{I} \otimes \bm{x}^\top) \bm{H}^{\star \; -1/2}_\beta \bm{H}^{\star \; 1/2}_\beta (\thetastar - \hat{\bm{\theta}}_\beta)}
    \overset{(i)}{\leq}\twonorm{(\bm{I} \otimes \bm{x}^\top) \bm{H}^{\star \; -1/2}_\beta}  \matnorm{\thetastar - \hat{\bm{\theta}}_\beta}{\bm{H}^\star_\beta}
    \\
    &\overset{(ii)}{\leq} \kappa^{1/2} \gamma(\lambda)\twonorm{(\bm{I} \otimes \bm{x}^\top) \kappa^{1/2} \tilde{\bm{V}}^{-1/2}_\beta}
    \overset{(iii)}{=}\kappa^{1/2} \gamma(\lambda)\sqrt{ \eigmax{\tilde{\bm{V}}^{-1/2}_\beta (\bm{I} \otimes \bm{x}) (\bm{I} \otimes \bm{x}^\top) \tilde{\bm{V}}^{-1/2}_\beta}}
    \\
    &\overset{(iv)}{=}\kappa^{1/2} \gamma(\lambda)\sqrt{ \eigmax{(\bm{I}\otimes \bm{V}^{-1/2}_\beta) (\bm{I} \otimes \bm{x}) (\bm{I} \otimes \bm{x}^\top) (\bm{I}\otimes \bm{V}^{-1/2}_\beta)}}
    \\
    &\overset{(v)}{=}\kappa^{1/2} \gamma(\lambda)\sqrt{ \eigmax{\bm{I} 
    \otimes \bm{V}^{-1/2}_\beta \bm{x}
    \bm{x}^\top\bm{V}^{-1/2}_\beta}}
    \overset{(vi)}{=} \kappa^{1/2} \gamma(\lambda)\matnorm{\bm{x}}{\bm{V}\inv_\beta}
\end{align*}
where $(i)$ follows from the sub-multiplicativity of the norm, $(ii)$ follows from Lemma \ref{lemma: confidence radius} and Lemma \ref{lemma : relation between V and H}, $(iii)$ follows from the definition of the norm, $(iv)$ follows from the definition of $\tilde{\bm{V}}_\beta$ and the fact that $(\bm{A} \otimes \bm{B})^n = \bm{A}^n \otimes \bm{B}^n$, $(v)$ follows from the Mixed-Product property, and $(vi)$ follows from $\lambda(\bm{A}\otimes\bm{B}) = \lambda(\bm{A}) \lambda(\bm{B})$.

Thus, we can say that $\twonorm{(\bm{I} \otimes \bm{x}^\top)(\thetastar - \hat{\bm{\theta}}_\beta)} \leq \min\cbrak{\gamma(\lambda)\kappa^{1/2}\matnorm{\bm{x}}{\bm{V}\inv_\beta} , 2S}$.

Define $B_\beta(\bm{x}) = \exp\pbrak{\sqrt{6}\min\cbrak{\gamma(\lambda)\kappa^{1/2}\matnorm{\bm{x}}{\bm{V}\inv_\beta} , 2S}}$. Then, $\bm{A}(\bm{x} , \hat{\bm{\theta}}_\beta) \mleq \bm{A}(\bm{x} , \thetastar)  B_\beta(\bm{x})$. Hence, we can say,
\[\bm{H}_\beta = \lambda\bm{I}_{Kd \times Kd} + \sum\limits_{t \in \beta} \frac{\bm{A}(\bm{x}_t , \hat{\bm{\theta}}_\beta)}{\bm{B}_\beta(\bm{x}_t)} \otimes \bm{x}_t\bm{x}_t^\top \mleq \lambda\bm{I}_{Kd \times Kd} + \sum\limits_{t \in \beta} \bm{A}(\bm{x}_t , \thetastar) \otimes \bm{x}_t\bm{x}_t^\top = \bm{H}^\star_\beta\]
\end{proof} 

\begin{lemma}
    (Proposition 1 , \cite{Zhang2023}) For any arm $\bm{x}$, we have that, 
    $$\modulus{\bm{\rho}^\top\bm{z}(\bm{x} , \thetastar) - \bm{\rho}^\top\bm{z}(\bm{x} , \bm{\theta}_j)} \leq \epsilon_1(j , \bm{x} , \lambda) + \epsilon_2(j , \bm{x} , \lambda)$$

    where 
    \[\epsilon_1(j , \bm{x} , \lambda) = \gamma(\lambda)\twonorm{\bm{H}_j^{-1/2}(\bm{I} \otimes \bm{x})\bm{A}(\bm{x} , \bm{\theta}_j)\bm{\rho}} \text{ and } \epsilon_2(j , \bm{x} , \lambda) = 3R\gamma(\lambda)^2\twonorm{(\bm{I} \otimes \bm{x}^\top)\bm{H}_j^{-1/2}}^2\]
    \label{lemma:UCB-LCB}
\end{lemma}
\begin{proof}
    We provide the proof for the sake of completeness:
\begin{align*}
    \modulus{\bm{\rho}^\top\bm{z}(\bm{x} , \thetastar) - \bm{\rho}^\top\bm{z}(\bm{x} , \bm{\theta}_j)} &= \modulus{\sum\limits_{i=1}^{K}\rho_i\sbrak{z_i(\bm{x} , \thetastar) - z_i(\bm{x} , \bm{\theta}_j)}}
    \\
    &= \modulus{\sum\limits_{i=1}^{K} \rho_i \bm{\nabla}z_i(\bm{x} , \bm{\theta}_j)^\top \sbrak{(\bm{I}_{K\times K} \otimes \bm{x}^\top)(\thetastar - \bm{\theta}_j)} + \sum\limits_{i=1}^{K} \rho_i \matnorm{(\bm{I}_{K\times K} \otimes \bm{x}^\top)(\thetastar - \bm{\theta}_j)}{\bm{Z}_i}}\\
    &\leq \modulus{\bm{\rho}^\top \bm{A}(\bm{x} , \bm{\theta}_j) (\bm{I}_{K\times K} \otimes \bm{x}^\top)(\thetastar - \bm{\theta}_j)} + \modulus{\sum\limits_{i=1}^{K} \rho_i \matnorm{(\bm{I}_{K\times K} \otimes \bm{x}^\top)(\thetastar - \bm{\theta}_j)}{\bm{Z}_i}^2}
\end{align*}

where 
\[\bm{Z}_i = \int\limits_{0}^1 (1-v) \nabla^2 z_i(\bm{x} , v \bm\theta^\star + (1-v)\bm\theta_j) \; \diff v\]

Beginning with the first term : 
\begin{align*}
  \modulus{\bm{\rho}^\top \bm{A}(\bm{x} , \bm{\theta}_j) (\bm{I}_{K\times K} \otimes \bm{x}^\top)(\thetastar - \bm{\theta}_j)}  &=  \modulus{\bm{\rho}^\top \bm{A}(\bm{x} , \bm{\theta}_j) (\bm{I}_{K\times K} \otimes \bm{x}^\top)\bm{H}^{\star\;-1/2}_j \bm{H}^{\star\;1/2}_j (\thetastar - \bm{\theta}_j)}
  \\
   &\overset{(i)}{\leq} \matnorm{\thetastar - \bm{\theta}_j}{\bm{H}^\star_j} \twonorm{\bm{\rho}^\top \bm{A}(\bm{x} , \bm{\theta}_j) (\bm{I}_{K\times K} \otimes \bm{x}^\top)\bm{H}^{\star\;-1/2}_j}
   \\
   &\leq \gamma(\lambda) \twonorm{\bm{H}^{\star\;-1/2}_j (\bm{I}_{K\times K} \otimes \bm{x}) \bm{A}(\bm{x} , \bm{\theta}_j)\bm{\rho} }\\
   &\overset{(ii)}\leq \gamma(\lambda) \twonorm{\bm{H}^{-1/2}_j (\bm{I}_{K\times K} \otimes \bm{x}) \bm{A}(\bm{x} , \bm{\theta}_j)\bm{\rho} }
\end{align*}
where $(i)$ follows from the sub-multiplicativity of the norm and $(ii)$ is due to Lemma \ref{lemma: relation between H and H}.

For the second term, for some $k \in [1,K]$, we make the following observation:
$$\bm{Z}_k = \int\limits_{0}^1 (1-v) \nabla^2 z_k(\bm{x} , v\bm\theta^\star + (1-v)\bm\theta_j) \; \diff v \mleq 3\bm{I}\int\limits_{0}^1 (1-v) \; \diff v \mleq 3\bm{I}$$

Thus, we have:
\begin{align*}
    \modulus{\sum\limits_{i=1}^{K} \rho_i \matnorm{(\bm{I}_{K\times K} \otimes \bm{x}^\top)(\thetastar - \bm{\theta}_j)}{\bm{Z}_i}^2} &\leq \modulus{\sum\limits_{i=1}^{K} 3\rho_i \twonorm{(\bm{I}_{K\times K} \otimes \bm{x}^\top)(\thetastar - \bm{\theta}_j)}^2}
    \\
    &\leq 3R \twonorm{(\bm{I}_{K\times K} \otimes \bm{x}^\top)\bm{H}^{\star\;-1/2}_j \bm{H}^{\star\;1/2}_j (\thetastar - \bm{\theta}_j)}^2
    \\
    &\leq 3R \matnorm{\thetastar - \bm{\theta}_j}{\bm{H}^\star_j}^2 \twonorm{(\bm{I}_{K\times K} \otimes \bm{x}^\top)\bm{H}^{\star\;-1/2}_j}^2
    \\
    &\leq 3R \gamma(\lambda)^2 \twonorm{(\bm{I}_{K\times K} \otimes \bm{x}^\top)\bm{H}^{\star\;-1/2}_j}^2
    \\
    &\leq 3R \gamma(\lambda)^2 \twonorm{(\bm{I}_{K\times K} \otimes \bm{x}^\top)\bm{H}^{-1/2}_j}^2
\end{align*}
\end{proof} 

\begin{lemma}
    Let $\bm{x}^\star_t$ be the optimal arm at round $t$, i.e $\bm{x}^\star_t = \argmax_{\bm{x} \in \mathcal{X}_t} \bm{\rho}^\top \bm{z}(\bm{x} , \bm\theta^\star)$. Then, the optimal arm never gets eliminated in any round.
    \label{lemma: optimal arm remains}
\end{lemma}

\begin{proof}
    From Lemma \ref{lemma:UCB-LCB}, we know that 
$$\modulus{\bm{\rho}^\top\bm{z}(\bm{x} , \thetastar) - \bm{\rho}^\top\bm{z}(\bm{x} , \bm{\theta}_j)} \leq \epsilon_1(j , \bm{x} , \lambda) + \epsilon_2(j , \bm{x} , \lambda)$$

Also, from Algorithm \ref{alg : B-MNL-CB}, we have the definitions of $\textrm{UCB}(j , \bm{x} , \lambda)$ and $\textrm{LCB}(j , \bm{x} , \lambda)$ as: 
\[\textrm{UCB}(j , \bm{x} , \lambda) = \bm{\rho}^\top\bm{z}(\bm{x} , \bm{\theta}_j) + \epsilon_1(j , \bm{x} , \lambda) + \epsilon_2(j , \bm{x} , \lambda)\]
\[\textrm{LCB}(j , \bm{x} , \lambda) = \bm{\rho}^\top\bm{z}(\bm{x} , \bm{\theta}_j) - \epsilon_1(j , \bm{x} , \lambda) - \epsilon_2(j , \bm{x} , \lambda)\]

From Algorithm \ref{alg : B-MNL-CB}, we know that an arm $\bm{x}\in \mathcal{X}_t$ gets eliminated if 
$\textrm{UCB}(j , \bm{x} , \lambda) \leq \max_{\bm{y} \in \mathcal{X}_t} \textrm{LCB}(j , \bm{y} , \lambda)$. Thus, showing that $\textrm{UCB}(j , \bm{x}^\star_t , \lambda) \geq \max_{\bm{y} \in \mathcal{X}_t} \textrm{LCB}(j , \bm{y} , \lambda)$ accounts to showing that $\bm{x}^\star_t$ never gets eliminated.

We assume that $\argmax_{\bm{y} \in \mathcal{X}_t} \textrm{LCB}(j , \bm{y} , \lambda) = \bm{y}$. Then, for any arm $\bm{x} \in \mathcal{X}_t$, we have that
\begin{align*}
    \textrm{LCB}(j ,\bm{x} , \lambda) &\leq \max\limits_{\bm{y} \in \mathcal{X}_t}\textrm{LCB}(j , \bm{y} , \lambda)\\
    &= \bm{\rho}^\top\bm{z}(\bm{y} , \bm{\theta}_j) - \epsilon_1(j , \bm{y} , \lambda) - \epsilon_2(j , \bm{y} , \lambda)\\
    &\overset{(i)}{\leq} \sbrak{\bm{\rho}^\top\bm{z}(\bm{y} , \thetastar) + \epsilon_1(j , \bm{y} , \lambda) + \epsilon_2(j , \bm{y} , \lambda)}- \epsilon_1(j , \bm{y} , \lambda) - \epsilon_2(j , \bm{y} , \lambda) \\
    &= \bm{\rho}^\top\bm{z}(\bm{y} , \thetastar)\\
    &\overset{(ii)}{\leq} \bm{\rho}^\top\bm{z}(\bm{x}^\star_t , \thetastar)\\
    &\overset{(iii)}{\leq}  \bm{\rho}^\top\bm{z}(\bm{x}^\star_t , \bm{\theta}_j) + \epsilon_1(j , \bm{x}^\star_t , \lambda) + \epsilon_2(j , \bm{x}^\star_t , \lambda)\\
    &= \textrm{UCB}(j , \bm{x}^\star_t , \lambda)
\end{align*}

where $(i)$ follows from Lemma \ref{lemma:UCB-LCB}, $(ii)$ follows from the fact that $\bm{x}_t^\star = \argmax\limits_{\bm{y} \in \mathcal{X}_t} \bm{\rho}^\top\bm{z}(\bm{y} , \thetastar)$, and $(iii)$ again follows from Lemma \ref{lemma:UCB-LCB}.
\end{proof}

\begin{lemma} Let $B_\beta(\bm{x})$ be as defined in Section \ref{appendix: b_mnl_notation}. Then, we have that
\[\sqrt{B_\beta(\bm{x})} \leq \frac{1}{2}e^{3S} \gamma(\lambda) \kappa^{1/2} S^{-1} \lVert \bm{x} \rVert_{\bm{V}_\beta^{-1}} + 1\]
\label{lemma: sqrt_B}
\end{lemma}
\begin{proof}
    \[\sqrt{B_\beta(\bm{x})} = \exp \left(\sqrt{6} \min \left\{ S , \frac{1}{2} \gamma(\lambda) \kappa^{1/2} \lVert \bm{x} \rVert_{\bm{V}_\beta^{-1}} \right\} \right) \leq \frac{1}{2}e^{3S} \gamma(\lambda) \kappa^{1/2} S^{-1} \lVert \bm{x} \rVert_{\bm{V}_\beta^{-1}} + 1\]

where the inequality follows from Lemma \ref{lemma: exp_relation} by choosing $\min\{2S , \gamma(\lambda) \kappa^
{1/2} \lVert \bm{x} \rVert_{\bm{V}_\beta^{-1}} \} = \gamma(\lambda) \kappa^
{1/2} \lVert \bm{x} \rVert_{\bm{V}_\beta^{-1}}$ and $M = \sqrt{6}S$.

\end{proof}

\begin{lemma}
Let $\epsilon_1(\beta , \bm{x} , \lambda)$ be as defined in Lemma \ref{lemma:UCB-LCB}. Then, we have
    \[\E_{\mathcal{X} \sim \mathcal{D}_{\beta+1}} \sbrak{\max\limits_{\bm{x}\in\mathcal{X}}\epsilon_1(\beta , \bm{x} , \lambda)} \leq \frac{8 R \kappa^{1/2} K^{5/2} d e^{3S}\gamma(\lambda)^2 S^{-1} \sqrt{\log Kd \log d}}{\tau_\beta} +  \frac{4R K^2 d^{1/2} \gamma(\lambda) \sqrt{\log (Kd)}}{\sqrt{\tau_\beta}}\]
    \label{lemma: expectation of eps1}
\end{lemma}

\begin{proof}
\begin{align*}
    \E_{\mathcal{X} \sim \mathcal{D}_{\beta+1}} \sbrak{\max\limits_{\bm{x}\in\mathcal{X}}\epsilon_1(\beta , \bm{x} , \lambda)} &= \E_{\mathcal{X} \sim \mathcal{D}_{\beta+1}}\sbrak{\max\limits_{\bm{x}\in\mathcal{X}}\gamma(\lambda)\twonorm{\bm{H}_\beta^{-1/2}(\bm{I} \otimes \bm{x})\bm{A}(\bm{x} , \hat{\bm\theta}_\beta)\bm{\rho}}}
    \\
    &\overset{(i)}{\leq} \gamma(\lambda) \E_{\mathcal{X} \sim \mathcal{D}_{\beta+1}}\sbrak{\max\limits_{\bm{x}\in\mathcal{X}}\twonorm{\bm{H}_\beta^{-1/2}(\bm{I} \otimes \bm{x})\bm{A}(\bm{x} , \hat{\bm\theta}_\beta)^{1/2}}\matnorm{\bm{\rho}}{\bm{A}(\bm{x} , \hat{\bm\theta}_\beta)}}\\
    &\overset{(ii)}{\leq} R\gamma(\lambda) \E_{\mathcal{X} \sim \mathcal{D}_{\beta+1}}\sbrak{\max\limits_{\bm{x}\in\mathcal{X}} \twonorm{\bm{A}(\bm{x} , \hat{\bm\theta}_\beta)^{1/2}(\bm{I} \otimes \bm{x}^\top)\bm{H}_\beta^{-1/2}}}
    \\
    &\overset{(iii)}{\leq} R\gamma(\lambda) \E_{\mathcal{X} \sim\mathcal{D}_{\beta+1}}\sbrak{\max\limits_{\bm{x}\in\mathcal{X}}\twonorm{\sqrt{B_\beta(\bm{x})}\tilde{\bm{X}}_
    \beta^\top\bm{H}_\beta^{-1/2}}}
    \\
    &\overset{(iv)}{\leq}  4R\gamma(\lambda)K^2 \sqrt{\frac{d\log Kd}{\tau_\beta}} \cbrak{\frac{1}{2} e^{3S}\gamma(\lambda)\kappa^{1/2}S^{-1} \E\limits_{\mathcal{X} \sim \mathcal{D}_{\beta+1}}\left[ \max\limits_{\bm{x} \in \mathcal{X}} \lVert \bm{x} \rVert_{\bm{V}_\beta^{-1}}\right] + 1 }
    \\
    &\overset{(v)}{\leq}  4R\gamma(\lambda)K^2 \sqrt{\frac{d\log Kd}{\tau_\beta}} \cbrak{2 e^{3S}\gamma(\lambda)\kappa^{1/2}S^{-1} \sqrt{\frac{K d \log d}{\tau_\beta}} + 1 }
    \\
    &\overset{}{\leq}  \frac{8 R \kappa^{1/2} K^{5/2} d e^{3S}\gamma(\lambda)^2 S^{-1} \sqrt{\log Kd \log d}}{\tau_\beta} +  \frac{4R K^2 d^{1/2} \gamma(\lambda) \sqrt{\log (Kd)}}{\sqrt{\tau_\beta}}
\end{align*}

where $(i)$ follows from $\twonorm{\bm{A}\bm{x}} \leq \twonorm{\bm{A}}\twonorm{\bm{x}}$, $(ii)$ follows from the fact that $\bm{A}(\bm{x},  \bm{\theta}) \mleq \bm{I}$, $(iii)$ follows from the defintion of $\tilde{\bm{X}}$, $(iv)$ follows from Lemma \ref{lemma: sqrt_B}, the fact that $\max\cbrak{ab}\leq\max\cbrak{a}\max\cbrak{b}$,  and Lemma \ref{lemma_OD: optimal_design_on_eigmax}, and $(v)$ follows from Lemma \ref{lemma_OD: optimal_design_wrt_V}.

\end{proof}

\begin{lemma}
Let $\epsilon_2(\beta , \bm{x} , \lambda)$ be as defined in Lemma \ref{lemma:UCB-LCB}. Then, we have
    $$\E_{\mathcal{X} \sim \mathcal{D}_{\beta+1}} \sbrak{\epsilon_2(\beta , \bm{x} , \lambda)} \leq \frac{96R\kappa\gamma(\lambda)^2}{\tau_\beta} Kd^2$$
    \label{lemma: expectation of eps2}
\end{lemma}
\begin{proof}
Recall from Lemma \ref{lemma:UCB-LCB}, in one of the intermediate steps, we have that 
\[\epsilon_2(\beta , \bm{x} , \lambda) = 3R\gamma(\lambda)^2\twonorm{(\bm{I} \otimes \bm{x}^\top)\bm{H}_\beta^{\star \; -1/2}}^2\]

Thus, we have

\begin{align*}
    \E_{\mathcal{X}\sim\mathcal{D}_{\beta+1}}\sbrak{\epsilon_2(\beta , \bm{x} , \lambda)} &= \E_{\mathcal{X}\sim\mathcal{D}_{\beta+1}}\sbrak{\max\limits_{\bm{x} \in \mathcal{X}} 3R\gamma(\lambda)^2\twonorm{(\bm{I} \otimes \bm{x}^\top)\bm{H}_\beta^{\star \; -1/2}}^2}
    \\
    &=3 R \gamma(\lambda)^2 \E_{\mathcal{X}\sim\mathcal{D}_{\beta+1}}\sbrak{\max\limits_{\bm{x} \in \mathcal{X}} \twonorm{(\bm{I} \otimes \bm{x}^\top)\bm{H}_\beta^{\star \; -1/2}}^2}
    \\
    &\overset{(i)}\leq 3R\kappa\gamma(\lambda)^2\E_{\mathcal{X}\sim\mathcal{D}_{\beta+1}}\sbrak{\max\limits_{\bm{x} \in \mathcal{X}} \twonorm{(\bm{I} \otimes \bm{x}^\top)\tilde{\bm{V}}_\beta^{-1/2}}^2}\\
    &\overset{(ii)}\leq 
    3R\kappa\gamma(\lambda)^2\E_{\mathcal{X}\sim\mathcal{D}_{\beta+1}}\sbrak{\max\limits_{\bm{x} \in \mathcal{X}} \twonorm{(\bm{I} \otimes \bm{x}^\top)(\bm{I} \otimes\bm{V}_\beta^{-1/2})}^2}\\
    &\overset{(iii)}\leq    3R\kappa\gamma(\lambda)^2\E_{\mathcal{X}\sim\mathcal{D}_{\beta+1}}\sbrak{\max\limits_{\bm{x} \in \mathcal{X}} \matnorm{\bm{x}}{\bm{V}_\beta\inv}^2 }
    \overset{(iv)}\leq \frac{48R\kappa\gamma(\lambda)^2}{\tau_\beta} (K+1)d^2 
    \leq \frac{96R\kappa\gamma(\lambda)^2}{\tau_\beta} Kd^2 \\
\end{align*}

where $(i)$ follows from Lemma \ref{lemma : relation between V and H}, $(ii)$ follows from the definition of $\tilde{\bm{V}}_\beta$, $(iii)$ follows from the Mixed-Product Property and the fact that $\lambda(\bm{A}\otimes\bm{B}) = \lambda(\bm{A})\lambda(\bm{B})$, and $(iv)$ follows from Lemma \ref{lemma_OD: optimal_design_norm_squared_for_V}.

\end{proof}

\begin{lemma}
    Let t be a time round in batch $\beta+1$, i.e $t \in \mathcal{T}_\beta$. Then, the expected regret incurred at round $t$, denoted as $R_t$ can be bounded as:
    \begin{align*}
        R_t \leq \frac{32RK \kappa^{1/2} d \gamma(\lambda)^2}{\tau_\beta}\cbrak{e^{3S} K^{3/2} S^{-1}\sqrt{\log (Kd) \log d} + 12 \kappa^{1/2}d} + \frac{16R K^2 d^{1/2}  \gamma(\lambda) \sqrt{\log (Kd)}}{\sqrt{\tau_\beta}} 
    \end{align*}
    \label{Lemma: regret_per_round}
\end{lemma}
\begin{proof}
Using Lemma \ref{lemma:UCB-LCB},
\[\bm{\rho}^\top\bm{z}(\bm{x}_t^\star , \thetastar) - \bm{\rho}^\top\bm{z}(\bm{x}_t , \thetastar) \leq \bm{\rho}^\top\bm{z}(\bm{x}_t^\star , \bm{\theta}_\beta) - \bm{\rho}^\top\bm{z}(\bm{x}_t , \bm{\theta}_\beta) + \epsilon_1(\beta , \bm{x}_t^\star , \lambda) + \epsilon_2(\beta, \bm{x}_t^\star , \lambda)  + \epsilon_1(\beta , \bm{x}_t , \lambda) + \epsilon_2(\beta, \bm{x}_t , \lambda)\]
Since $\bm{x}_t$ was not eliminated, we have $\textrm{UCB}(\beta , \bm{x}_t , \lambda) \geq \max\limits_{\bm{y}\in\mathcal{X}}\textrm{LCB}(\beta , \bm{y} , \lambda) \geq \textrm{LCB} (\beta , \bm{x}_t^\star , \lambda)$ since $\bm{x}_t^\star$ never gets eliminated (Lemma \ref{lemma: optimal arm remains}). Thus,
\[\bm{\rho}^\top\bm{z}(\bm{x}_t , \bm{\theta}_\beta) + \epsilon_1(\beta , \bm{x}_t , \lambda) + \epsilon_2(\beta, \bm{x}_t , \lambda) \geq  \bm{\rho}^\top\bm{z}(\bm{x}_t^\star , \bm{\theta}_\beta) - \epsilon_1(\beta , \bm{x}_t^\star , \lambda) - \epsilon_2(\beta, \bm{x}_t^\star , \lambda)\]
Thus, we get
\begin{align*}
    \bm{\rho}^\top\bm{z}(\bm{x}_t^\star , \thetastar) - \bm{\rho}^\top\bm{z}(\bm{x}_t , \thetastar) &\leq 2\epsilon_1(\beta , \bm{x}_t , \lambda) + 2\epsilon_2(\beta, \bm{x}_t , \lambda) + 2\epsilon_1(\beta , \bm{x}_t^\star , \lambda) + 2\epsilon_2(\beta, \bm{x}_t^\star , \lambda) 
    \\
    &\leq 4\max\limits_{\bm{x} \in \mathcal{X}}\epsilon_1(\beta , \bm{x} , \lambda) + 4\max\limits_{\bm{x} \in \mathcal{X}}\epsilon_2(\beta, \bm{x} , \lambda)
\end{align*}

Taking an expectation on both sides, we get
\begin{align*}
    &\E_{\mathcal{X}\sim\mathcal{D}_{\beta+1}}\sbrak{\bm{\rho}^\top\bm{z}(\bm{x}_t^\star , \thetastar) - \bm{\rho}^\top\bm{z}(\bm{x}_t , \thetastar)} 
    \leq 4 \left( \E_{\mathcal{X}\sim\mathcal{D}_{\beta+1}}\sbrak{\max\limits_{\bm{x} \in \mathcal{X}}\epsilon_1(\beta , \bm{x} , \lambda) + \max\limits_{\bm{x} \in \mathcal{X}}\epsilon_2(\beta, \bm{x} , \lambda)} \right)
    \\
    &\leq \frac{32RK \kappa^{1/2} d \gamma(\lambda)^2}{\tau_\beta}\cbrak{e^{3S} K^{3/2} S^{-1}\sqrt{\log (Kd) \log d} + 12 \kappa^{1/2}d} + \frac{16R K^2 d^{1/2}  \gamma(\lambda) \sqrt{\log (Kd)}}{\sqrt{\tau_\beta}} 
\end{align*}
which follows from Lemma \ref{lemma: expectation of eps1} and Lemma \ref{lemma: expectation of eps2}.

\end{proof}

\subsection{Supporting Results on Optimal Designs for \ref{section: regret_b_mnl}}
\label{appendix: optimal design}

Recall from Section \ref{appendix: b_mnl_notation}, \[\tilde{\bm{X}}_\beta = \frac{\bm{A}(\bm{x} , \hat{\bm{\theta}}_\beta)^\frac{1}{2}}{\sqrt{B_\beta(\bm{x})}}\otimes \bm{x}\]

Also, recall that at each round $t \in [T]$, the feasible set of context vectors $\mathcal{X}_t$ is being sampled from some distribution $\mathcal{D}$. For a given batch $\beta$, we denote $\mathcal{D}_\beta$ to be the distribution of the pruned arm-sets post the successive elimination procedure (Section \ref{subsection:successive_elimination}). Thus, we have that $\mathcal{D}_{\beta + 1} \subset \mathcal{D}_{\beta}$.

We now define $K$ different partitions of $\tilde{\bm{X}}_\beta$ as follows:
$$\tilde{\bm{x}}^{(i)}_\beta = \frac{\bm{A}(\bm{x} , \hat{\bm{\theta}}_\beta)^\frac{1}{2}}{\sqrt{B_\beta(\bm{x})}}\bm{e}_i\otimes \bm{x}$$
where $i \in [K]$ and $\bm{e}_i$ is the $K-$dimensional standard  basis vector. We first show a few relations between $\tilde{\bm{X}}_\beta$ and $\tilde{\bm{x}}^{(i)}_\beta$:

\begin{lemma}
Let $\tilde{\bm{X}}_\beta$ and $\tilde{\bm{x}}^{(i)}_\beta$ be defined as above. Then, we have 
    \[\tilde{\bm{X}}_\beta\tilde{\bm{X}}_\beta^\top = \sum\limits_{i=1}^{K}\tilde{\bm{x}}^{(i)}_\beta\tilde{\bm{x}}^{(i)\; \top}_\beta\]
    \label{lemma_OD: decomposition of x_tilde}
\end{lemma}
\begin{proof}
    \begin{align*}
    \sum\limits_{i=1}^{K} \tilde{\bm{x}}^{(i)}_\beta\tilde{\bm{x}}^{(i)\; \top}_\beta &= \sum\limits_{i=1}^{K}\pbrak{\frac{\bm{A}(\bm{x} , \hat{\bm{\theta}}_\beta)^\frac{1}{2}}{\sqrt{B_\beta(\bm{x})}}\bm{e}_i\otimes \bm{x}}\pbrak{\bm{e}_i^\top\frac{\bm{A}(\bm{x} , \hat{\bm{\theta}}_\beta)^\frac{1}{2}}{\sqrt{B_\beta(\bm{x})}}\otimes \bm{x}^\top}
    \\
    &= \frac{1}{B_\beta(\bm{x})}\sum\limits_{i=1}^{K}\bm{A}(\bm{x} , \hat{\bm{\theta}}_\beta)^\frac{1}{2}\bm{e}_i\bm{e}_i^\top\bm{A}(\bm{x} , \hat{\bm{\theta}}_\beta)^\frac{1}{2}\otimes \bm{x}\bm{x}^\top
    \\
    &= \frac{1}{B_\beta(\bm{x})}\bm{A}(\bm{x} , \hat{\bm{\theta}}_\beta)^\frac{1}{2}\pbrak{\sum\limits_{i=1}^{K}\bm{e}_i\bm{e}_i^\top}\bm{A}(\bm{x} , \hat{\bm{\theta}}_\beta)^\frac{1}{2}\otimes \bm{x}\bm{x}^\top
    \\
    &= \frac{\bm{A}(\bm{x} , \hat{\bm{\theta}}_\beta)}{B_\beta(\bm{x})} \otimes \bm{x}\bm{x}^\top
    = \tilde{\bm{X}}_\beta\tilde{\bm{X}}_\beta^\top
\end{align*}
where we use the fact that $\sum\limits_{i=1}^{K}\bm{e}_i\bm{e}_i^\top = \bm{I}_{K\times K}$.
\end{proof}

\begin{lemma}
    Let $\bm{M} \in \R^{Kd}$ be any positive-semidefinite matrix. Then,
    \[\eigmax{\tilde{\bm{X}}^\top_\beta \bm{M} \tilde{\bm{X}}_\beta} \leq \sum\limits_{i=1}^{K} \matnorm{\tilde{\bm{x}}^{(i)}_\beta}{\bm{M}}^2\]
    \label{lemma_OD: eig_relationship_between_x_tilde}
\end{lemma}
\begin{proof}
    \begin{align*}
    \eigmax{\tilde{\bm{X}}^\top_\beta \bm{M} \tilde{\bm{X}}_\beta} &\overset{(i)}{=} \eigmax{\tilde{\bm{X}}_\beta\tilde{\bm{X}}^\top_\beta \bm{M} }
    \overset{(ii)}{=} \eigmax{\sum\limits_{i=1}^{K} \tilde{\bm{x}}^{(i)}_\beta\tilde{\bm{x}}^{(i)\; \top}_\beta \bm{M} } \\
    &\overset{(iii)}{\leq}\sum\limits_{i=1}^{K} \eigmax{ \tilde{\bm{x}}^{(i)}_\beta\tilde{\bm{x}}^{(i)\; \top}_\beta \bm{M} }
    \overset{(iv)}{=}\sum\limits_{i=1}^{K} \eigmax{ \tilde{\bm{x}}^{(i)\; \top}_\beta \bm{M} \tilde{\bm{x}}^{(i)}_\beta}
    =\sum\limits_{i=1}^{K} \matnorm{\tilde{\bm{x}}^{(i)}_\beta}{\bm{M}}^2
\end{align*}

where $(i)$ follows from the cyclic property of eigenvalues, $(ii)$ follows from Lemma \ref{lemma_OD: decomposition of x_tilde}, $(iii)$ follows from the fact that $\eigmax{\bm{A} + \bm{B}}\leq \eigmax{\bm{A}} + \eigmax{\bm{B}}$, and $(iv)$ again follows from the cyclic property of eigenvalues.
\end{proof}

We first define the Distributional Optimal Design \citep{Ruan2021} for a set $\mathcal{X}$. 
\[\pi(\mathcal{X}) = 
    \begin{cases}
        \pi_G(\mathcal{X}) & \text{w.p.} \frac{1}{2}\\
        \pi^{S}_{\bm{M}_i}(\mathcal{X}) & \text{w.p.} \frac{p_i}{2}
    \end{cases}
\]
where $\pi_G$ is the G-optimal design and $\pi^{S}_{\bm{M_i}}$ represents the Softmax Policy with respect to $\bm{M}_i$. We refer the reader to \cite{Ruan2021} for more details.

We now define a few notations regarding some of the information and design matrices used throughout this section.

\begin{enumerate}
    \item $\bm{\mathbb{I}}^\lambda_\mathcal{D}(\pi) = \E\limits_{\mathcal{X}\sim\mathcal{D}}\sbrak{\E\limits_{\bm{x}\sim\pi(\mathcal{X})}\tilde{\bm{X}}_\beta\tilde{\bm{X}}_\beta^\top}$
    \item $\mathbb{W}^{(i)}_\mathcal{D}(\pi) = \E\limits_{\mathcal{X}\sim\mathcal{D}}\sbrak{\E\limits_{\bm{x}\sim\pi(\mathcal{X})}\tilde{\bm{x}}_\beta^{(i)}\tilde{\bm{x}}_\beta^{(i) \; \top}}$
     \item $\mathbb{W}^{(0)}_\mathcal{D}(\pi) =     \E\limits_{\mathcal{X}\sim\mathcal{D}}\sbrak{\E\limits_{\bm{x}\sim\pi(\mathcal{X})}\bm{x}\bm{x}^\top}$
\end{enumerate}

Suppose Algorithm \ref{alg: D-Opt_MNL} is called with the inputs $\beta$ and $\{\mathcal{X}_j\}_j$, where $\beta$ is the current batch index. Then, the policy returned by the algorithm is denoted by $\pi_\beta$, where

\[\pi_\beta = \frac{1}{K+1}\pbrak{\sum\limits_{i=0}^{K} \pi_{\beta,i}}\]

where $\pi_{\beta,0}$ and $\pi_{\beta,i} \; i \in [K]$ represents the Distributional Optimal Design learned over $\{\mathcal{X}_j\}_j$ and $F_i( \{\mathcal{X}_j\}_j, \beta), i \in [K]$ respectively. Here $F_i$ is as defined in Equation \ref{equation: scaled sets}. 


We now state a few results that relate the design matrices $\bm{H}$ and $\bm{V}$ as well as the matrices $\mathbb{I}$ and $\mathbb{W}$.

\begin{lemma}
    (Lemma A.16, \cite{sawarni24}) Let $\bm{V}_\beta$ and $\bm{H}_\beta$ as defined in Section \ref{appendix: b_mnl_notation} and $\bm{W}^{(0)}_{\mathcal{D}} (\pi_\beta)$ and $\bm{I}^\lambda_{\mathcal{D}}(\pi_\beta)$ be as defined in Section \ref{appendix: optimal design}. 
    
    Then, with probability at least $1 - \frac{1}{T^2}$, we have that
    \[\bm{V}_\beta \mgeq \frac{\tau_\beta}{8}\mathbb{W}^{(0)}_\mathcal{D}(\pi_\beta)\]
    \[\bm{H}_\beta \mgeq \frac{\tau_\beta}{8}\bm{\mathbb{I}}^\lambda_\mathcal{D}(\pi_\beta)\]
    \label{lemma_OD: relation between V,H,designs}
\end{lemma}

\begin{lemma} 
    For all $i \in [0,K]$, we have that
    \[(K+1)\bm{\mathbb{I}}^\lambda_\mathcal{D}(\pi_\beta) \mgeq \bm{\mathbb{I}}^\lambda_\mathcal{D}(\pi_{\beta,i}) \]
    \label{lemma_OD: relation between I_pi_beta and I_pi_i}
\end{lemma}
\begin{proof}
\begin{align*}
    \bm{\mathbb{I}}^\lambda_\mathcal{D}(\pi_\beta) &= \E\limits_{\mathcal{X}\sim\mathcal{D}}\sbrak{\E\limits_{\bm{x}\sim\pi_\beta(\mathcal{X})}\tilde{\bm{X}}_\beta \tilde{\bm{X}}^\top_\beta}
    \overset{(i)}{\mgeq} (K+1)\inv \E\limits_{\mathcal{X}\sim\mathcal{D}}\sbrak{\sum\limits_{i=0}^{K}\E\limits_{ \bm{x}\sim\pi_{\beta,i}(\mathcal{X})}\tilde{\bm{X}}_\beta \tilde{\bm{X}}^\top_\beta}\\
    &\mgeq (K+1)\inv \E\limits_{\mathcal{X}\sim\mathcal{D}}\sbrak{\E\limits_{ \bm{x}\sim\pi_{\beta,i}(\mathcal{X})}\tilde{\bm{X}}_\beta \tilde{\bm{X}}^\top_\beta}
    = (K+1)\inv \bm{\mathbb{I}}^\lambda_\mathcal{D}(\pi_{\beta,i})
\end{align*}

where $(i)$ follows from the definition of $\pi_\beta$.
\end{proof}

\begin{lemma} For all $i \in [K]$, we have that
    $$\bm{\mathbb{I}}^{\lambda}_\mathcal{D}(\pi) \mgeq \mathbb{W}^{(i)}_\mathcal{D}(\pi)$$
    \label{lemma_OD: relation_bw_W_and_I}
\end{lemma}
\begin{proof}
\begin{align*}
    \bm{\mathbb{I}}^{\lambda}_\mathcal{D}(\pi) &= \E\limits_{\mathcal{X}\sim\mathcal{D}}\sbrak{\E\limits_{\bm{x}\sim\pi(\mathcal{X})}\tilde{\bm{X}}_\beta\tilde{\bm{X}}_\beta^\top}
    \overset{(i)}{=} \E\limits_{\mathcal{X}\sim\mathcal{D}}\sbrak{\E\limits_{\bm{x}\sim\pi(\mathcal{X})}\sum\limits_{i=1}^{K} \tilde{\bm{x}}^{(i)}_\beta\tilde{\bm{x}}^{(i)\; \top}_\beta}
    \mgeq \E\limits_{\mathcal{X}\sim\mathcal{D}}\sbrak{\E\limits_{\bm{x}\sim\pi(\mathcal{X})}\tilde{\bm{x}}^{(i)}_\beta\tilde{\bm{x}}^{(i)\; \top}_\beta}
    = \mathbb{W}^{(i)}_\mathcal{D}(\pi)
\end{align*}
\end{proof}

Using the lemmas stated above, we now derive a few results.

\begin{lemma}
    Let $\bm{V}_\beta$ be as defined in Section \ref{appendix: b_mnl_notation} and $\tau_\beta$ be the length of the $\beta$ batch, i.e $\mathcal{T}_\beta = \tau_\beta$. Then, we have 
    \[ \E\limits_{\mathcal{X}\sim\mathcal{D}_{\beta+1}}\sbrak{\max\limits_{\bm{x}\in\mathcal{X}} \matnorm{\bm{x}}{\bm{V}_\beta\inv}^2} \leq \frac{16}{\tau_\beta}(K+1)d^2
    \]
    \label{lemma_OD: optimal_design_norm_squared_for_V}
\end{lemma}
\begin{proof}
\begin{align*}
    \E\limits_{\mathcal{X}\sim\mathcal{D}_{\beta+1}}\sbrak{\max\limits_{\bm{x}\in\mathcal{X}} \matnorm{\bm{x}}{\bm{V}_\beta\inv}^2} &\overset{(i)}{\leq} \frac{8}{\tau_\beta} \E\limits_{\mathcal{X}\sim\mathcal{D}_{\beta+1}}\sbrak{\max\limits_{\bm{x}\in\mathcal{X}} \matnorm{\bm{x}}{\mathbb{W}^{(0) \; -1}_{\mathcal{D}_{\beta}}(\pi_\beta)}^2}
    \\
    &\overset{(ii)}{\leq} \frac{8}{\tau_\beta} (K+1) \E\limits_{\mathcal{X}\sim\mathcal{D}_{\beta}}\sbrak{\max\limits_{\bm{x}\in\mathcal{X}} \matnorm{\bm{x}}{\mathbb{W}^{(0) \; -1}_{\mathcal{D}_{\beta}}(\pi_{\beta,0})}^2}
    \\
    &\overset{(iii)}{\leq} \frac{16}{\tau_\beta} (K+1)
    \E\limits_{\mathcal{X}\sim\mathcal{D}_{\beta}}\sbrak{\max\limits_{\bm{x}\in\mathcal{X}} \matnorm{\bm{x}}{\mathbb{W}^{(0) \; -1}_{\mathcal{D}_{\beta}}(\pi_G)}^2}\\
     &\overset{(iv)}{\leq} \frac{16}{\tau_\beta} (K+1) d^2
\end{align*}
where $(i)$ follows from Lemma \ref{lemma_OD: relation between V,H,designs}, $(ii)$ follows from Lemma \ref{lemma_OD: relation between I_pi_beta and I_pi_i} and the fact that $\mathcal{D}_{\beta+1} \subset \mathcal{D}_\beta$ and hence, $\E_{\mathcal{D}_{\beta+1}} \leq \E_{\mathcal{D}_\beta}$, $(iii)$ follows from the definition of $\pi_{\beta,0}$ and uses the fact that $\pi_{\beta,0} \mgeq \frac{\pi_G}{2}$, and $(iv)$ follows from Lemma \ref{Lemma_OD: opt_design_wrt_G}.

\end{proof}

\begin{lemma}
    Let $\bm{V}_\beta$ be as defined in Section \ref{appendix: b_mnl_notation} and $\tau_\beta$ be the length of the $\beta$ batch, i.e $|\mathcal{T}_\beta| = \tau_\beta$. Then, we have 
    \[ \E\limits_{\mathcal{X}\sim\mathcal{D}_{\beta+1}}\sbrak{\max\limits_{\bm{x} \in \mathcal{X}} \matnorm{\bm{x}}{\bm{V}_\beta\inv}} \leq 4\sqrt{\frac{Kd\log d}{\tau_\beta}} \]
    \label{lemma_OD: optimal_design_wrt_V}
\end{lemma}
\begin{proof}
\begin{align*}
    \E\limits_{\mathcal{X}\sim\mathcal{D}_{\beta+1}}\sbrak{\max\limits_{\bm{x} \in \mathcal{X}} \matnorm{\bm{x}}{\bm{V}_\beta\inv}} &\overset{(i)}{\leq} \sqrt{\frac{8}{\tau_\beta}} \E\limits_{\mathcal{X}\sim\mathcal{D}_{\beta}}\sbrak{\max\limits_{\bm{x} \in \mathcal{X}} \matnorm{\bm{x}}{\mathbb{W}^{(0) \; -1}_{\mathcal{D}_{\beta}}(\pi_\beta)}}
    \\
    &\overset{(ii)}{\leq} \sqrt{\frac{8}{\tau_\beta}(K+1)} \E\limits_{\mathcal{X}\sim\mathcal{D}_{\beta}}\sbrak{\max\limits_{\bm{x} \in \mathcal{X}} \matnorm{\bm{x}}{\mathbb{W}^{(0) \; -1}_{\mathcal{D}_{\beta}}(\pi_{\beta,0})}}
    \\
     &\overset{(iii)}{\leq} \sqrt{\frac{8}{\tau_\beta}(K+1)d\log d} 
     \\
     &\overset{}{\leq} 4\sqrt{\frac{Kd\log d}{\tau_\beta}} 
\end{align*}

where $(i)$ follows from Lemma \ref{lemma_OD: relation between V,H,designs} and the fact that $\mathcal{D}_{\beta+1} \subset \mathcal{D}_\beta$, $(ii)$ follows in a similar manner as Lemma \ref{lemma_OD: relation between I_pi_beta and I_pi_i}, and $(iii)$ follows from Lemma \ref{lemma_OD: opt_design_wrt_Distributional_design}.

\end{proof}

\begin{lemma}
Let $\tilde{\bm{X}}_\beta$ and $\bm{H}_\beta$ be as defined in Section \ref{appendix: b_mnl_notation}. Denote $\tau_\beta = |\mathcal{T}_\beta|$. Then, we have that
\[\E\limits_{\mathcal{X}\sim\mathcal{D}_{\beta+1}}\sbrak{\max\limits_{\bm{x}\in\mathcal{X}} \twonorm{\tilde{\bm{X}}^\top_\beta \bm{H}^{-1/2}_\beta}} \leq 4K^2\sqrt{\frac{d\log (Kd)}{\tau_\beta}}\]
\label{lemma_OD: optimal_design_on_eigmax}
\end{lemma}
\begin{proof}
\begin{align*}
    \E\limits_{\mathcal{X}\sim\mathcal{D}_{\beta+1}}\sbrak{\max\limits_{\bm{x}\in\mathcal{X}} \twonorm{\tilde{\bm{X}}^\top_\beta \bm{H}^{-1/2}_\beta}} &\overset{(i)}{\leq} \E\limits_{\mathcal{X}\sim\mathcal{D}_{\beta}}\sbrak{\max\limits_{\bm{x}\in\mathcal{X}} \sqrt{\eigmax{\bm{H}^{-1/2}_\beta\tilde{\bm{X}}_\beta \tilde{\bm{X}}^\top_\beta \bm{H}^{-1/2}_\beta}}}
    \\
    &\overset{(ii)}{=} \E\limits_{\mathcal{X}\sim\mathcal{D}_{\beta}}\sbrak{\max\limits_{\bm{x}\in\mathcal{X}} \sqrt{\eigmax{\tilde{\bm{X}}^\top_\beta \bm{H}\inv_\beta \tilde{\bm{X}}_\beta}}}
    \\
    &\overset{(iii)}{\leq} \E\limits_{\mathcal{X}\sim\mathcal{D}_{\beta}}\sbrak{\max\limits_{\bm{x}\in\mathcal{X}} \sqrt{\sum\limits_{i=1}^{K} \matnorm{\tilde{\bm{x}}_\beta^{(i)}}{\bm{H}\inv_\beta}^2}}
    \\
    &\overset{(iv)}{\leq} \sqrt\frac{8}{\tau_\beta} \E\limits_{\mathcal{X}\sim\mathcal{D}_{\beta}}\sbrak{\max\limits_{\bm{x}\in\mathcal{X}} \sqrt{\sum\limits_{i=1}^{K} \matnorm{\tilde{\bm{x}}_\beta^{(i)}}{\mathbb{I}^{\lambda \; -1}_{\mathcal{D}_{\beta}}(\pi_\beta)}^2}}
    \\
    &\overset{(v)}{\leq} \sqrt{\frac{8}{\tau_\beta}(K+1)}
    \E\limits_{\mathcal{X}\sim\mathcal{D}_{\beta}}\sbrak{\max\limits_{\bm{x}\in\mathcal{X}} \sqrt{\sum\limits_{i=1}^{K} \matnorm{\tilde{\bm{x}}_\beta^{(i)}}{\mathbb{I}^{\lambda \; -1}_{\mathcal{D}_{\beta}}(\pi_{\beta,i})}^2}}
    \\
    &\overset{(vi)}{\leq} \sqrt{\frac{8}{\tau_\beta}(K+1)}
    \E\limits_{\mathcal{X}\sim\mathcal{D}_{\beta}}\sbrak{\max\limits_{\bm{x}\in\mathcal{X}} \sqrt{\sum\limits_{i=1}^{K} \matnorm{\tilde{\bm{x}}_\beta^{(i)}}{\mathbb{W}^{(i) \; -1}_{\mathcal{D}_{\beta}}(\pi_{\beta,i})}^2}}
    \\
    &\overset{(vii)}{\leq} \sqrt{\frac{8}{\tau_\beta}(K+1)}
    \E\limits_{\mathcal{X}\sim\mathcal{D}_{\beta}}\sbrak{\max\limits_{\bm{x}\in\mathcal{X}}\sum\limits_{i=1}^{K} \matnorm{\tilde{\bm{x}}_\beta^{(i)}}{\mathbb{W}^{(i) \; -1}_{\mathcal{D}_{\beta}}(\pi_{\beta,i})}}\\
    &\overset{(viii)}{\leq} \sqrt{\frac{8}{\tau_\beta}(K+1)}
    \sum\limits_{i=1}^{K} \E\limits_{\mathcal{X}\sim\mathcal{D}_{\beta}}\sbrak{ \max\limits_{\bm{x}\in\mathcal{X}}\matnorm{\tilde{\bm{x}}_\beta^{(i)}}{\mathbb{W}^{(i) \; -1}_{\mathcal{D}_{\beta}}(\pi_{\beta,i})}}\\
    &\overset{(ix)}{\leq} K\sqrt{\frac{8}{\tau_\beta}(K+1)Kd\log (Kd)}
    \leq 4K^2\sqrt{\frac{d\log (Kd)}{\tau_\beta}}
\end{align*}

where $(i)$ follows from the definition of the norm $\twonorm{\bm{A}} = \sqrt{\eigmax{\bm{A}^\top\bm{A}}}$ and the fact that $\mathcal{D}_{\beta+1} \subset \mathcal{D}_\beta$, $(ii)$ follows from the cyclic property of eigenvalues, $(iii)$ follows from Lemma \ref{lemma_OD: eig_relationship_between_x_tilde}, $(iv)$ follows from Lemma \ref{lemma_OD: relation between V,H,designs}, $(v)$ follows from Lemma \ref{lemma_OD: relation between I_pi_beta and I_pi_i}, $(vi)$ follows from Lemma \ref{lemma_OD: relation_bw_W_and_I}, $(vii)$ uses the fact that for $\cbrak{a_i}_{i=1}^N$,  $\sqrt{\sum\limits_{i=1}^{N}a_i^2} \leq \sqrt{\pbrak{\sum\limits_{i=1}^{N}a_i}^2} = \sum\limits_{i=1}^{N}a_i$, $(viii)$ uses the linearity of expectations and the fact that $\max\limits_x [f(x) + g(x)] \leq \max\limits_x f(x) + \max\limits_x g(x)$, and $(ix)$ follows from Lemma \ref{lemma_OD: opt_design_wrt_Distributional_design}.
 \end{proof}

\section{Rarely Switching Multinomial Contextual Bandit Algorithm: \texttt{RS-MNL}}
\label{section: regret_rs_mnl}


\subsection{Notations}
\label{appendix: rs_mnl_notation}
We first define a few matrices, vectors, and scalars that are used throughout this section (here, $\bm{e}_i$ denotes the $i^{th}-$standard basis vector):
\begin{enumerate}
    \item $\bm{V}_t = \lambda\bm{I}_{d \times d} + \sum\limits_{s\in[t]} \bm{x}_s \bm{x}_s^\top $
    \item $\tilde{\bm{V}}_t = \bm{I}_{K \times K} \otimes \bm{V}_t$
    \item $\bm{A}(\bm{x} , \bm{\theta}) = \textrm{diag}(\bm{z}(\bm{x} , \bm{\theta})) - \bm{z}(\bm{x} , \bm{\theta}) \bm{z}(\bm{x} , \bm{\theta})^\top$
    \item $\bm{M}(\bm{x} , \bm{\theta}_1 , \bm{\theta}_2) = \int\limits_{0}^1 \bm{A}(\bm{x} , v\bm{\theta}_1 + (1-v)\bm{\theta}_2) \; dv$
    \item $\bm{H}^\star_t = \lambda \bm{I}_{Kd \times Kd} + \sum\limits_{s \in [t]} \bm{A}(\bm{x}_s, \bm{\theta}^\star)\otimes \bm{x}_s \bm{x}_t^\top$
    \item $\gamma(\delta) = C S^{5/4} \sqrt{Kd \log (T/\delta)}$
    \item $B_t(\bm{x}) = \exp\pbrak{\sqrt{6}\min\cbrak{2\kappa^{1/2}\gamma(\delta) \lVert \bm{x} \rVert_{\bm{V}_t^{-1}}, 2S}}$
    \item $\bm{H}_t (\bm\theta) = \lambda\bm{I}_{Kd \times Kd} + \sum\limits_{s \in [t]} \frac{\bm{A}(\bm{x}_s , \bm\theta)}{B_s(\bm{x}_s)} \otimes \bm{x}_s\bm{x}_s^\top$
    \item $\tilde{\bm{X}_t}(\bm\theta) = \frac{\bm{A}(\bm{x}_t , \bm\theta)^{\frac{1}{2}}}{\sqrt{B_t(\bm{x}_t)}} \otimes \bm{x}_t$
    \item $\tilde{\bm{x}}_t^{(i)}(\bm\theta) = \frac{\bm{A}(\bm{x}_t , \bm\theta)^{\frac{1}{2}}}{\sqrt{B_t(\bm{x}_t)}} \bm{e}_i \otimes \bm{x}_t$
    \item $\bm{H}^{i}_t(\bm\theta) = \sum\limits_{s \in [t]} \tilde{\bm{x}}^{(i)}_s (\bm\theta) \tilde{\bm{x}}^{(i) }_s (\bm\theta)^\top + \lambda\bm{I}$
\end{enumerate}

We now present the regret upper bound for \texttt{RS-MNL} by restating Theorem \ref{thm: regret_rs_mnl}.

\begin{theorem}
\label{theorem RS-MNL}
    With high probability, the regret incurred by Algorithm \ref{alg : RS-MNL} is bounded above by $R_T$ where:
    \[R_T \leq C R K^{3/2} S^{5/4} (\log T \log (T/\delta))^{1/2} d  \sqrt{T} + C R K^2 d^2 S^{5/2} \log T \log(T/\delta) \kappa^{1/2} e^{2S}(e^S + K \kappa^{1/2})\]
    
\end{theorem}
\begin{proof} For any round $t \in [T]$, let $\tau_t \leq t$ denote the last round at which a switch was made. Then, using the value of $\gamma(\delta)$ alongside Lemma \ref{Lemma_RS: per_round_regret}, Lemma \ref{lemma_RS: bound on epsilon_1}, and Lemma \ref{lemma_RS: bound on epsilon_2}, we get:
\begin{align*}
    R(T) &\leq \sum_{t \in [T]} \lvert \bm{\rho}^\top \bm{z} (\bm{x}_t^\star , \bm\theta^\star) - \bm\rho^\top \bm{z}(\bm{x}_t , \bm\theta^\star)  \rvert \leq \sum\limits_{t\in[T]} {2\epsilon_1(t , \tau_t , \bm{x}_t ) + 2\epsilon_2(t , \tau_t , \bm{x}_t )}
    \\ 
    &\leq   4 R K d^{1/2} (\log T)^{1/2} \gamma(\delta)\sqrt{T} + 8 R K d \log T \kappa^{1/2} e^{3S} \gamma(\delta)^2+ 24 d R K^2 e^{2S} \kappa \gamma(\delta)^2 \log T  
    \\
    &\leq C R K^{3/2} S^{5/4} (\log T \log (T/\delta))^{1/2} d  \sqrt{T} + C R K^2 d^2 S^{5/2} \log T \log(T/\delta) \kappa^{1/2} e^{2S}(e^S + K \kappa^{1/2})
\end{align*}
\end{proof}
\subsection{Supporting Lemmas for \ref{section: regret_rs_mnl}}

\begin{lemma}
    Let $\{\bm{x}_1 , \ldots , \bm{x}_\tau\}$ be a set of arms and $\{r_1 , \ldots , r_\tau\}$ be the set of corresponding rewards associated with the arms. Define $\hat{\bm\theta}_\tau$ be the MLE estimate calculated using this set of arms and rewards, i.e
    \[\hat{\bm\theta}_\tau = \argmin_{\bm\theta} \sum_{s \in [\tau]} \sum_{i=1}^K \mathbbm{1}\{y_s = i\} \log z_i(\bm{x}_s , \bm\theta) + \frac{\lambda}{2} \lVert \bm\theta \rVert^2_2\]
    Let $\bm{H}^\star_\tau$ be as defined in Section \ref{appendix: rs_mnl_notation}. Then, with high probability, and the choice of $\lambda = K d S^{-1/2} \log(T/\delta)$, we have that
    \[\lVert \hat{\bm\theta}_\tau - \bm\theta^\star \rVert_{\bm{H}_\tau^\star} \leq C S^{5/4} \sqrt{Kd \log (T/\delta)}\]
    \label{lemma: RS-Confidence bound}
\end{lemma}

\begin{proof}
We define $G_\tau(\bm\theta_1 , \bm\theta_2)$ as:
\[G_\tau(\bm\theta_1 , \bm\theta_2) = \sum_{t \in [\tau]} M(\bm{x}_t , \bm\theta_1 , \bm\theta_2) \otimes \bm{x}_t \bm{x}_t^\top + \lambda \bm{I}\]

where $M(\bm{x} , \bm\theta_1 , \bm\theta_2)$ is as defined in Section \ref{appendix: rs_mnl_notation}. Thus, from Lemma \ref{lemma: H-G relation}, we have that 
\[(1+2S)\inv \bm{G}_\tau \mgeq \bm{H}^\star_\tau\]

Thus, we have
\begin{align*}
   \matnorm{\hat{\bm{\theta}}_\tau - \thetastar}{\bm{H}_\tau^\star} &\leq \sqrt{1+2S} \matnorm{\hat{\bm{\theta}}_\tau - \thetastar}{\bm{G}_\tau(\hat{\bm{\theta}}_\tau , \thetastar)}
   \\
    &\leq \sqrt{1+2S} \matnorm{\bm{G}_\tau(\hat{\bm{\theta}}_\tau , \thetastar)\pbrak{\hat{\bm{\theta}}_\tau - \thetastar}}{\bm{G}_\tau\inv(\hat{\bm{\theta}}_\tau , \thetastar)}
    \\
    &\overset{(i)}{\leq} \sqrt{1+2S} \matnorm{\sbrak{\sum\limits_{t \in [\tau]} \bm{M}(\hat{\bm{\theta}}_\tau , \thetastar) \otimes \bm{x}_t\bm{x}_t^\top + \lambda \bm{I}_{Kd \times Kd}}\pbrak{\hat{\bm{\theta}}_\tau - \thetastar}}{\bm{G}_\tau\inv(\hat{\bm{\theta}}_\tau , \thetastar)}
    \\
    &\overset{(ii)}{\leq} \sqrt{1 + 2S}\matnorm{\sum\limits_{t \in [\tau]} \sbrak{\bm{M}(\bm{x} , \thetastar , \hat{\bm{\theta}}_\tau) \otimes \bm{x}_t^\top}\pbrak{\thetastar - \hat{\bm{\theta}}_\tau} \otimes \bm{x}_t + \lambda \pbrak{\hat{\bm{\theta}}_\tau - \thetastar}}{\bm{G}^{-1}_\tau(\bm{\theta}_1 , \bm{\theta}_2)}
    \\
    &\overset{(iii)}{\leq} \sqrt{1 + 2S}\matnorm{\sum\limits_{t \in [\tau]} \sbrak{\bm{z}(\bm{x_t} , \hat{\bm{\theta}}_\tau) - \bm{z}(\bm{x}_t  , \thetastar)} \otimes \bm{x}_t + \lambda \pbrak{\hat{\bm{\theta}}_\tau - \thetastar}}{\bm{G}^{-1}_\tau(\bm{\theta}_1 , \bm{\theta}_2)}
    \\ 
    &\overset{(iv)}{\leq} \sqrt{1 + 2S}\matnorm{\sum\limits_{t \in [\tau]} \sbrak{\bm{m}_t - \bm{z}(\bm{x}_t  , \thetastar)} \otimes \bm{x}_t - \lambda \thetastar}{\bm{G}^{-1}_\tau(\bm{\theta}_1 , \bm{\theta}_2)}
    \\ 
    &\overset{(v)}{\leq} \sqrt{1 + 2S}\matnorm{\sum\limits_{t \in [\tau]} \sbrak{\bm{m}_t - \bm{z}(\bm{x}_t  , \thetastar)} \otimes \bm{x}_t}{\bm{G}^{-1}_\tau(\bm{\theta}_1 , \bm{\theta}_2)}  +\lambda \sqrt{1+2S} \matnorm{\thetastar}{\bm{G}^{-1}_\tau(\bm{\theta}_1 , \bm{\theta}_2)}
    \\ 
    &\leq (1 + 2S)\matnorm{\sum\limits_{t \in [\tau]} \sbrak{\bm{m}_t - \bm{z}(\bm{x}_t  , \thetastar)} \otimes \bm{x}_t}{\bm{H}_\tau^{\star \; -1}}  +\lambda \sqrt{1+2S} \matnorm{\thetastar}{\bm{G}^{-1}_\tau(\bm{\theta}_1 , \bm{\theta}_2)}
    \\ 
    &\overset{(vi)}{\leq} (1 + 2S)\matnorm{\sum\limits_{t \in [\tau]} \bm{\epsilon}_t \otimes \bm{x}_t}{\bm{H}_\tau^{\star \; -1}}  + S\sqrt{1+2S}\sqrt{\lambda}
    \\ 
    &\overset{(vi)}{\leq} 3S\matnorm{\sum\limits_{t \in [\tau]} \bm{\epsilon}_t \otimes \bm{x}_t}{\bm{H}_\tau^{\star \; -1}}  + \sqrt{3} \lambda^{1/2} S^{3/2} 
\end{align*}

where $(i)$ follows from Lemma \ref{lemma: H-G relation} , $(ii)$ follows from Mixed Product Property , $(iii)$ follows from the Mean value Theorem, $(iv)$ from Lemma \ref{lemma: MLE}, $(v)$ follows from Cauchy-Schwarz, and $(vi)$ follows from the fact that $\bm{G}_\tau \geq \lambda\bm{I}$ and $\twonorm{\bm{\theta}} \leq S$.

Note that $\bm{\epsilon}_t = \bm{m}_t - \bm{z}(\bm{x}_t , \thetastar)$ and since $\E\sbrak{\bm{m}_t} = \bm{z}(\bm{x}_t,\thetastar)$, we get $\E\sbrak{\bm{\epsilon}_t\bm{\epsilon}_t^\top} = \bm{A}(\bm{x}_t , \thetastar)$. Also, note that $\lVert \epsilon_t \rVert_1 \leq \lVert \bm{m}_t \rVert_1 + \lVert\bm{z}(\bm{x}_t , \thetastar) \rVert_1 \leq 2$. Thus, using Lemma \ref{lemma: martingale confidence bound}, we get

\[\matnorm{\hat{\bm{\theta}}_\tau - \thetastar}{\bm{H}_\tau^\star} \leq 3S\pbrak{\frac{\sqrt\lambda}{4} + \frac{4}{\sqrt\lambda}\log\pbrak{\frac{\textrm{det } \bm{H}_\tau^{1/2}}{\delta\lambda^{\frac{dK}{2}}}} + \frac{4}{\sqrt\lambda} Kd\log 2} + 2S^{3/2}\lambda^{1/2}\]

where $\bm{H}_\tau = \lambda \bm{I} + \sum_{t \in \tau} A(\bm{x}_t , \bm\theta^\star) \otimes \bm{x}_t \bm{x}_t^\top$.

We can calculate $\textrm{det }\bm{H}_\tau$ as follows:
\begin{align*}
    \textrm{det }\bm{H}_\tau &\overset{(i)}{\leq} \left( \frac{\textrm{trace } \bm{H}_\tau}{Kd} \right)^{Kd} 
    \\
    &\leq \left( \frac{\textrm{trace } \lambda \bm{I} + \textrm{trace } \sum_{t \in \tau} A(\bm{x}_t , \bm\theta^\star) \otimes \bm{x}_t \bm{x}_t^\top}{Kd} \right)^{Kd}
    \\
    &\overset{(ii)}{\leq} \left( \frac{\lambda Kd + \tau \lVert \bm{x}_t \rVert_2^2}{Kd} \right)^{Kd}
    \\
    &\overset{(iii)}{\leq} \lambda^{Kd} \left(1 +  \frac{\tau}{\lambda Kd} \right)^{Kd}
\end{align*}
where $(i)$ follows from Lemma \ref{lemma: determinant-trace inequality}, $(ii)$ follows from the fact that $\textrm{tr }(\bm{A}\otimes\bm{B}) = \sum \lambda(\bm{A}) \lambda(\bm{B})$ and the fact that $\bm{A}(\bm{x} , \bm\theta^\star) \mleq \bm{I}$ and the only non-zero eigenvalue of $\bm{x}_t\bm{x}_t^\top$ is $\lVert \bm{x}_t \rVert_2^2$, and $(iii)$ follows since $ \lVert \bm{x}\rVert \leq 1$. 

Thus, we have 
\begin{align*}
    \matnorm{\hat{\bm{\theta}}_\tau - \thetastar}{\bm{H}_\tau^\star} &\leq 3S\pbrak{\frac{\sqrt\lambda}{4} + \frac{4}{\sqrt\lambda}\log\pbrak{\frac{\pbrak{1 + \frac{\tau}{\lambda Kd}}^\frac{Kd}{2}}{\delta}} + \frac{4}{\sqrt\lambda}Kd\log 2} + 2S^{3/2} \lambda^{1/2}
    \\
    &= 3S\pbrak{\frac{\sqrt\lambda}{4} + \frac{2Kd}{\sqrt\lambda} \log\pbrak{1 + \frac{\tau}{\lambda d}} + \frac{4}{\sqrt\lambda}\log\frac{1}{\delta} + \frac{4}{\sqrt\lambda}Kd\log 2} + 2S^{3/2} \lambda^{1/2}
\end{align*}

Finally, by setting $\lambda = K d S^{-1/2} \log (T/\delta)$ and simplifying the constants, we get that for some appropriately tuned constant $C$
\[\matnorm{\hat{\bm{\theta}}_\tau - \thetastar}{\bm{H}_\tau^\star} \leq CS^{5/4}\sqrt{Kd\log(T/\delta)}\]
\end{proof}

From here on, we shall use the notation $\gamma(\delta) = C S^{5/4} \sqrt{Kd \log (T/\delta)}$.

\begin{lemma}
    Let $\tilde{\bm{V}}_t$ and $\bm{H}_t^\star$ be defined as in Section \ref{appendix: rs_mnl_notation}. Then, for any round $t \in [T]$, we have that
    \[\tilde{\bm{V}}_t \mleq \kappa \bm{H}_t^\star\]
    \label{lemma_RS: relation_between_V_and_H}
\end{lemma}
\begin{proof}
   From the definition of $\kappa$, we have $\bm{A}(\bm{x} , \bm\theta) \mgeq \frac{1}{\kappa} \bm{I}$.  Hence, using the fact that $\kappa \geq 1$, we have 
   \begin{align*}
       \tilde{\bm{V}}_t &= \bm{I}_{K \times K} \otimes \bm{V}_t
    =\bm{I}_{K \times K} \otimes \pbrak{\lambda\bm{I}_{d\times d} + \sum\limits_{s \in [t]}\bm{x}_s \bm{x}_s^\top}
    \\
    &= \lambda\bm{I}_{Kd \times Kd} +  \bm{I}_{K\times K}\otimes \sum\limits_{s \in [t]} \bm{x}_t \bm{x}_t^\top 
    \\
    &\mleq\kappa \lambda \bm{I}_{Kd \times Kd} + \kappa \sum\limits_{s \in [t]} \bm{A}(\bm{x}_t , \thetastar) \otimes \bm{x}_t \bm{x}_t ^\top 
    \\
    &\mleq \kappa \bm{H}^\star_t 
   \end{align*}
\end{proof}

\begin{lemma}
    Let $\tilde{\bm{V}}_t$ and $\bm{H}_t(\bm\theta)$ be defined as in Section \ref{appendix: rs_mnl_notation}. Then, for any round $t \in [T]$, we have that
    \[\tilde{\bm{V}}_t \mleq \kappa \bm{H}_t(\bm\theta)\]
    \label{lemma_RS: relation_between_V_and_H_theta}
\end{lemma}
\begin{proof}
   From the definition of $\kappa$, we have $\bm{A}(\bm{x} , \bm\theta) \mgeq \frac{1}{\kappa} \bm{I}$.  Hence, using the fact that $\kappa \geq 1$, we have 
   \begin{align*}
       \tilde{\bm{V}}_t &= \bm{I}_{K \times K} \otimes \bm{V}_t
    =\bm{I}_{K \times K} \otimes \pbrak{\lambda\bm{I}_{d\times d} + \sum\limits_{s \in [t]}\bm{x}_s \bm{x}_s^\top}
    \\
    &= \lambda\bm{I}_{Kd \times Kd} +  \bm{I}_{K\times K}\otimes \sum\limits_{s \in [t]} \bm{x}_t \bm{x}_t^\top 
    \\
    &\mleq\kappa \lambda \bm{I}_{Kd \times Kd} + \kappa \sum\limits_{s \in [t]} \bm{A}(\bm{x}_t , \bm\theta) \otimes \bm{x}_t \bm{x}_t ^\top 
    \\
    &\mleq \kappa \lambda \bm{I}_{Kd \times Kd} + \kappa \sum\limits_{s \in [t]} \frac{\bm{A}(\bm{x}_t , \bm\theta)}{B_s(\bm{x}_s)} \otimes \bm{x}_t \bm{x}_t ^\top 
    \\
    &\mleq \kappa \bm{H}_t(\bm\theta)
   \end{align*}
   where the second to last inequality follows since $B_t(\bm{x}) \geq 1$.
\end{proof}

\begin{lemma}
    Let $1 , \tau_1 , \ldots, \tau_m$ be the rounds at which a switch occurs, i.e $\textrm{det }\bm{H}_{\tau_{i+1}}(\hat{\bm\theta}_{\tau_i}) \geq 2\textrm{ det } \bm{H}_{\tau_{i}}(\hat{\bm\theta}_{\tau_i}) \forall i \in [m]$. Let $\bm{H}_t(\bm\theta)$ and $\bm{H}_t^\star$ be defined as in Section \ref{appendix: rs_mnl_notation}. Then, for all $i \in [m]$, we have that
    \[\bm{H}_{\tau_i}(\hat{\bm\theta}_{\tau_{i}}) \mleq \bm{H}^\star_{\tau_i}\]
    \label{lemma_RS: relation between H and H for switch rounds}
\end{lemma}
\begin{proof} 

From Lemma \ref{lemma: self-concordance}, for some $\bm{x}$ such that $\lVert \bm{x} \rVert \leq 1$ and some $\tau \in \{\tau_1 , \ldots , \tau_m\}$, we have that
\[A(\bm{x} , \hat{\bm\theta}_{\tau}) \mleq A(\bm{x} , \bm\theta^\star) \exp \left(\sqrt{6} \left \lVert (\bm{I} \otimes \bm{x}^\top)(\bm\theta^\star - \hat{\bm\theta}_{\tau}) \right \rVert_2 \right)\]

Now, we can bound $\left \lVert (\bm{I} \otimes \bm{x}^\top)(\bm\theta^\star - \hat{\bm\theta}_{\tau}) \right \rVert_2$ as follows:
\begin{align*}
    \left \lVert (\bm{I} \otimes \bm{x}^\top)(\bm\theta^\star - \hat{\bm\theta}_{\tau_t}) \right \rVert_2 &\overset{(i)}{\leq} 2S  \lVert (\bm{I} \otimes \bm{x}^\top) \rVert_2 
    \overset{(ii)}{=} 2S \sqrt{\eigmax{(\bm{I} \otimes \bm{x})(\bm{I} \otimes \bm{x}^\top)}} 
    \\
    &\overset{(iii)}{=} 2S \sqrt{\eigmax{\bm{I} \otimes \bm{x}\bm{x}^\top}}
    \overset{(iv)}{\leq} 2S
\end{align*}
where $(i)$ uses Cauchy-Schwarz inequality and the fact that $\lVert \bm\theta \rVert_2 \leq S$, $(ii)$ uses the definition of the norm as $\lVert \bm{A} \rVert_2 = \sqrt{\eigmax{\bm{A}^\top \bm{A}}}$, $(iii)$ follows from the mixed product property of tensor products, and $(iv)$ follows from the fact that $\eigmax{\bm{A} \otimes \bm{B}} = \eigmax{\bm{A}}\eigmax{\bm{B}}$ and  $\eigmax{\bm{x}\bm{x}^\top} = \lVert \bm{x} \rVert_2^2 \leq 1$.

We can also bound $\left \lVert (\bm{I} \otimes \bm{x}^\top)(\bm\theta^\star - \hat{\bm\theta}_{\tau}) \right \rVert_2$ in the following way (note that the $d-$ dimensional unit ball is represented as $\mathcal{B}_2(d)$):
\begin{align*}
    \lVert (\bm{I} \otimes \bm{x}^\top) (\bm\theta^\star - \hat{\bm\theta}_\tau) \rVert_2 &= \lVert (\bm{I} \otimes \bm{x}^\top) {\bm{H}_\tau^\star}^{-1/2} {\bm{H}_\tau^\star}^{1/2}  (\bm\theta^\star - \hat{\bm\theta}_\tau) \rVert_2
    \\
    &\overset{(i)}{\leq} \lVert (\bm{I} \otimes \bm{x}^\top) {\bm{H}_\tau^\star}^{-1/2}\rVert_2 \gamma(\delta)
    \\
    &\overset{(ii)}{\leq} \kappa^{1/2} \lVert (\bm{I} \otimes \bm{x}^\top) {\tilde{\bm{V}}_\tau}^{-1/2}\rVert_2 \gamma(\delta)
    \\
    &\overset{(iii)}{\leq} \kappa^{1/2} \lVert (\bm{I} \otimes \bm{x}^\top) (\bm{I} \otimes \bm{V}_\tau^{-1/2}) \rVert_2 \gamma(\delta)
    \\
    &\overset{(iv)}{=} \kappa^{1/2} \sqrt{\eigmax{(\bm{I} \otimes \bm{V}_\tau^{-1/2}) (I \otimes \bm{x}) (\bm{I} \otimes \bm{x}^\top) (\bm{I} \otimes \bm{V}_\tau^{-1/2})}}\gamma(\delta)
    \\
    &\overset{(v)}{=} \kappa^{1/2} \gamma(\delta) \lvert \bm{x} \rVert_{\bm{V}_\tau^{-1}}
    \\
    &\leq 2\kappa^{1/2}\gamma(\delta) \lvert \bm{x} \rVert_{\bm{V}_\tau^{-1}}
\end{align*}
where $(i)$ is obtained from the fact that $\lVert \bm{A} \bm{x} \rVert_2 \leq \lVert \bm{A} \rVert_2 \lVert \bm{x} \rVert_2$ and from Lemma \ref{lemma: RS-Confidence bound}, $(ii)$ follows from Lemma \ref{lemma_RS: relation_between_V_and_H}, $(iii)$ is obtained from the definition of $\tilde{\bm{V}}$ and the fact that $(\bm{A} \otimes \bm{B})^n = \bm{A}^n \otimes \bm{B}^n$, $(iv)$ follows from the definition of the norm, i.e, $\lVert \bm{A} \rVert_2 = \sqrt{\eigmax{\bm{A}^\top \bm{A}}}$, and $(v)$ follows from the cyclic property of eigenvalues and the fact that $\eigmax{\bm{A} \otimes \bm{B}} = \eigmax{\bm{A}} \eigmax{\bm{B}}$.

Thus, by combining both bounds, we obtain

\[A(\bm{x} , \hat{\bm\theta}_{\tau}) \mleq A(\bm{x} , \bm\theta^\star) \exp \left(\sqrt{6} \min \left\{  \sqrt{2}\kappa^{1/2}\gamma(\delta) \lvert \bm{x} \rVert_{\bm{V}_\tau^{-1}}, 2S \right\} \right) \]

Let $B_\tau(\bm{x})$ denote the value $ \exp \left(\sqrt{6} \min \left\{  \sqrt{2}\kappa^{1/2}\gamma(\delta) \lvert \bm{x} \rVert_{\bm{V}_\tau^{-1}}, 2S \right\} \right)$. Then, we have that

\[\bm{H}_{\tau}^\star = \lambda \bm{I} + \sum_{s \in [\tau]} \bm{A}(\bm{x}_s , \bm\theta^\star) \otimes \bm{x}_s \bm{x}_s \mgeq \lambda \bm{I} + \sum_{s \in [\tau]} \frac{\bm{A}(\bm{x}_s , \hat{\bm\theta}_{\tau})}{B_\tau(\bm{x}_s)} \otimes \bm{x}_s \bm{x}_s = \bm{H}_\tau(\hat{\bm\theta}_{\tau})\]

\end{proof}

\begin{lemma}
    For time round $t$, let $\tau_t \leq t$ be the last time round at which a switch occurred, i.e $\textrm{det } \bm{H}_t (\hat{\bm\theta}_{\tau_t}) \leq 2 \textrm{ det }\bm{H}_{\tau_t}(\hat{\bm\theta}_{\tau_t})$. Let $\bm{H}_t(\bm\theta)$ and $\bm{H}_t^\star$ be defined as in Section \ref{appendix: rs_mnl_notation}.
    \[\bm{H}_t(\hat{\bm\theta}_{\tau_t}) \mleq \bm{H}^\star_t\]
    \label{lemma_RS: relation between H and H}
\end{lemma}
\begin{proof} 

Similar to Lemma \ref{lemma_RS: relation between H and H for switch rounds} for some $\bm{x}$ such that $\lVert \bm{x} \rVert \leq 1$, we have that
\[A(\bm{x} , \hat{\bm\theta}_{\tau_t}) \mleq A(\bm{x} , \bm\theta^\star) \exp \left(\sqrt{6} \left \lVert (\bm{I} \otimes \bm{x}^\top)(\bm\theta^\star - \hat{\bm\theta}_{\tau_t}) \right \rVert_2 \right)\]

Now, we can bound $\left \lVert (\bm{I} \otimes \bm{x}^\top)(\bm\theta^\star - \hat{\bm\theta}_{\tau_t}) \right \rVert_2$ in two different ways: the first way results in $2S$, following the same method as Lemma \ref{lemma_RS: relation between H and H for switch rounds}. We can also bound it in the following way:

\begin{align*}
    \lVert (\bm{I} \otimes \bm{x}^\top) (\bm\theta^\star - \hat{\bm\theta}_{\tau_t}) \rVert_2 &= \lVert (\bm{I} \otimes \bm{x}^\top) {\bm{H}_{\tau_t}^\star}^{-1/2} {\bm{H}_{\tau_t}^\star}^{1/2}  (\bm\theta^\star - \hat{\bm\theta}_{\tau_t}) \rVert_2
    \\
    &\overset{(i)}{\leq} \lVert (\bm{I} \otimes \bm{x}^\top) {\bm{H}_{\tau_t}^\star}^{-1/2}\rVert_2 \gamma(\delta)
    \\
    &\overset{(ii)}{\leq} \lVert (\bm{I} \otimes \bm{x}^\top) \bm{H}_{\tau_t}(\hat{\bm\theta}_{\tau_t})^{-1/2}\rVert_2 \gamma(\delta)
    \\
    &\overset{(iii)}{\leq} \sqrt{2} \lVert (\bm{I} \otimes \bm{x}^\top) \bm{H}_t(\hat{\bm\theta}_{\tau_t})^{-1/2}\rVert_2 \gamma(\delta)
    \\
    &\overset{(iv)}{\leq} \sqrt{2} \kappa^{1/2} \lVert (\bm{I} \otimes \bm{x}^\top) \tilde{\bm{V}}^{-1/2} \rVert_2 \gamma(\delta)
    \\
    &\leq 2\kappa^{1/2}\gamma(\delta) \lvert \bm{x} \rVert_{\bm{V}_t^{-1}}
\end{align*}
where $(i)$ is obtained from the fact that $\lVert \bm{A} \bm{x} \rVert_2 \leq \lVert \bm{A} \rVert_2 \lVert \bm{x} \rVert_2$ and from Lemma \ref{lemma: RS-Confidence bound}, $(ii)$ follows from Lemma \ref{lemma_RS: relation between H and H for switch rounds}, $(iii)$ follows from the combination of Lemma \ref{lemma: determinant_condition} and the fact that $\textrm{det } \bm{H}_t (\hat{\bm\theta}_{\tau_t}) \leq 2 \textrm{ det }\bm{H}_{\tau_t}(\hat{\bm\theta}_{\tau_t})$, $(iv)$ follows from Lemma \ref{lemma_RS: relation_between_V_and_H_theta}, and $(v)$ follows from the same steps used in Lemma \ref{lemma_Rs: relation between H and H for x_tilde}.

Combining the bounds in the same manner as Lemma \ref{lemma_RS: relation between H and H for switch rounds} finishes the proof.

\end{proof}

\begin{lemma} 
For time round $t$, let $\tau_t \leq t$ be the last time round at which a switch occurred. Let $\bm{H}^{(i)}_t(\hat{\bm\theta}_{\tau_t})$ and $\bm{H}_t(\hat{\bm\theta}_{\tau_t})$ be defined as in Section \ref{appendix: rs_mnl_notation}. Then, we have
    \[\bm{H}^{(i)}_t(\hat{\bm\theta}_{\tau_t}) \mleq \bm{H}_t(\hat{\bm\theta}_{\tau_t})\]
    \label{lemma_Rs: relation between H and H for x_tilde}
\end{lemma}
\begin{proof} We have:
\begin{align*}
    \bm{H}_t(\hat{\bm\theta}_{\tau_t}) &= \lambda\bm{I} + \sum\limits_{s \in [t]}\tilde{\bm{X}}_s(\hat{\bm\theta}_{\tau_t}) \tilde{\bm{X}}_s(\hat{\bm\theta}_{\tau_t})^\top 
    \\
    &\overset{(i)}{=} \lambda \bm{I} +  \sum\limits_{s \in [t]}\sum\limits_{i=1}^{K}\tilde{\bm{x}}^{(i)}_s(\hat{\bm\theta}_{\tau_t}) \tilde{\bm{x}}^{(i)}_s (\hat{\bm\theta}_{\tau_t})^\top
    \\
    &\mgeq \lambda \bm{I} + \sum\limits_{s \in [t]}\tilde{\bm{x}}^{(i)}_s (\hat{\bm\theta}_{\tau_t}) \tilde{\bm{x}}^{(i)}_s (\hat{\bm\theta}_{\tau_t})^\top
    \\
    &= \bm{H}^i_\tau(\hat{\bm\theta}_{\tau_t})
\end{align*}
where $(i)$ follows from Lemma \ref{lemma_OD: decomposition of x_tilde}. 
\end{proof}

\begin{lemma}
    Let $\tau_t \leq t$ be the last time round at which a switch was made. In other words, $\textrm{det }\bm{H}_t(\hat{\bm\theta}_{\tau_t}) \leq 2\textrm{ det }\bm{H}_{\tau_t}(\hat{\bm\theta}_{\tau_t})$. Then, for any arm $\bm{x}$, we have that, 
    \[\modulus{\bm{\rho}^\top\bm{z}(\bm{x} , \thetastar) - \bm{\rho}^\top\bm{z}(\bm{x} , \hat{\bm{\theta}}_\tau)} \leq \epsilon_1(t , \tau_t , \bm{x}) + \epsilon_2(t , \tau_t , \bm{x})\]

    where 
    \[\epsilon_1(t , \tau_t , \bm{x}) = \sqrt{2}\gamma(\delta)\twonorm{\bm{H}_t(\hat{\bm\theta}_{\tau_t})^{-1/2}(\bm{I} \otimes \bm{x})\bm{A}(\bm{x} , \hat{\bm{\theta}}_{\tau_t})\bm{\rho}}\]
    \[\epsilon_2(t  ,\tau_t , \bm{x}) = 6R\gamma(\delta)^2\twonorm{(\bm{I} \otimes \bm{x}^\top)\bm{H}_t(\hat{\bm\theta}_{\tau_t})^{-1/2}}^2\]
    \label{lemma_RS:UCB-LCB}
\end{lemma}
\begin{proof}
The proof follows on the same lines as that of Lemma \ref{lemma:UCB-LCB} and uses the fact that $\frac{\textrm{det }\bm{H}_{\tau_t}(\hat{\bm\theta}_{\tau_t})\inv}{\textrm{det } \bm{H}_t(\hat{\bm\theta}_{\tau_t})\inv}\leq 2$ combined with Lemma \ref{lemma: determinant_condition} to convert $\bm{H}_{\tau_t}(\hat{\bm\theta}_{\tau_t})$ to $\bm{H}_t(\hat{\bm\theta}_{\tau_t})$.
\end{proof}

\begin{lemma}
    Let $\tau_t \leq t$ be the last time step at which a switch was made. Let $\epsilon_1(t , \tau_t , \bm{x})$ and $\epsilon_2(t , \tau_t , \bm{x})$ be as defined in Lemma \ref{lemma_RS:UCB-LCB}. Then, the regret at time step $t$ can be bounded as 
    \[\modulus{\bm{\rho}^\top\bm{z}(\bm{x}^* , \thetastar) - \bm{\rho}^\top\bm{z}(\bm{x}_t , \thetastar)} \leq  2\epsilon_1(t , \tau_t , \bm{x}_t) + 2\epsilon_2(t , \tau_t , \bm{x}_t)\]
    \label{Lemma_RS: per_round_regret}
\end{lemma}
\begin{proof}
\begin{align*}
    \modulus{\bm{\rho}^\top\bm{z}(\bm{x}^\star , \thetastar) - \bm{\rho}^\top\bm{z}(\bm{x}_t , \thetastar)} &\overset{(i)}{\leq} \bm{\rho}^\top\bm{z}(\bm{x}^\star , \hat{\bm\theta}_{\tau_t} ) + \epsilon_1(t , \tau_t , \bm{x}^\star) + \epsilon_2(t , \tau_t , \bm{x}^\star) - \bm{\rho}^\top\bm{z}(\bm{x}_t , \hat{\bm\theta}_{\tau_t})+ \epsilon_1(t , \tau_t , \bm{x}_t) + \epsilon_2(t , \tau_t , \bm{x}_t)
    \\
    &\overset{(ii)}{\leq} 2\epsilon_1(t , \tau_t , \bm{x}_t) + 2\epsilon_2(t , \tau_t , \bm{x}_t)
\end{align*}

where $(i)$ follows from Lemma \ref{lemma_RS:UCB-LCB} and $(ii)$ follows from the fact that $\bm{x}_t = \argmax\limits_{\bm{x}\in\mathcal{X}}\textrm{UCB}(t , \tau_t , \bm{x}) = \argmax\limits_{\bm{x}\in\mathcal{X}} \left[ \bm{\rho}^\top \bm{z}(\bm{x} , \hat{\bm\theta}_{\tau_t}) + \epsilon_1(t , \tau_t , \bm{x}) + \epsilon_2(t , \tau_t , \bm{x}) \right]$ and hence, gets selected at round $t$.
\end{proof}

\begin{lemma} Let $B_t(\bm{x)}$ be as defined in Section \ref{appendix: rs_mnl_notation}. Then, we have that
\[\sqrt{B_t(\bm{x)}} \leq e^{3S} \kappa^{1/2} \gamma(\delta) \lVert \bm{x} \rVert_{\bm{V}_t^{-1}}  + 1\]
\label{Lemma_RS: bound_on_B}
\end{lemma}
\begin{proof}
    \begin{align*}
    \sqrt{B_t(\bm{x})} &= \exp\pbrak{\sqrt{6} \min\cbrak{ \kappa^{1/2} \gamma(\delta) \lVert \bm{x} \rVert_{\bm{V}_t^{-1}}, S}} 
    \\
    &\overset{(i)}{\leq}e^{3S} \kappa^{1/2} \gamma(\delta) \lVert \bm{x} \rVert_{\bm{V}_t^{-1}}  + 1
    \end{align*}

where $(i)$ follows from Lemma \ref{lemma: exp_relation} by choosing $\min\cbrak{\kappa^{1/2} \gamma(\delta) \lVert \bm{x} \rVert_{\bm{V}_t^{-1}} , S} = \kappa^{1/2} \gamma(\delta) \lVert \bm{x} \rVert_{\bm{V}_t^{-1}}$ and $M = \sqrt{6}S$.

\end{proof}

\begin{lemma}
Let $\tilde{\bm{X}}_\tau (\bm\theta)$ and $\tilde{\bm{x}}^{(i)}_\tau (\bm\theta)$ be defined as in Section \ref{appendix: rs_mnl_notation}. Then, we have 
    \[\tilde{\bm{X}}_\tau(\bm\theta) \tilde{\bm{X}}_\tau (\bm\theta)^\top = \sum\limits_{i=1}^{K}\tilde{\bm{x}}^{(i)}_\tau (\bm\theta) \tilde{\bm{x}}^{(i)}_\tau(\bm\theta)^\top\]
    \label{lemma_RS: decomposition of x_tilde}
\end{lemma}
\begin{proof}
    The proof follows on the same lines as Lemma \ref{lemma_OD: decomposition of x_tilde}.
\end{proof}

\begin{lemma}
    Let $\bm{M} \in \R^{Kd}$ be any positive-semidefinite matrix. Then,
    \[\eigmax{\tilde{\bm{X}}_\tau (\bm\theta)^\top \bm{M} \tilde{\bm{X}}_\tau (\bm\theta)} \leq \sum\limits_{i=1}^{K} \matnorm{\tilde{\bm{x}}^{(i)}_\tau (\bm\theta)}{\bm{M}}^2\]
    \label{lemma_RS: eig_relationship_between_x_tilde}
\end{lemma}
\begin{proof}
    The proof follows on the same lines as Lemma \ref{lemma_OD: eig_relationship_between_x_tilde}.
\end{proof}

\begin{lemma}
    Let $\epsilon_1(t , \tau , \bm{x})$ be as defined in Lemma \ref{lemma_RS:UCB-LCB}, and $\tau_t$ be the last switching round before round $t$. Then, we have that
    \[\sum\limits_{t \in [T]} \epsilon_1(t , \tau_t , \bm{x}_t) \leq  8 R K d \log T \kappa^{1/2} e^{3S} \gamma(\delta)^2 + 4 R K d^{1/2} (\log T)^{1/2} \gamma(\delta)\sqrt{T}  \]
    
    \label{lemma_RS: bound on epsilon_1}
\end{lemma}
\begin{proof}
\begin{align*}
    \sum\limits_{t \in [T]} \epsilon_1(t , \tau_t , \bm{x}_t) &=  \sqrt{2}\gamma(\delta) \sum\limits_{t \in [T]} \twonorm{\bm{H}_t(\hat{\bm\theta}_{\tau_t})^{-1/2}(\bm{I} \otimes \bm{x}_t)\bm{A}(\bm{x}_t , \hat{\bm{\theta}}_{\tau_t})\bm{\rho}}
    \\
    &\overset{(i)}{\leq}\sqrt{2}\gamma(\delta) \sum\limits_{t \in [T]} \twonorm{\bm{H}_t(\hat{\bm\theta}_{\tau_t})^{-1/2}(\bm{I} \otimes \bm{x}_t)\bm{A}(\bm{x}_t , \hat{\bm{\theta}}_{\tau_t})^{1/2}}\matnorm{\bm{\rho}}{\bm{A}(\bm{x}_t , \hat{\bm{\theta}}_{\tau_t})}\\
    &\leq \sqrt 2 R\gamma(\delta) \sum\limits_{t \in [T]}\twonorm{\bm{A}(\bm{x}_t , \hat{\bm{\theta}}_{\tau_t})^{1/2}(\bm{I} \otimes \bm{x}_t^\top) {\bm{H}_t(\hat{\bm\theta}_{\tau_t})^{-1/2}}}
    \\
    &\overset{(ii)}{\leq} \sqrt 2 R\gamma(\delta) \sum\limits_{t \in [T]}\twonorm{\sqrt{B_t(\bm{x}_t)} \tilde{\bm{X}}_t(\hat{\bm\theta}_{\tau_t})^\top {\bm{H}_t(\hat{\bm\theta}_{\tau_t})^{-1/2}}}
    \\
     &\overset{(iii)}{\leq} \sqrt{2} R\gamma(\delta) \sum\limits_{t \in [T]}\twonorm{\tilde{\bm{X}}_t(\hat{\bm\theta}_{\tau_t})^\top {\bm{H}_t(\hat{\bm\theta}_{\tau_t})^{-1/2}}} \left\{ e^{3S} \kappa^{1/2} \gamma(\delta) \lVert \bm{x}_t \rVert_{\bm{V}_t^{-1}}  + 1 \right\}
     \\
\end{align*}
where $(i)$ follows from $\twonorm{\bm{A}\bm{x}} \leq \twonorm{\bm{A}}\twonorm{\bm{x}}$, $(ii)$ follows from the definition of $\tilde{\bm{X}}(\bm\theta)$, and $(iii)$ follows from Lemma \ref{Lemma_RS: bound_on_B}.

We now bound the term $\sum\limits_{t \in [T]}\twonorm{\tilde{\bm{X}}_t(\hat{\bm\theta}_{\tau_t})^\top {\bm{H}_t(\hat{\bm\theta}_{\tau_t})^{-1/2}}}$:
\begin{align*}
     \sum\limits_{t \in [T]}\twonorm{\tilde{\bm{X}}_t(\hat{\bm\theta}_{\tau_t}) {\bm{H}_t(\hat{\bm\theta}_{\tau_t})^{-1/2}}} &=  \sum\limits_{t \in [T]}\sqrt{\eigmax{\bm{H}_t(\hat{\bm\theta}_{\tau_t})^{-1/2} \tilde{\bm{X}}_t(\hat{\bm\theta}_{\tau_t}) \tilde{\bm{X}}_t(\hat{\bm\theta}_{\tau_t})^\top \bm{H}_t(\hat{\bm\theta}_{\tau_t})^{-1/2}}}
   \\
    &=  \sum\limits_{t \in [T]}\sqrt{\eigmax{ \tilde{\bm{X}}_t(\hat{\bm\theta}_{\tau_t})^\top \bm{H}_t(\hat{\bm\theta}_{\tau_t})^{-1} \tilde{\bm{X}}_t(\hat{\bm\theta}_{\tau_t})}}
    \\
    &\overset{(i)}{=}  \sum\limits_{t \in [T]}\sqrt{\sum\limits_{i=1}^{K} \matnorm{\tilde{\bm{x}}^{(i)}_t(\hat{\bm\theta}_{\tau_t})}{\bm{H}_t(\hat{\bm\theta}_{\tau_t})^{-1}}^2}
    \\
    &\overset{(ii)}{\leq}  \sum\limits_{t \in [T]}\sqrt{\sum\limits_{i=1}^{K} \matnorm{\tilde{\bm{x}}^{(i)}_t(\hat{\bm\theta}_{\tau_t})}{\bm{H}_t^{i}(\hat{\bm\theta}_{\tau_t})^{-1}}^2}
    \\
    &\overset{(iii)}{\leq}  \sqrt T\sqrt{\sum\limits_{t \in [T]}\sum\limits_{i=1}^{K} \matnorm{\tilde{\bm{x}}^{(i)}_t(\hat{\bm\theta}_{\tau_t})}{\bm{H}_t^{i}(\hat{\bm\theta}_{\tau_t})^{-1}}^2}
    \\
    &\overset{(iv)}{\leq} 2K \sqrt{d T \log T}
\end{align*}
where $(i)$ follows from Lemma \ref{lemma_RS: eig_relationship_between_x_tilde}, $(ii)$ follows from Lemma \ref{lemma_Rs: relation between H and H for x_tilde}, $(iii)$ follows from Cauchy-Schwarz, and $(iv)$ follows from Lemma \ref{lemma: elliptical potential lemma} and the fact that $\twonorm{\tilde{\bm{x}}^{(i)} (\bm\theta)} \leq \twonorm{\bm{A}(\bm{x} , \bm{\theta})}\twonorm{\bm{x}} \leq 1$.

We also bound the term $\sum\limits_{t \in [T]}\twonorm{\tilde{\bm{X}}_t(\hat{\bm\theta}_{\tau_t})^\top {\bm{H}_t(\hat{\bm\theta}_{\tau_t})^{-1/2}}} \lVert \bm{x}_t \rVert_{\bm{V}_t^{-1}}$ as follows:

\begin{align*}
    \sum\limits_{t \in [T]}\twonorm{\tilde{\bm{X}}_t(\hat{\bm\theta}_{\tau_t})^\top {\bm{H}_t(\hat{\bm\theta}_{\tau_t})^{-1/2}}} \lVert \bm{x}_t \rVert_{\bm{V}_t^{-1}} &\overset{(i)}{\leq} \sqrt{ \sum\limits_{t \in [T]}\twonorm{\tilde{\bm{X}}_t(\hat{\bm\theta}_{\tau_t})^\top {\bm{H}_t(\hat{\bm\theta}_{\tau_t})^{-1/2}}}^2} \sqrt{ \sum\limits_{t \in [T]}\lVert \bm{x}_t \rVert_{\bm{V}_t^{-1}}^2}
    \\
    &\overset{(ii)}{\leq} 2 K \sqrt{d \log T}  \sqrt{ \sum\limits_{t \in [T]}\lVert \bm{x}_t \rVert_{\bm{V}_t^{-1}}^2}
    \\
    &\overset{(ii)}{\leq} 4 K d \log T
\end{align*}

where $(i)$ follows from Cauchy-Schwarz, $(ii)$ follows from the same steps used to bound $\sum\limits_{t \in [T]}\twonorm{\tilde{\bm{X}}_t(\hat{\bm\theta}_{\tau_t})^\top {\bm{H}_t(\hat{\bm\theta}_{\tau_t})^{-1/2}}}$ above, and $(iii)$ follows from Lemma \ref{lemma: elliptical potential lemma}.

Substituting back, we get
\begin{align*}
    \sum\limits_{t \in [T]} \epsilon_1(t , \tau_t , \bm{x}_t) &\leq 4 \sqrt{2} R K d \log T \kappa^{1/2} e^{3S} \gamma(\delta)^2 + 2\sqrt{2} R K d^{1/2} (\log T)^{1/2} \gamma(\delta)\sqrt{T} 
    \\
    &\leq 8 R K d \log T \kappa^{1/2} e^{3S} \gamma(\delta)^2 + 4 R K d^{1/2} (\log T)^{1/2} \gamma(\delta)\sqrt{T} 
\end{align*}
\end{proof}

\begin{lemma}
    Let $\epsilon_2(t , \tau , \bm{x})$ be as defined in Lemma \ref{lemma_RS:UCB-LCB}, and $\tau_t$ be the last switching round before round $t$. Then, we have that
    \[\sum\limits_{t \in [T]} \epsilon_2(t , \tau_t , \bm{x}_t) \leq 24 d R K^2 e^{2S} \kappa \gamma(\delta)^2 \log T   \]
    \label{lemma_RS: bound on epsilon_2}
\end{lemma}
\begin{proof}
\begin{align*}
    \sum\limits_{t \in [T]} \epsilon_2(t , \tau , \bm{x}_t) &= 6R\gamma(\delta)^2 \sum\limits_{t \in [T]} \twonorm{(\bm{I} \otimes \bm{x}^\top)\bm{H}_t(\hat{\bm\theta}_{\tau_t})^{-1/2}}^2
    \\
    &\overset{(i)}{=} 6R\gamma(\delta)^2 \sum\limits_{t \in [T]} \twonorm{\bm{A}(\bm{x}_t , \bm{\hat{\bm{\theta}}}_\tau)^{-1/2}}\twonorm{\tilde{\bm{X}}_t(\hat{\bm\theta}_{\tau_t}) \bm{H}_t(\hat{\bm\theta}_{\tau_t})^{-1/2}}^2 B_t(\bm{x}_t)
    \\
    &\overset{(ii)}{\leq} 6R\gamma(\delta)^2 e^{2S} \sum\limits_{t \in [T]} \twonorm{\bm{A}(\bm{x}_t , \bm{\hat{\bm{\theta}}}_{\tau_t})^{-1/2}}^2\twonorm{\tilde{\bm{X}}_t(\hat{\bm\theta}_{\tau_t}) \bm{H}_t(\hat{\bm\theta}_{\tau_t})^{-1/2}}^2 
    \\
    &\overset{(iii)}{\leq} 6 R\gamma(\delta)^2 e^{2S} \kappa \sum\limits_{t \in [T]} \twonorm{\tilde{\bm{X}}_t(\hat{\bm\theta}_{\tau_t}) \bm{H}_t(\hat{\bm\theta}_{\tau_t})^{-1/2}}^2 
    \\
    &\overset{(iv)}{\leq} 24 d R K^2 e^{2S} \kappa \gamma(\delta)^2 \log T 
\end{align*}
where $(i)$ follows from the definition of $\tilde{\bm{X}}$ and Lemma \ref{lemma_Rs: relation between H and H for x_tilde}, $(ii)$ follows from the definition of $B_t(\bm{x})$, $(iii)$ follows from the fact that $\bm{A}(\bm{x} , \bm{\theta}) \leq \frac{1}{\kappa}\bm{I}$, and $(iv)$ follows from the methods used in Lemma \ref{lemma_RS: bound on epsilon_1}.
\end{proof}

\begin{lemma}
Let Algorithm \ref{alg : RS-MNL} be run for $t$ rounds. Then, the switching criterion is triggered a maximum of $dK \log(1 + \frac{t}{d\lambda})$ times.
\label{lemma: number_switches}
\end{lemma}
\begin{proof}

 Let $\tau_0, \tau_1,\ldots,\tau_m \in [1,t]$ be the time steps at which the switching criterion in Algorithm \ref{alg : RS-MNL} is triggered, i.e, $2 \textrm{ det }\bm{H}_{\tau_i} (\hat{\bm\theta}_{\tau_i}) \leq \textrm{det }\bm{H}_{\tau_{i+1}} (\hat{\bm\theta}_{\tau_i})$ for $i \in [m-1]$, and $\tau_m = t$.
Note that $\bm{H}_{\tau_0}(\bm\theta) = \lambda \bm{I}_{Kd \times Kd}$.

\begin{align*}
\frac{\textrm{det }\bm{H}_{t}(\hat{\bm\theta}_{\tau_{m-1}})}{\textrm{det }\bm{H}_{\tau_0} (\bm\theta)} &= \frac{\textrm{det }\bm{H}_{\tau_m}(\hat{\bm\theta}_{\tau_{m-1}})}{\textrm{det }\bm{H}_{\tau_{m-1}}(\hat{\bm\theta}_{\tau_{m-1}})}\times \frac{\textrm{det }\bm{H}_{\tau_{m-1}}(\hat{\bm\theta}_{\tau_{m-2}})}{\textrm{det }\bm{H}_{\tau_{m-2}}(\hat{\bm\theta}_{\tau_{m-2})}}\times\ldots\times\frac{\textrm{det }\bm{H}_{\tau_1}(\hat{\bm\theta}_{\tau_{0}})}{\textrm{det }\bm{H}_{\tau_{0}} (\bm\theta)}\\
&\geq 2^m
\end{align*}
and hence, $\textrm{det }\bm{H}_t(\hat{\bm\theta}_{\tau_{m-1}}) \geq 2^m \lambda^{Kd}$ since $\textrm{det }\bm{H}_1 = \lambda^{Kd}$.
Also, we can say that:

\begin{align*}
\textrm{det }\bm{H}_t(\hat{\bm\theta}_{\tau_{m-1}}) &\overset{(i)}\leq \left(\frac{\textrm{trace }\bm{H}_t(\hat{\bm\theta}_{\tau_{m-1})}}{Kd}\right)^{Kd}\\
&\overset{(ii)}\leq \left(\frac{\sum_{i \in [K]}\textrm{trace }\bm{H}_t^i(\hat{\bm\theta}_{\tau_{m-1}})}{Kd}\right)^{Kd} \\
&\overset{(iii)}\leq \left(\frac{\lambda Kd + \sum_{i \in [K]}\sum_{s \in [t]} \|\tilde{\bm{x}}^{(i)}_s(\hat{\bm\theta}_{\tau_{m-1}})\|_2^2}{Kd}\right)^{Kd} \\
&\overset{(iv)}\leq \left( \lambda + \frac{t}{d}\right)^{Kd}
\end{align*}

Here $(i)$ follows from Lemma \ref{lemma: determinant-trace inequality}, $(ii)$ follows from Lemma \ref{lemma_Rs: relation between H and H for x_tilde} alongside the linearity of the trace operator, $(iii)$ follows from the definition of $\bm{H}^i_t(\bm\theta)$ and the fact that the only non-zero eigenvalue of $\bm{x} \bm{x}^\top$ is $\lVert \bm{x} \rVert_2^2$, and $(iv)$ is due to the fact that $\|\tilde{\bm{x}}_t^{(i)}(\bm\theta)\|_2 \leq \|\bm{A}(\bm{x}_t , \bm\theta)\| \leq 1$.
Thus, we have
\[
2^m\lambda^{Kd} \leq \det(\bm{H}_t (\hat{\bm\theta}_{\tau_{m-1}})\leq \left(\lambda + \frac{t}{d}\right)^{Kd}
\]
and hence, $2^m \leq \pbrak{1 + \frac{t}{\lambda d}}^{Kd}$. This finishes the proof.

\end{proof}

\section{General Lemmas and Results}

\begin{lemma}
    (Self-Concordance) Let $\bm{A}(\bm{x} , \bm\theta) = \nabla \bm{z}(\bm{x} , \bm\theta)$. Then, $\bm{A}(\bm{x} , \bm\theta)$ is $(M,v)-$generalized self-concordant with $v = 1$ and $M = \sqrt{6}$. In other words, for any given $\bm{x}_1 , \bm{x}_2 , \bm{\theta}_1 , \bm{\theta}_2$, denote $\bm{A}_1 = \bm{A}(\bm{x}_1 , \bm\theta_1)$ and $\bm{A}_2 = \bm{A}(\bm{x}_2 , \bm\theta_2)$. Then, we have
    \[\bm{A}_2\exp\pbrak{-\sqrt{6}\twonorm{\pbrak{\bm{I} \otimes \bm{x}_1^\top}\bm{\theta}_1 - \pbrak{\bm{I} \otimes \bm{x}_2^\top}\bm{\theta}_2}} \mleq \bm{A}_1 \mleq \bm{A}_2 \exp\pbrak{\sqrt{6}\twonorm{\pbrak{\bm{I} \otimes \bm{x}_1^\top}\bm{\theta}_1 - \pbrak{\bm{I} \otimes \bm{x}_2^\top}\bm{\theta}_2}}\]
    \label{lemma: self-concordance}
\end{lemma}


\begin{lemma}
    (Lemma 13, \cite{Amani2021}) Let $\beta = \{t_1 , \ldots , t_N\}$ be a set of time indices and define
    \[\bm{G}_{\beta}(\bm{\theta}_1 , \bm{\theta}_2) = \sum\limits_{t \in \beta} \bm{M}(\bm{x} , \bm{\theta}_1 , \bm{\theta}_2) \otimes \bm{x}_t\bm{x}_t^\top + \lambda \bm{I}_{Kd \times Kd}\]
    and 
    \[\bm{H}_\beta^\star = \sum\limits_{t \in \beta} \bm{A}(\bm{x}_t , \bm\theta^\star) \otimes \bm{x}_t\bm{x}_t^\top + \lambda \bm{I}_{Kd \times Kd}\]
    where 
    \[\bm{M}(\bm{x} , \bm\theta_1 , \bm\theta_2) =  \int\limits_{0}^1 \bm{A}(\bm{x} , v\bm{\theta}_1 + (1-v)\bm{\theta}_2) \; \diff v\] 
    Then,
    \[\bm{G}_{\beta}(\bm\theta , \bm\theta^\star) \mgeq \frac{1}{1 + 2S} \bm{H}_\beta^\star\]
    \label{lemma: H-G relation}
\end{lemma}

\begin{lemma}
    Define the log-likelihood function as follows:
    \[\mathcal{L}_t(\bm{\theta}) = \sum\limits_{s=1}^{t-1}\sum\limits_{i=1}^{K} \mathbbm{1}\cbrak{y_s = i} \log\frac{1}{\bm{z}_i(\bm{x}_s , \bm{\theta})} + \frac{\lambda}{2} \lVert \bm\theta \rVert_2^2 \]
    Let $\hat{\bm{\theta}}$ be the MLE of $\thetastar$, i.e., $\hat{\bm{\theta}} = \argmin\limits_{\bm{\theta}} \mathcal{L}_t(\bm{\theta})$, then
    $$\sum\limits_{s=1}^{t-1}\bm{z}(\bm{x}_s , \hat{\bm{\theta}}) \otimes \bm{x}_s + \lambda \hat{\bm\theta}= \sum\limits_{s=1}^{t-1}\bm{m}_s \otimes \bm{x}_s$$
    where $\bm{m}_s = \pbrak{\mathbbm{1}\cbrak{y_s = 1} , \ldots , \mathbbm{1}\cbrak{y_s = K}}^\top$ is the user-response vector.
    \label{lemma: MLE}
\end{lemma}
\begin{proof} For the sake of convenience, define the loss incurred at round $t$ (without the regularization term) as 
\[l_t(\bm{\theta}) = \sum\limits_{i=1}^{K} \mathbbm{1}\cbrak{y_s = i} \log\frac{1}{\bm{z}_i(\bm{x}_s , \bm{\theta})}\]
Then, it is easy to see that
\begin{align*}
    \frac{\partial l_t(\bm{\theta})}{\partial \theta_m} &= -\sum\limits_{i=1}^{K} \mathbbm{1}\cbrak{y_s = i} \frac{1}{\bm{z}_i(\bm{x}_s , \bm{\theta})} \frac{\partial \bm{z}_i(\bm{x}_s , \bm{\theta})}{\partial \theta_m}\\
    &= -\sum\limits_{i=1}^{K} \mathbbm{1}\cbrak{y_s = i} \frac{1}{\bm{z}_i(\bm{x}_s , \bm{\theta})}\sbrak{\mathbbm{1}\cbrak{i=m}\bm{z}_i(\bm{x}_s , \bm{\theta}) - \bm{z}_i(\bm{x}_s , \bm{\theta})\bm{z}_m(\bm{x}_s , \bm{\theta})}\otimes\bm{x}_s\\
    &= \sbrak{\mathbbm{1}\cbrak{y_s = m} - \bm{z}_m(\bm{x}_s , \bm{\theta})}\otimes\bm{x}_s\\
 \end{align*}
and hence, 
$$\nabla l_t(\bm{\theta}) = \sbrak{\bm{m}_s - \bm{z}(\bm{x}_s , \bm{\theta})}\otimes\bm{x}_s$$

Since $\hat{\bm{\theta}} = \argmin\limits_{\bm{\theta}}\mathcal{L}_t(\bm{\theta}) $, we have that $\nabla\mathcal{L}_t(\hat{\bm{\theta}}) = \argmin\limits_{\bm{\theta}} \sum_{s=1}^{t-1} l_s(\hat{\bm\theta}) + \lambda \hat{\bm\theta} = 0$, which results in the claim.

\end{proof}

\begin{lemma}
    (Bernstein's Inequality) Let $X_1 , \ldots ,X_n$ be a sequence of independent random variables with $\modulus{X_t - \E\sbrak{X_t}} \leq b$. Let $S = \sum\limits_{t=1}^{n}\pbrak{X_t - \E\sbrak{X_t}}$ and $v = \sum\limits_{t=1}^{n} \mathbb{V}[X_t]$. Then, for any $\delta \in [0,1]$, we have
    \[\P\cbrak{S \geq \sqrt{2v\log \frac{1}{\delta}} + \frac{2b}{3}\log\frac{1}{\delta}} \leq \delta\]
    \label{lemma: bernstein}
\end{lemma}

\begin{lemma}
    Let $\bm{m}_s = \pbrak{\mathbbm{1}\cbrak{y_s = 1} , \ldots , \mathbbm{1}\cbrak{y_s = K}}$ be the user-response vector as defined in Section \ref{appendix: b_mnl_notation}. Then,
    \[\E\sbrak{\bm{m}_s} = \bm{z}(\bm{x}_s , \thetastar)\text{ and } \E\sbrak{\bm{m}_s\bm{m}_s^\top} = \textrm{diag}(\bm{z}(\bm{x}_s , \thetastar))\]
    \label{lemma: expectation of m}
\end{lemma}
\begin{proof}
    Since $\bm{m}_s = \pbrak{\mathbbm{1}\cbrak{y_s = 1} , \ldots , \mathbbm{1}\cbrak{y_s = K}}$,
 we have 
 \begin{align*}
     \E\sbrak{\bm{m}_s} &= \pbrak{\E\sbrak{\mathbbm{1}\cbrak{y_s = 1}} , \ldots , \E\sbrak{\mathbbm{1}\cbrak{y_s = K}}}
     = \pbrak{{z}_1(\bm{x}_s , \thetastar) , \ldots , {z}_K(\bm{x}_s , \thetastar)}
     = \bm{z}(\bm{x}_s , \thetastar)
 \end{align*}
For the second part, note that 
$$\sbrak{\bm{m}_s\bm{m}_s^\top}_{i,j} = \mathbbm{1}\cbrak{y_s = i}\mathbbm{1}\cbrak{y_s = j} = 
\begin{cases}
\mathbbm{1}\cbrak{y_s = i} & i = j\\
0 & i \neq j
\end{cases}
$$
Thus, we have 
\begin{align*}
    \E\sbrak{\bm{m}_s\bm{m}_s^\top} &= \E\sbrak{\textrm{diag}\pbrak{\mathbbm{1}\cbrak{y_s = 1} , \ldots, \mathbbm{1}\cbrak{y_s = K}}}
    = \textrm{diag}\pbrak{\E\sbrak{\mathbbm{1}\cbrak{y_s = 1}} , \ldots, \E\sbrak{\mathbbm{1}\cbrak{y_s = K}}}\\
    &= \textrm{diag}\pbrak{{z}_1(\bm{x}_s , \thetastar) , \ldots , {z}_K(\bm{x}_s , \thetastar)}
    = \textrm{diag}(\bm{z}(\bm{x}_s , \thetastar))
\end{align*}
\end{proof}

\begin{lemma}
    (Claim A.8, \cite{sawarni24}) For any $x \in [0,M]$, 
    $$e^x \leq e^M\pbrak{\frac{x}{M}} + 1$$
    \label{lemma: exp_relation}
\end{lemma}

\begin{lemma}
    (Theorem 5, \cite{Ruan2021}) Let $\pi$ represent the G-Optimal Distributional Design learnt from $\mathcal{X}_1 \ldots \mathcal{X}_s \overset{\text{i.i.d}}{\sim} \mathcal{D}$ and let $\bm{W}$ be the expected data matrix, i.e. $\bm{W} = \lambda\bm{I} + \E\limits_{\mathcal{X}\sim\mathcal{D}}\sbrak{\E\limits_{\bm{x}\sim\pi(\mathcal{X})}\bm{x}\bm{x}^\top|\mathcal{X}}$, then, we have
    \[P\cbrak{\E\limits_{\mathcal{X}\sim\mathcal{D}}\sbrak{\max\limits_{\bm{x}\in\mathcal{X}} \matnorm{\bm{x}}{\bm{W}\inv}} \leq O(\sqrt{d\log d})} \geq 1 - \exp\pbrak{O(d^4\log^2 d) - sd^{-12}2^{-16}}\]
    \label{lemma_OD: opt_design_wrt_Distributional_design}
\end{lemma}

\begin{lemma}
    (Lemma 4, \cite{Ruan2021}) Let $\pi_G$ represent the G-Optimal design and define the design matrix $\bm{W}_G = \lambda\bm{I} + \E\limits_{\mathcal{X} \sim \mathcal{D}}\sbrak{\E\limits_{\bm{x}\in\pi_G(\mathcal{X})} \bm{x}\bm{x}^\top \mid \mathcal{X}}$, then we have
    \[\E\limits_{\mathcal{X}\sim\mathcal{D}}\sbrak{\max\limits_{\bm{x}\in\mathcal{X}}\matnorm{\bm{x}}{\bm{W}_G\inv}^2} \leq O(d^2)\]
    \label{Lemma_OD: opt_design_wrt_G}
\end{lemma}

\begin{lemma}
    (Lemma A.15, \cite{sawarni24} , \cite{Ruan2021}) Let $\bm{x}_1 \ldots \bm{x}_n \sim \mathcal{D}$ be vectors with $\twonorm{x}\leq 1$, then
    \[\P\cbrak{3\epsilon N\bm{I} + \sum\limits_{i=1}^{n} \bm{x}_i\bm{x}_i^\top \mgeq \frac{n}{8}\E\limits_{\bm{x}\sim\mathcal{D}}\sbrak{\bm{x}\bm{x}^\top}} \geq 1 - 2d\exp\pbrak{-\frac{\epsilon n}{8}}\]
\end{lemma}

\begin{lemma}
    (Lemma 6, \cite{Zhang2023}) Let $\cbrak{\mathcal{F}_t}_{t=1}^\infty$ be a filteration and $\cbrak{\bm{x}_t}_{t=1}^\infty$ be a stochastic process in $\mathcal{B}_2(d) = \cbrak{\bm{x}\in\R^d \mid \twonorm{\bm{x}_t} \leq 1}$ such that $\bm{x}_t$ is $\mathcal{F}_t-$ measurable. Let $\cbrak{\bm{\epsilon_t}}_{t=1}^\infty$ be a martingale difference sequence such that $\bm{\epsilon}_t$ is $\mathcal{F}_{t+1}-$measurable. Assume that conditioned on $\mathcal{F}_t$, we have $\norm{\bm{\epsilon}_t}_1 \leq 2$ almost surely, and is denoted by $\bm{\eta_t} = \E\sbrak{\bm{\epsilon}_t\bm{\epsilon}_t^\top \mid \mathcal{F}_t}$. Let $\lambda > 0$ and for any $t \geq 1$, define
    \[\bm{S}_t = \sum\limits_{s=1}^{t-1} \bm{\epsilon}_s \otimes \bm{x}_s \text{ and } \bm{H}_t = \lambda \bm{I}_{dK \times dK} + \sum\limits_{s=1}^{t-1}\bm{\eta}_s \otimes \bm{x}_s\bm{x}_s^\top\]
    Then, for any $\delta \in (0,1)$, we have
    \[\P\cbrak{\exists t > 1, \matnorm{\bm{S}_t}{\bm{H}_t\inv} \geq \frac{\sqrt{\lambda}}{4} + \frac{4}{\sqrt{\lambda}}\log\pbrak{\frac{\text{det }\bm{H}_t^
    {1/2}}{\delta\lambda^{\frac{dK}{2}}}} + \frac{4}{\sqrt{\lambda}}Kd\log 2} \leq \delta\]
    \label{lemma: martingale confidence bound}
\end{lemma}

\begin{lemma}
    (Determinant-Trace Inequality) Let the determinant and trace of a p.s.d matrix $\bm{A}\in \R^{d \times d}$ be denoted by $\textrm{det } \bm{A}$ and $\textrm{trace } \bm{A}$. Then, we have
    \[\textrm{det } \bm{A} \leq \left( \frac{\textrm{trace } \bm{A}}{d} \right)^d\]
    \label{lemma: determinant-trace inequality}
\end{lemma}
\begin{proof}
    Let the eigenvalues of $\bm{A}$ be denoted by $\lambda(\bm{A}) \geq 0$ since $\bm{A} \mgeq 0$. Then, we know, $\textrm{det } \bm{A} = \prod \lambda(\bm{A})$ and $\textrm{trace } \bm{A} = \sum \lambda(\bm{A})$. Thus, applying the inequality for arithmetic means and geometric means, we get that
    \[\left(\prod \lambda(\bm{A}) \right)^{1/d} \leq \frac{\sum \lambda(\bm{A})}{d} \implies \textrm{det } \bm{A} \leq \left( \frac{\textrm{trace } \bm{A}}{d} \right)^d\]
\end{proof}

\begin{lemma}
    (Elliptical Potential Lemma, Lemma 11, \cite{AbbasiYadkori2011} ) Let $\cbrak{\bm{x}_s}_{s=1}^t$ represent a set of vectors in $\R^d$ and let $\twonorm{\bm{x}_s} \leq L$. Let $\bm{V}_s = \lambda\bm{I}_{d\times d} + \sum\limits_{m=1}^{s-1}x_mx_m^\top$. Then, for $\lambda \geq 1$
    \[\sum\limits_{s=1}^{t}\matnorm{\bm{x}_s}{\bm{V}_s\inv}^2 \leq 2d\log\pbrak{1 + \frac{tL^2}{\lambda d}} \leq 4d\log({tL^2})\]
    \label{lemma: elliptical potential lemma}
\end{lemma}

\begin{lemma}
    (Lemma 12, \cite{AbbasiYadkori2011}) If $\bm{A}\mgeq\bm{B}\mgeq0$, then
    $$\sup\limits_{\bm{x}\neq 0} \frac{\bm{x}^\top\bm{A}\bm{x}}{\bm{x}^\top\bm{B}\bm{x}} \leq \frac{\textrm{det }(\bm{A)}}{\textrm{det }(\bm{B})}$$
    \label{lemma: determinant_condition}
\end{lemma}

\section{Additional Experiments}
\label{appendix: Additional Experiments}

In this section, we supplement the experiments from Section \ref{sec: Experiments} (in particular, \textbf{Experiment 1} and \textbf{Experiment 2}). 

\begin{figure*}[h]
	\centering
	\begin{subfigure}[b]{0.45\columnwidth}  
		\centering 
		\includegraphics[width=59mm]{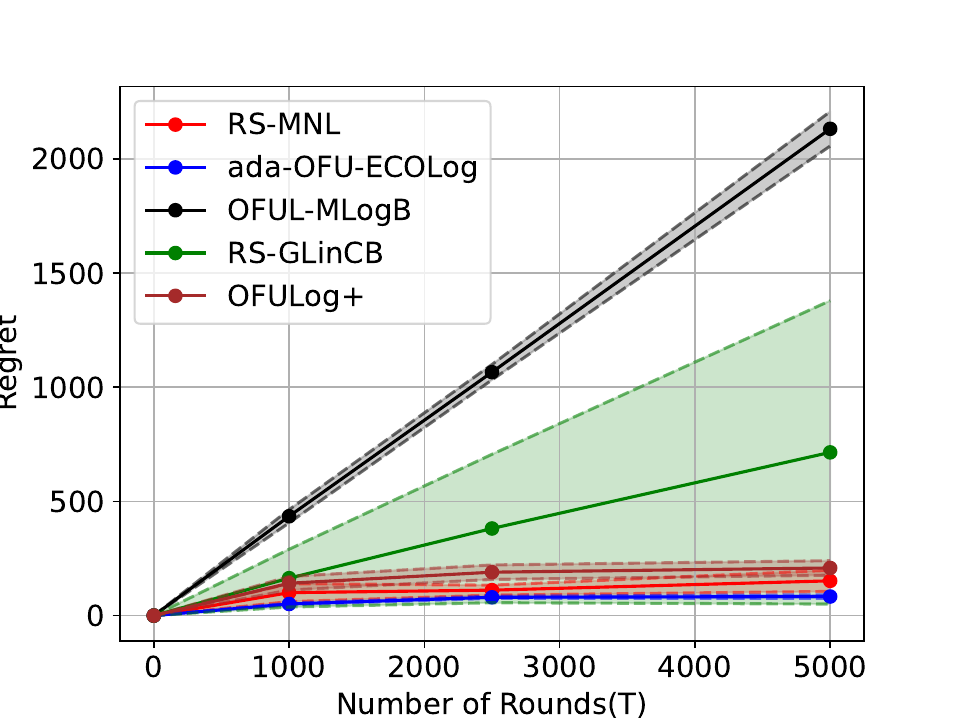}
		\caption[]{{\small Regret vs.\ $T$: Logistic ($K=1$) Setting}}   
		\label{fig:logistic_error}
	\end{subfigure}
	\hfill
	\begin{subfigure}[b]{0.45\columnwidth}   
		\centering 
	\includegraphics[width=59mm]{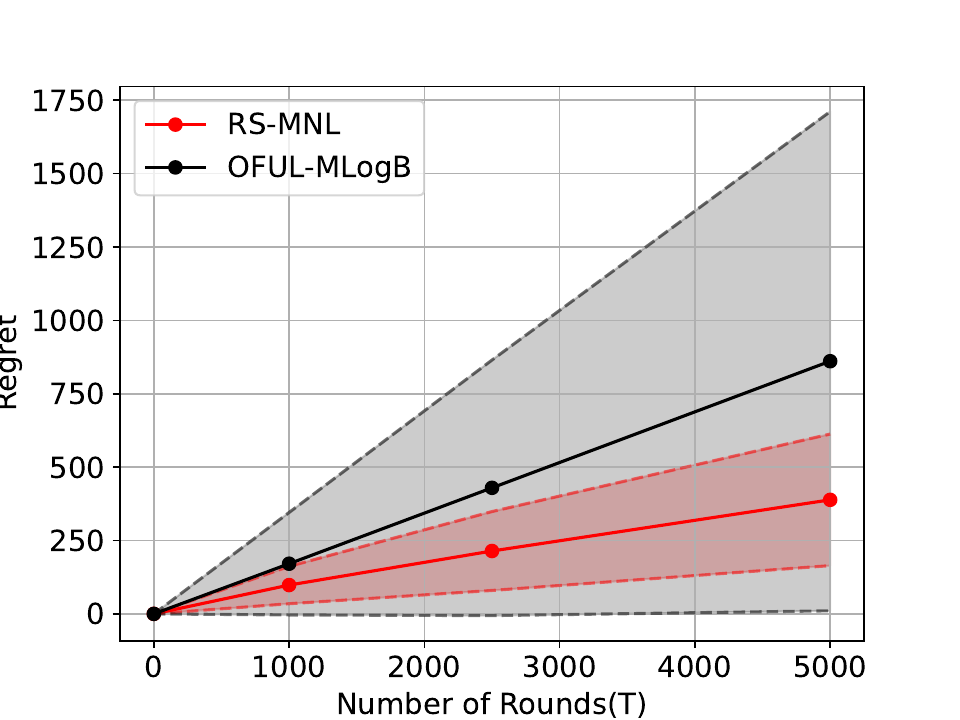}
		\caption[]{{\small Regret vs.\ $T$: $K = 3$}}   
		\label{fig:3 outcomes error}
	\end{subfigure}
\end{figure*}

\textbf{Experiment 1 ($R(T)$ vs.\ $T$ for the Logistic $(K=1)$ Setting):} In this experiment, we use the same instance as in \textbf{Experiment 1} (Section \ref{sec: Experiments}) and average the regret over 10 different seeds for sampling rewards. The averaged results with two standard deviations can be found in Figure \ref{fig:logistic_error}.

\textbf{Experiment 2 ($R(T)$ vs.\ $T$ for $K=3$):} In this experiment, we use the same instance as in \textbf{Experiment 2} (Section \ref{sec: Experiments}) and average the regret over 10 different seeds for sampling rewards. The averaged results with two standard deviations are reported in Figure \ref{fig:3 outcomes error}.

\end{document}